\documentclass[journal]{IEEEtran}
\ifCLASSINFOpdf
\else
\fi

\usepackage{graphicx}
\usepackage{comment}
\usepackage{times}
\usepackage{amsmath}
\usepackage{amssymb}
\usepackage{multirow}
\usepackage{color}
\usepackage{caption}
\usepackage{subcaption}
\usepackage{color}
\usepackage{hyperref}
\hypersetup{
	colorlinks=true,
	linkcolor=blue,
	filecolor=blue,
	urlcolor=blue,
	citecolor=cyan,
}
\begin{document}
%
\title{Towards Disentangling Latent Space for Unsupervised Semantic Face Editing}
%
%
%

\author{Kanglin Liu, Gaofeng Cao, Fei Zhou, Bozhi Liu, Jiang Duan, Guoping Qiu
\thanks{Kanglin Liu is with Pengcheng Laboratory, P.R.China, e-mail: max.liu.426@gmail.com}
\thanks{Guoping Qiu is with Shenzhen University and University of Nottingham,  e-mail: guoping.qiu@nottingham.ac.uk}
\thanks{Manuscript received *, *; revised *,*.}}

%
%

\markboth{}%
{Shell \MakeLowercase{\textit{et al.}}: Bare Demo of IEEEtran.cls for IEEE Journals}
%



\makeatletter
\let\@oldmaketitle\@maketitle
\renewcommand{\@maketitle}{\@oldmaketitle
	\begin{center}
		\captionsetup{type=figure}\setcounter{figure}{0}
		\centering
		\includegraphics[height=0.12\linewidth]{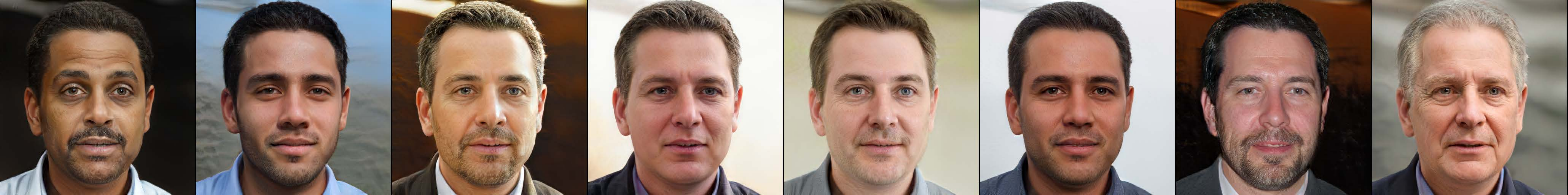}\\
		(a) synthesizing face images with identical $w_1$ but different $w_2$. \\
		\includegraphics[height=0.12\linewidth]{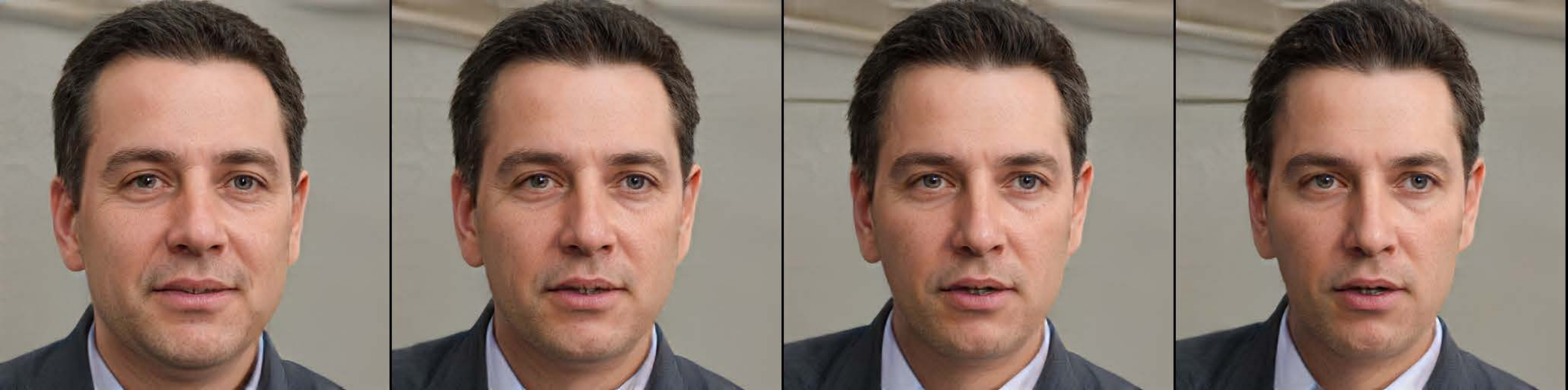}
		\includegraphics[height=0.12\linewidth]{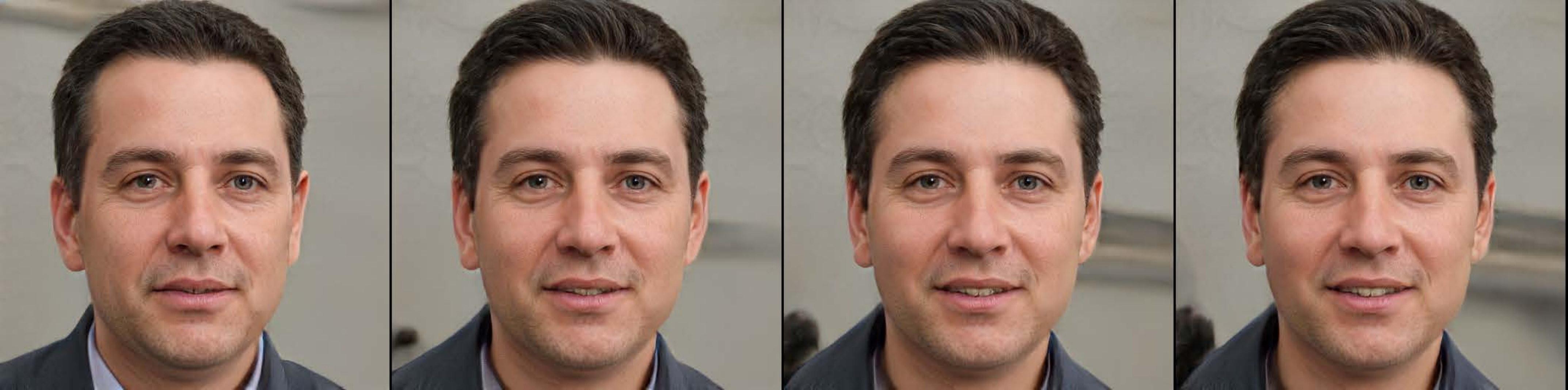}\\
		(b) round face $\rightarrow$ thin face \qquad \qquad \qquad \qquad \qquad \qquad \qquad \qquad 
		(c) no simile $\rightarrow$ simile\\
		\includegraphics[height=0.12\linewidth]{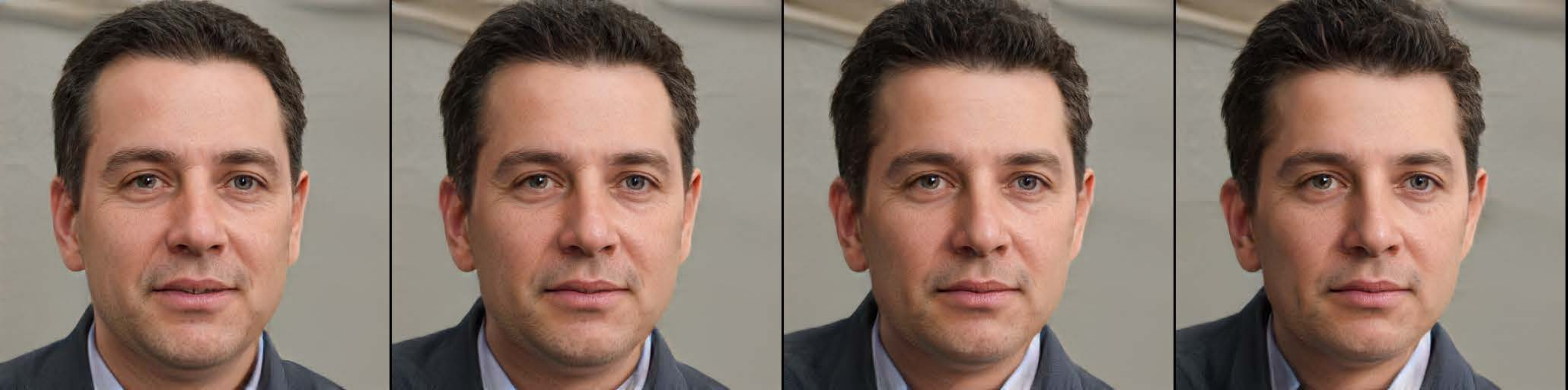}
		\includegraphics[height=0.12\linewidth]{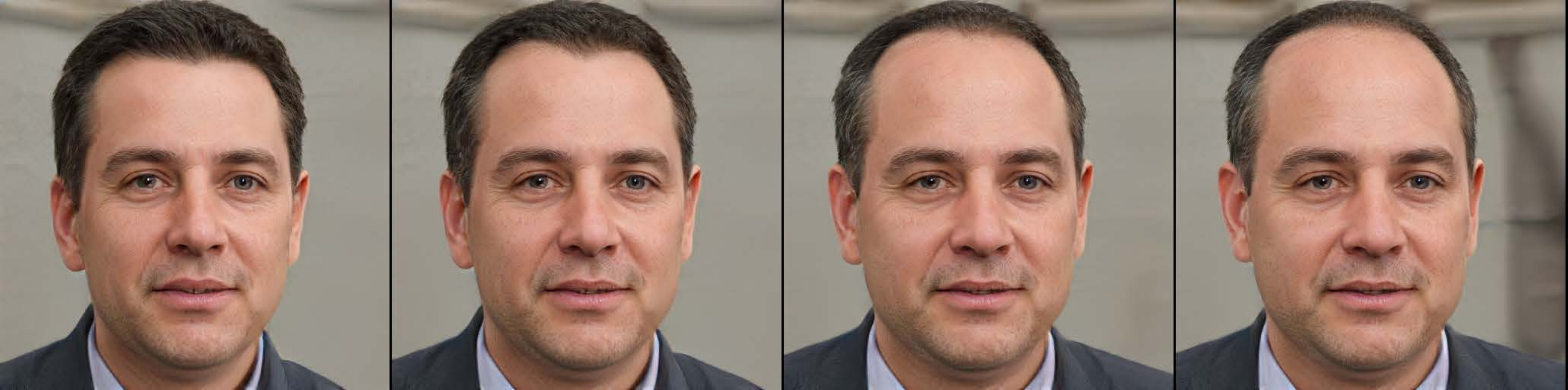}\\
		(d) hair $\rightarrow$ more hair \qquad \qquad \qquad \qquad \qquad \qquad \qquad \quad 
		(e) hair $\rightarrow$ almost no hair\\
		\includegraphics[height=0.12\linewidth]{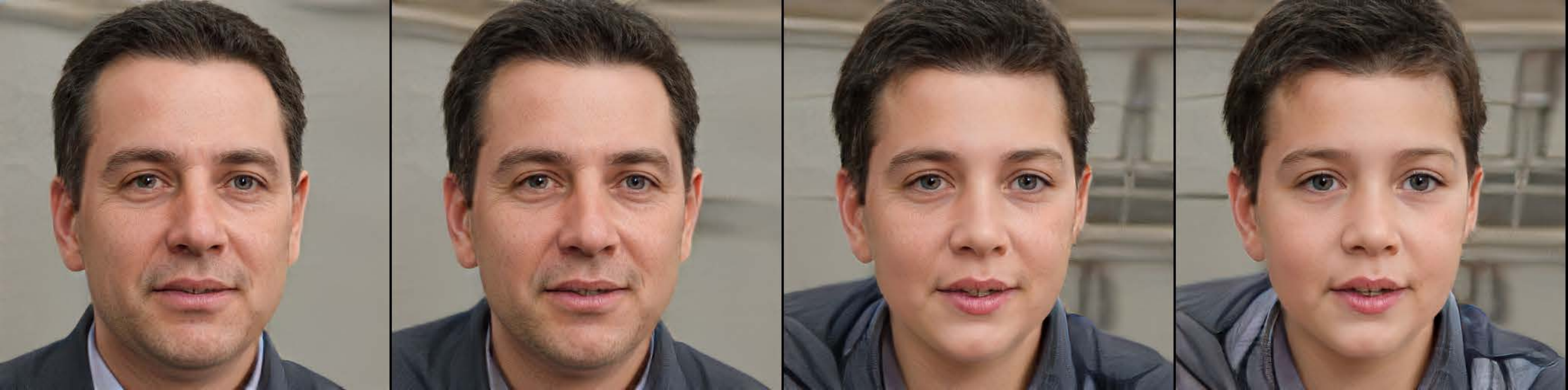}
		\includegraphics[height=0.12\linewidth]{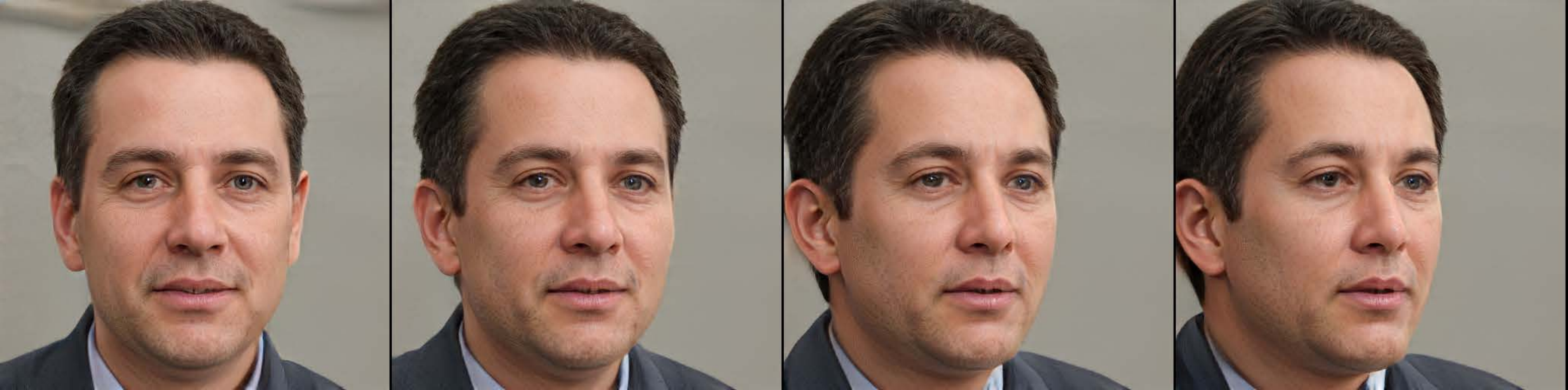}\\
		(f) adult $\rightarrow$ teenager\qquad \qquad \qquad \qquad \qquad \qquad \qquad \qquad \qquad \qquad 
		(g) pose\\
		\includegraphics[height=0.12\linewidth]{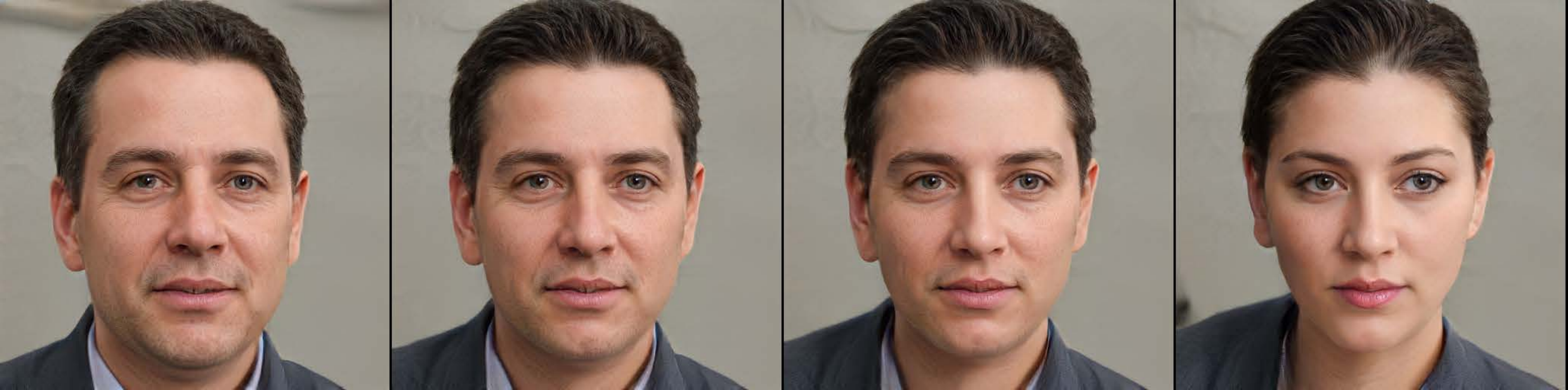}
		\includegraphics[height=0.12\linewidth]{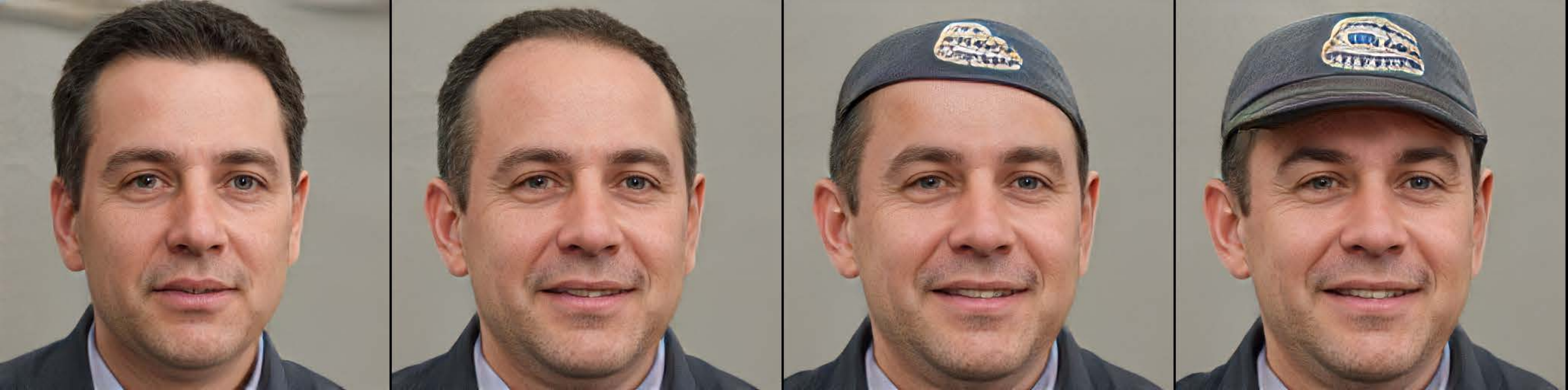}\\
		(h) male $\rightarrow$ female\qquad \qquad \qquad \qquad \qquad \qquad \qquad \qquad \quad 
		(i) no hat $\rightarrow$ hat\\
		\captionof{figure}{Attribute editing using the propsed STGAN-WO. (a) generating face images using identical $w_1$ but different $w_2$. Hence, all the faces share identical face shapes, \textit{i.e.}, the texture component, and have different structure parts. (b)-(i) synthesizing face images using identical $w_2$ but different $w_1$. Moving the intermediate latent code $w_1$ along its orthogonal directions can achieve the goal of attribute editing such that some specific attribites can be changed indivually.}
		\label{fig:fig1}
		\vspace*{-1.2cm}
	\end{center}
	\bigskip}
\makeatother

\maketitle


\begin{abstract}
Facial attributes in StyleGAN generated images are entangled in the latent space which makes it very difficult to independently control a specific attribute without affecting the others. Supervised attribute editing requires annotated training data which is difficult to obtain and limits the editable attributes to those with labels.
Therefore, unsupervised attribute editing in an disentangled latent space is key to performing neat and versatile semantic face editing. In this paper, we present a new technique termed Structure-Texture Independent Architecture with Weight Decomposition and Orthogonal Regularization (STIA-WO) to disentangle the latent space for unsupervised semantic face editing. By applying STIA-WO to GAN, we have developed a StyleGAN termed STGAN-WO which performs weight decomposition through utilizing the style vector to construct a fully controllable weight matrix to regulate image synthesis, and employs orthogonal regularization to ensure each entry of the style vector only controls one independent feature matrix. To further disentangle the facial attributes, STGAN-WO introduces a structure-texture independent architecture which utilizes two independently and identically distributed (i.i.d.) latent vectors to control the synthesis of the texture and structure components in a disentangled way. Unsupervised semantic editing is achieved by moving the latent code in the coarse layers along its orthogonal directions to change texture related attributes or changing the latent code in the fine layers to manipulate structure related ones. We present experimental results which show that our new STGAN-WO can achieve better attribute editing than state of the art methods$\footnote{The code is available at \href{https://github.com/max-liu-112/STGAN-WO}{STGAN-WO}}$.
\end{abstract}


%
\IEEEpeerreviewmaketitle

\section{Introduction}
%
%
%
%
\IEEEPARstart{G}{enerative} Adversarial Networks (GANs) \cite{Goodfellow2014} are one of the most significant developments in machine learning research of the past decades.
The rationale behind GANs is to learn the mapping from a latent distribution to the real data through adversarial training\cite{Radford2015,Arjovsky2017,Andrew2018,Gulrajani2017}. After learning such a non-linear mapping, GAN is capable of producing photo-realistic images from randomly sampled latent codes. The resolution and quality of images produced by GANs have seen rapid improvements \cite{Andrew2018,Karras2019}. Beyond this, efforts have been made towards intuitive control of the synthesis, for example, manipulating the facial expressions \cite{Karras2019}.

Conventional facial attribute manipulation can be achieved in a supervised  or unsupervised manner.  To be specific, the supervised scheme would treat the facial attribute manipulation as a image-to-image translation, where a particular aspect of a given image is changed to another, \textit{e.g.}, changing the facial expression of a person from smiling to frowning \cite{Choi2018,Huang2018,He2019,Lee2018}. Research studies have shown remarkable success in handling multi-domain image-to-image translation, where only a single model is used \cite{Choi2018}.
More recently, interpreting the latent semantics learned by GANs provides another way of semantic face editing \cite{Shen2020_cvpr,Zhu2020}. It has been found that the latent code of well-trained GAN models actually learns an entangled representation. 
Thus, precise control of facial attributes can be achieved by decoupling some entangled semantics via subspace projection \cite{Shen2020_cvpr}. 
To specify a domain for image-to-image translation or semantic interpretation, each face image needs to be annotated with different attributes, which is rather labor-consuming.

In contrast to the supervised methods, the unsupervised scheme does not rely on labeled datasets to conduct the facial attribute manipulation.
As the current state-of-the-art method for high-resolution images synthesis, StyleGANs \cite{Karras2019,Karras2020} utilize the style vector $S$ to adjust the style of the image at each convolutional layer via adaptive instance normalization (AdaIN) \cite{Karras2019} or its improved technique - weight demodulation \cite{Karras2020}, therefore directly controlling the strength of image features at different scales, and eventually leading to unsupervised separation of low- and high-level features. In other word, low-level features can be changed while maintaining high-level features, or vice verse. 
However, it is unclear how those high- or low-level features are related to the facial attributes. 
Attempts have been made towards better disentanglement by conducting principal directions variation \cite{Harkonen2020} in the latent/feature space or proposing a structured noise injection method, where the input noises are injected to GANs for controlling  specific parts of the generated images \cite{Alharbi2020}.
Such a noise injection method allows to change global or local features in a disentangled way. The noise injection method  shares the same problem as StyleGAN in that it is unclear how global or local features are related to the facial attributes.
Thus, it is incredibly difficult to change specific attributes individually. 

Following StyleGAN, our goal is towards disentangling the latent space for the purpose of achieving better face attribute editing in an unsupervised way. Firstly, we have found that the weight demodulation technique used in StyleGAN would lead to the entanglement problem between the weight matrix and the style vector $S$, inconsistent with the common goal of attribute disentanglement to find a latent space consists of linear subspace.
To address this, we introduce weight decomposition (WD), which constructs a controllable matrix via the style vector $S$, each entry of which corresponds to a specific feature matrix, and the orthogonal regularization would guarantee the orthogonality of each feature matrix. 
Thus, each entry in $S$ would only control one independent feature matrix, thus contributing to attribute disentanglement.

Besides, we introduce a structure-texture independent architecture (STIA) for better attribute disentanglement.
Based on  previous research, we have found that convolutional neural networks (CNNs) as well as facial attributes are structure or texture sensitive. For example, the texture component is closely related to attributes like facial outline, while the structure part has a strong relationship to attributes like face color, hair color, \textit{etc}.
Motivated by this, we reason that facial manipulation would benefit from independent generation of the texture and structure component.
Hence, STIA would generate the texture component first, then synthesize the corresponding structure part to obtain the final face image, where the multi-scale gradient strategy \cite{Karnewar2020} is used to guarantee stable training. Meanwhile, weight decomposition and orthogonal regularization are applied to control the synthesis of the structure and texture components.
To encourage the separation of structure and texture component, two independently and identically distributed latent vector $z_1$, $z_2$ are utilized to control the synthesis of the texture and structure parts, respectively. 
The texture component can be changed via $z_1$ while maintaining the structure component, or vice verse.
In addition, two independent gradient flows are constructed based on the multi-scale gradient strategy when calculating the adversarial loss, which encourages the outputs of the generator at multiple scales to match the structure and texture distributions individually, and therefore contributes to the separation of the structure and texture parts.
Eventually, as shown in Fig. \ref{fig:fig1}, better face editing is achieved in an unsupervised way by the proposed GAN model, referred to as STGAN-WO.
To summarize, our contributions are as follow:

(1) We introduce weight decomposition to control the image synthesis process. In contrast to weight demodulation, weight decomposition allows each entry in the style vector correspond to a specific feature matrix, whose orthogonality is guaranteed by the orthogonal regularization.

(2) We introduce the structure-texture independent architecture to hierarchically generate the texture and structure parts, thus allowing to edit specific attributes individually, and therefore contributing to attribute disentanglement.

(3) We experimentally confirm that the proposed techniques contribute to attribute disentanglement. Besides, STGAN-WO shows excellent performance on face editing tasks where specific attributes can be manipulated individually.

\begin{figure*}[htp]
	\centering
	\begin{subfigure}[b]{0.5\textwidth}
		\centering
		\includegraphics[height=0.5\linewidth]{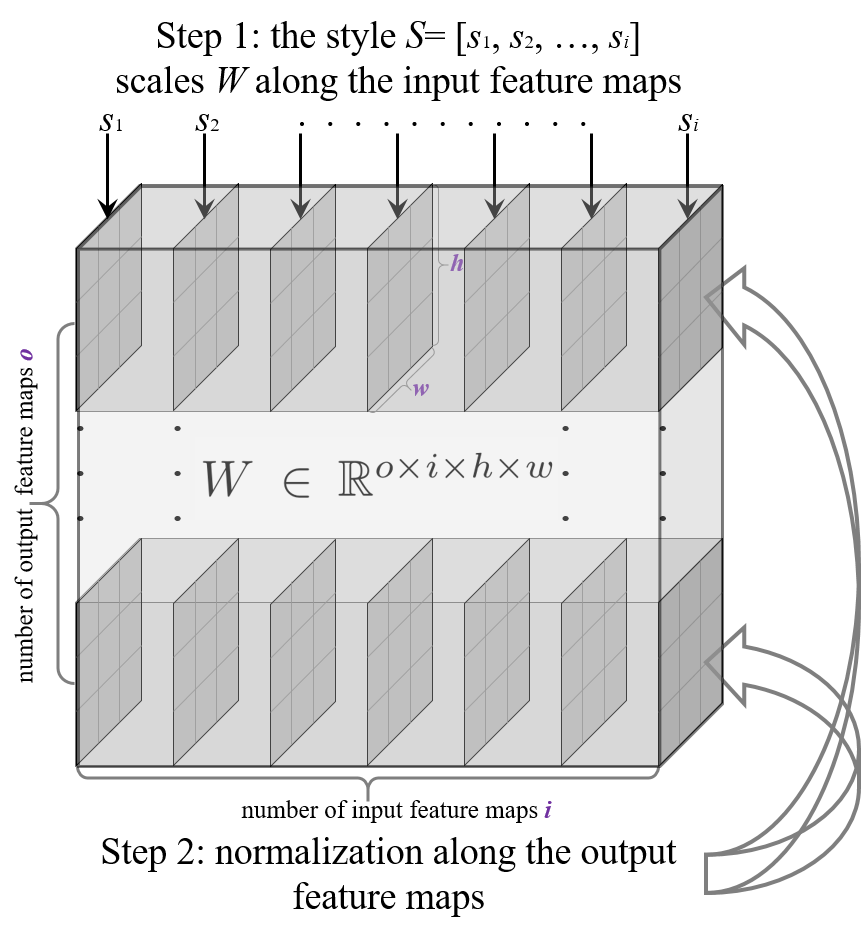}	
		\caption{weight demodulation}
	\end{subfigure}	
	\begin{subfigure}[b]{0.45\textwidth}
		\centering
		\includegraphics[height=0.5\linewidth]{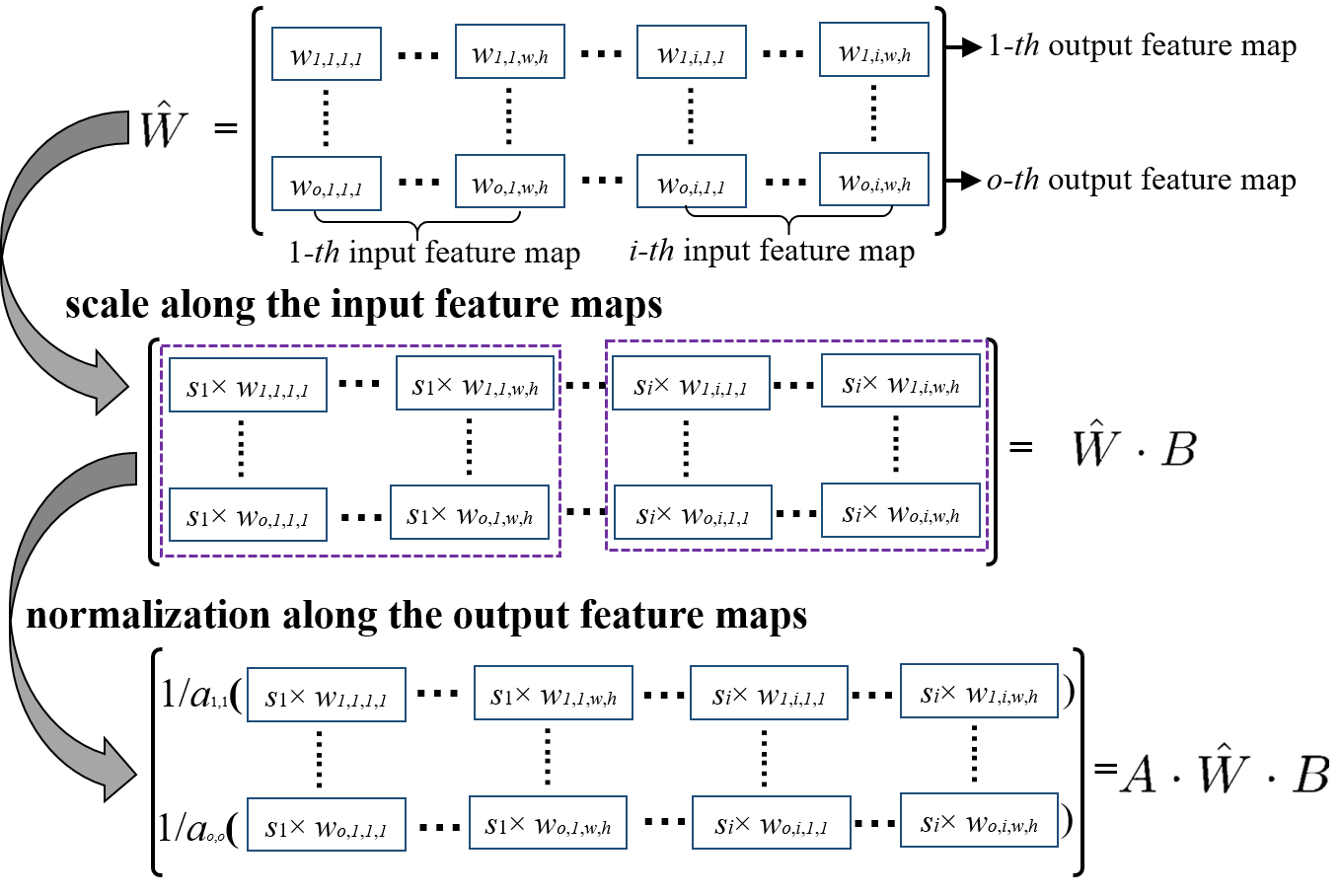}	
		\caption{equivalent form of weight demodulation}
	\end{subfigure}		
	\caption{Visualization of the weight demodulation  technique. (a) visualizes the weight demodulation. The style $S$ would channel-wisely scale the weight $W$ along the input feature maps, then normalization is conducted along the output feature maps. (b) shows the equivalent transformation of the weight demodulation. Scaling the weight $W$ along the input feature maps via $S$ is equivalent to the right multiplication of $\hat{W}$ by the diagonal matrix $B$, and normalization along the output feature maps is equivalent to the left multiplication of $\hat{W}$ by the diagonal matrix $A$.}
	\label{fig:fig2}
\end{figure*}

\section{Weight demodulation vs Weight decomposition}

\subsection{Weight Demodulation}

For easy discussion , we first briefly recap the essential idea of the weight demodulation technique proposed by the style-based GAN (StyleGAN) \cite{Karras2020}. The distinguishing feature of StyleGAN is its unconventional generator architecture. Instead of feeding the input latent code $\mathbf{z} \in \mathbb{Z}$ only to the beginning of the network, the mapping network $f$ first transforms it to an intermediate latent code $\mathbf{w} \in \mathbb{W}$. The affine transforms then produce styles $S$ that control the layers of the synthesis network $g$ via adaptive instance normalization (AdaIN) \cite{Karras2019,Huang2017,Ghiasi2017}, which normalizes the mean and variance of each feature map separately. However, AdaIN can potentially destroy any information found in the magnitudes of the features relative to each other, then causing characteristic blob-shaped artifacts in the generated images. To alleviate this, weight demodulation \cite{Karras2020} is applied in the improved version of StyleGAN. Instead of style control via AdaIN, weight demodulation conducts modulation on the convolutional weight matrix $W$ via the style vector $S$:
\begin{equation}\label{key1}
	W^{'}=\frac{W*S}{\sigma}
\end{equation}
\begin{equation}\label{key2}
	\sigma = \sqrt{\sum\limits_{i, h, w}(W*S)^2+\epsilon}
\end{equation}
where $W \in \mathbb{R}^{o\times i \times h \times w}$, $W^{'} \in \mathbb{R}^{o\times i \times h \times w}$ are the original and demodulated weights, respectively. $o$ and $i$ enumerate the output and input feature maps,  respectively. $h$ and $w$ represent the height and width of the kernel,  $S = [s_1, \cdots, s_i]$ is the style vector which scales the corresponding input feature maps, and $\epsilon$ is a small constant to avoid numerical issues. It is clearly seen that the weight $W$ is channel-wisely scaled by the incoming style $S$, then normalized by $\sigma$ along the output feature maps.

Fig. \ref{fig:fig2} shows the weight demodulation technique in detail. Specifically, weight demodulation in (\ref{key1}) is equivalent to the following formula.
\begin{equation}\label{key3}
	\hat{W}^{'}= A \cdot \hat{W} \cdot B
\end{equation}
where $\hat{W}^{'} \in \mathbb{R}^{out \times in}$, $\hat{W} \in \mathbb{R}^{out \times in}$ are the transformation of $W^{'}$ and $W$, respectively. $out = o$, and  $in=i \times h \times w$.  The transformation only gives a new shape to the weight matrix without changing its data.
$A \in \mathbb{R}^{out \times out}$  is a diagonal matrix, whose $n$-th diagonal entry $a_{n,n}$ is determined by:
\begin{equation}\label{key4}
	a_{n,n}=\frac{1}{\sqrt{\sum\limits_{o=n, i, h, w}(W*S)^2+\epsilon}} = \frac{1}{\sigma_n}
\end{equation}
$B \in \mathbb{R}^{in \times in}$ is a diagnol matrix as well, and its $n$-th diagonal entry $b_{n,n}$ is determined by:
\begin{equation}\label{key5}
	b_{n, n} = s_{mod(n, h \times w)}
\end{equation}
where  $mod(*, *)$ is the modulo operation, \textit{e.g.}, $b_{1,1} = s_1$ and $b_{in,in}=s_i$.

As indicated by Fig. \ref{fig:fig2} (b), scaling the weight $\hat{W}$ along the input feature maps via $S$ is equivalent to the right multiplication of $\hat{W}$ by the diagonal matrix $B$. In a similar way, normalization of $\hat{W}$ along the output feature maps via $\sigma$ is equivalent to the left multiplication of $\hat{W}$ by the diagonal matrix $A$.

Equation (\ref{key3}) gives an interpretation of the weight demodulation technique. To be specific, the style $S$ determines two diagonal matrices $A$ and $B$ to accordingly modulate the weight $W$. In the generation process, each style vector $S$ would determine a specific $W^{'}$ in the coarse layers and fine layers, thus controlling the low- and high-level features at corresponding convolutional layers, and further achieving the goal of style control.

However, as indicated by (\ref{key3}) and Fig. \ref{fig:fig2} (b), the weight matrix $W$ is implicitly modulated by the style vector $S$ in a nonlinear way, leading to the highly entanglement between $\hat{W}^{'}$ and $S$. Even though (\ref{key3}) is a linear equation, there is a non-linear relationship between $\hat{W}^{'}$ and $S$.
Based on the following evidence, we argue that the weight demodulation technique may potentially prevent the style-based generator from achieving good attribute disentanglement.
Specifically, a common goal for attribute disentanglement is a latent space consists of linear subspace \cite{Karras2019}. Intuitively, a less curved latent space should result in perceptually smoother transition than a highly curved latent space. However, a slight perturbation $[0, \cdots, 0, \delta s_n, 0, \cdots, 0]$ on the $n$-th element of $S$ would lead to huge changes on matrices $A$ and $B$, thus affecting all the entries in $\hat{W}^{'}$. This can be verified by (\ref{key3}), (\ref{key4}) and (\ref{key5}). 
To address this, a weight decomposition technique is introduced to obtain a better modulation of $W$.

\subsection{Weight Decomposition} \label{w_d}
It is clear that the weight demodulation technique implements style control via modulating the weight matrix. However, weight demodulation causes the entanglement problem between  $\hat{W}^{'}$ and $S$.
Motivated by singular value decomposition (SVD), weight decomposition (WD) technique  is proposed to construct a controllable weight matrix $\hat{W}_{WD}$ via  three matrices:
\begin{equation}\label{key6}
	\begin{split}
		\hat{W}_{WD}& =  U \cdot diag\{S\} \cdot V^{T}\\
		& = s_1 \cdot u_1 v_1^{T} + \cdots + s_i \cdot u_i v_i^{T}
	\end{split}
\end{equation}
where $\hat{W}_{WD} \in \mathbb{R}^{out \times in}$ is the constructed weight matrix for convolutional operation, $U \in \mathbb{R}^{out \times i} = [u_1, \cdots, u_i]$ and $V \in \mathbb{R}^{in \times i} = [v_1, \cdots, v_i]$ are learned matrices, and $diag\{S\} \in \mathbb{R}^{i \times i}$ is a diagonal matrix, whose diagonal entry is the style vector $S=[s_1, \cdots, s_i]$. In the generation process, $\hat{W}_{WD}$ is reshaped to $W_{WD} \in \mathbb{R}^{o \times i \times h \times w}$ for the convolutional operation.

We would like to comment on the implication of (\ref{key6}).
Firstly, $u_nv_n^T (n=1, \cdots, i)$ can be regarded as the $n$-th feature matrix corresponding to $s_n$, and $\hat{W}_{WD}$ can be regarded as the weighted composition of $u_n v_n^{T}$, where the weight is decided by the style $S$. Therefore, the convolutional operation $\hat{W}_{WD}x$  can adaptively utilize different combinations of each component of $u_nv_n^T$ to complete the forward generation process. It is clearly seen that different combination of $[s_1, \cdots, s_i]$ would produce totally different weight matrices $\hat{W_{WD}}$, thus synthesizing faces with different attributes and achieving the goal of semantic face editing.

Secondly, each $u_nv_n^T$ can be regarded as a learned feature matrix, corresponding to $s_n$. 
A slight perturbation $[0, \cdots, 0, \delta s_n, 0, \cdots, 0]$ on the $n$-th element of $S$ would only affect the weight of the $n$-th component of $u_nv_n^T$ in the forward generation process, guaranteeing the independence of each $s_i$, and therefore contributing to attribute disentanglement.

Thirdly, the formula of (\ref{key6}) may intuitively cause the misunderstanding that each entry $s_n$ corresponds to a particular attribute, and attribute editing is achieved by changing $s_n$. On the one hand, $u_nv_n^T$ is merely a feature matrix with the size of $out \times in$. Considering its capacity, $s_n$ is incapable of determining a particular attribute via controlling $u_nv_n^T$. Through the ablation study in Section \ref{ap1}, it is clear that  multiple $s_n$ across different layers (instead of only one $s_n$) would be changed when editing one particular attribute.
On the other hand, establishing the correspondence between $s_n$ and one particular attribute is sub-optimal for learning a disentangled representation. To explain this, we show a specific and illustrative example. Suppose that each entry in the style vector corresponds to one particular vector, and let $s_m$ and $s_k$ denote the attribute of age and hair color, respectively. 
Conducting face aging would potentially turn the hair color to gray or silver, which in turn affect the attribute of hair color. In the case above, $s_m$ would be entangled with $s_k$ as they both control the hair color. Such a case would obviously prevent the model from learning a disentangled representation.
Considering this, the style $S$ has a close relationship to matrix-related features, \textit{i.e.}, $u_nv_n^T$, instead of the facial attributes.
Actually, the latent code $z$ and $w$ are the vectors that determine the facial attributes. Because the style vectors across different layers are all produced by $w$, which controls the entire synthesis process and therefore determines the facial attributes.

Weight decomposition of (\ref{key6}) has a similar formula to SVD. Similar techniques like singular value clipping \cite{Saito2017} and singular value bounding \cite{Jia2017} have been utilized in regularizing the singular values. The differences between (\ref{key6}) and SVD include that $U$ and $V$ in the proposed weight decomposition are not the singular vectors of $\hat{W}$, and $S$ has no relationship with the singular values of $\hat{W}$ as well. We just utilize such a similar formula to construct a controllable matrix to implement style control.

\subsection{Orthogonal Regularization}
When utilizing (\ref{key6}) to construct the weight matrix, a special case may exist where the feature matrix $u_nv_n^T$ can be obtained by another two feature matrices $u_mv_m^T$ and $u_kv_k^T$:
\begin{equation}\label{key61}
	u_nv_n^T= \beta_1 \cdot u_mv_m^T + \beta_2 \cdot  u_kv_k^T
\end{equation}
where $\beta_1$ and $\beta_2$ are the coefficients.
Suppose that $S^1=[s_1, \cdots, s_n, \cdots, s_m, \cdots, s_k, \cdots, s_i]$, $S^2=[s_1, \cdots, 0, \cdots, s_m+\beta_1 \cdot s_n, \cdots, s_k+\beta_2 \cdot s_n, \cdots, s_i]$, and $S^1$, $S^2$ are produced by $w^1$ and $w^2$, respectively.
It is obvious that $U\cdot diag\{S^1\} \cdot V^T$ = $U\cdot diag\{S^2\} \cdot V^T$.
Even though $w^1$ is different from $w^2$ in the latent space, $w^1$ and $w^2$ can produce identical weight matrix to control the synthesis process. Obviously, such a case would prevent the model from learning a disentangled representation in the latent space.
To avoid such a case and encourage the independence of the feature matrices, orthogonal regularization \cite{Brock2016,Miyato2018}  can be applied in the adversarial objective function of the generator by adding the following term:
\begin{equation}\label{key7}
	\alpha \cdot(||U^TU - I|| + ||V^TV -  I||)
\end{equation}
where $\alpha$ is a hyper-parameter, and empirically taken as 1 in the following experiments. $I$ is the identity matrix. 
Applying orthogonal regularization, $u_n$, $v_n$ in $U$ and $V$ would satisfy:
\begin{equation}\label{key8}
	u_n\cdot u_m ( v_n \cdot v_m )=\begin{cases}
		1& m = n \\ 
		0& m \neq n
	\end{cases}
\end{equation}

\begin{figure*}[htp]
	\centering
	\begin{subfigure}[b]{0.32\textwidth}
		\centering
		\includegraphics[height=0.6\linewidth]{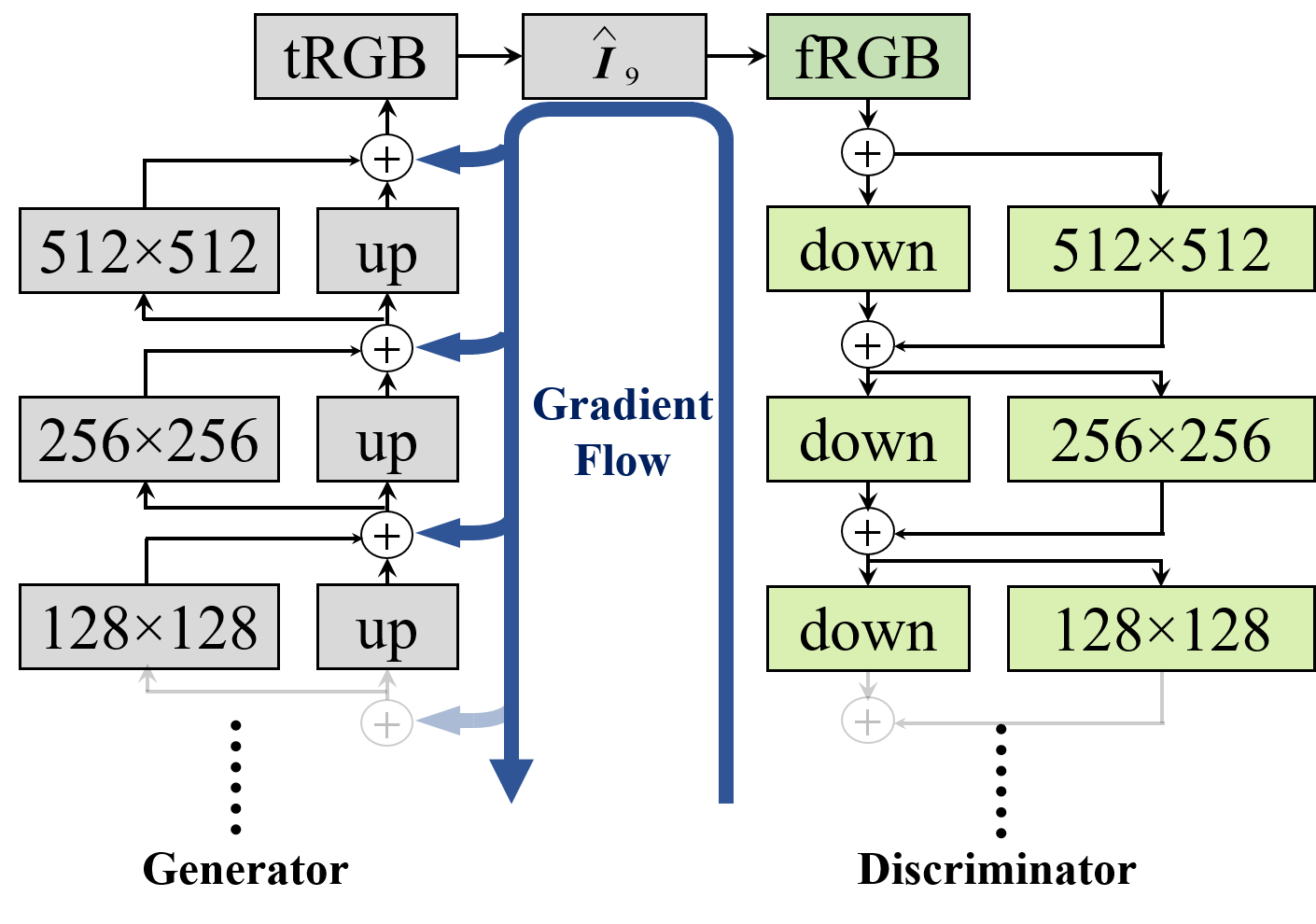}	
		\caption{StyleGAN}
	\end{subfigure}	
	\begin{subfigure}[b]{0.32\textwidth}
		\centering
		\includegraphics[height=0.6\linewidth]{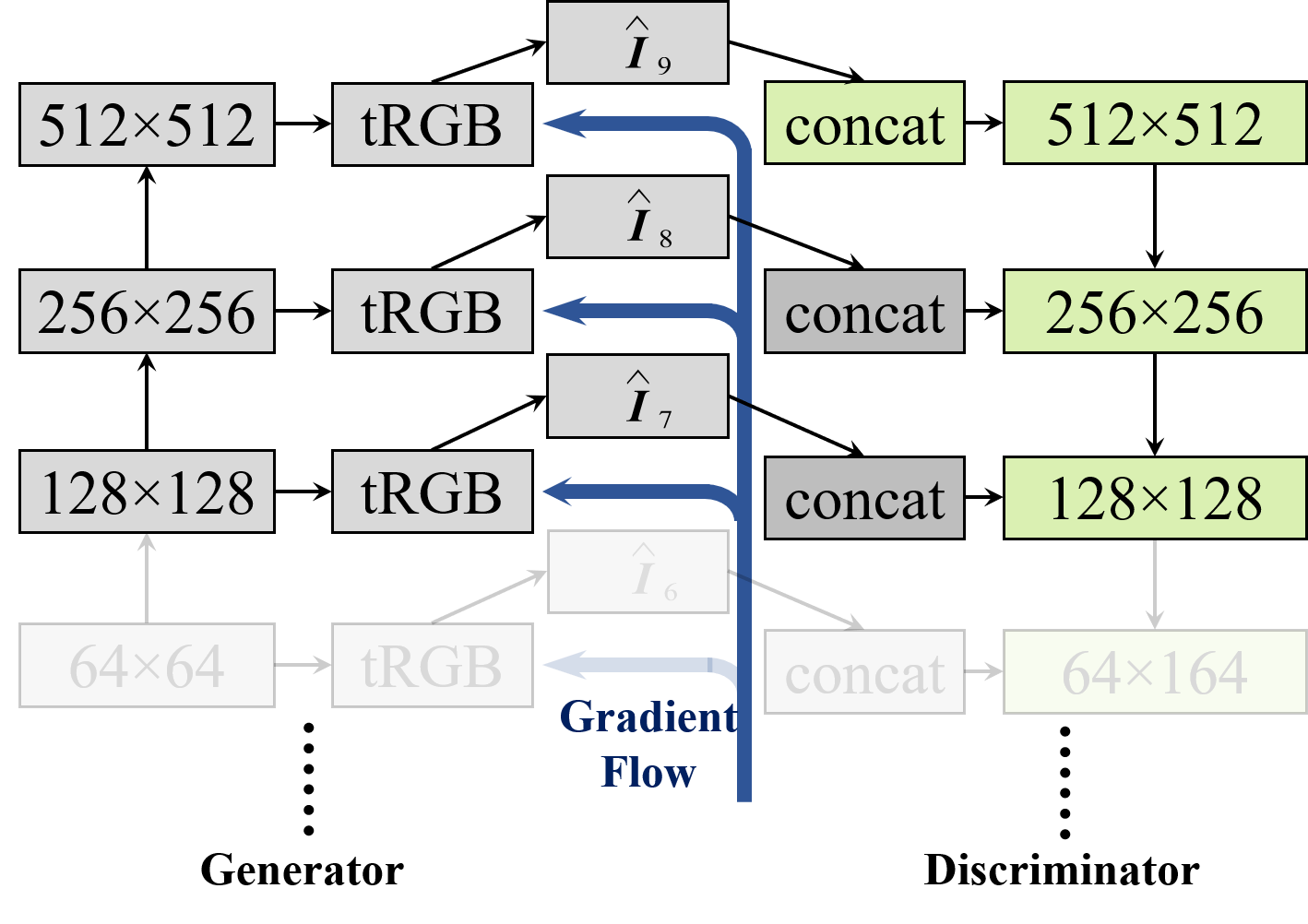}	
		\caption{MSG-GAN}
	\end{subfigure}	
	\begin{subfigure}[b]{0.32\textwidth}
		\centering
		\includegraphics[height=0.6\linewidth]{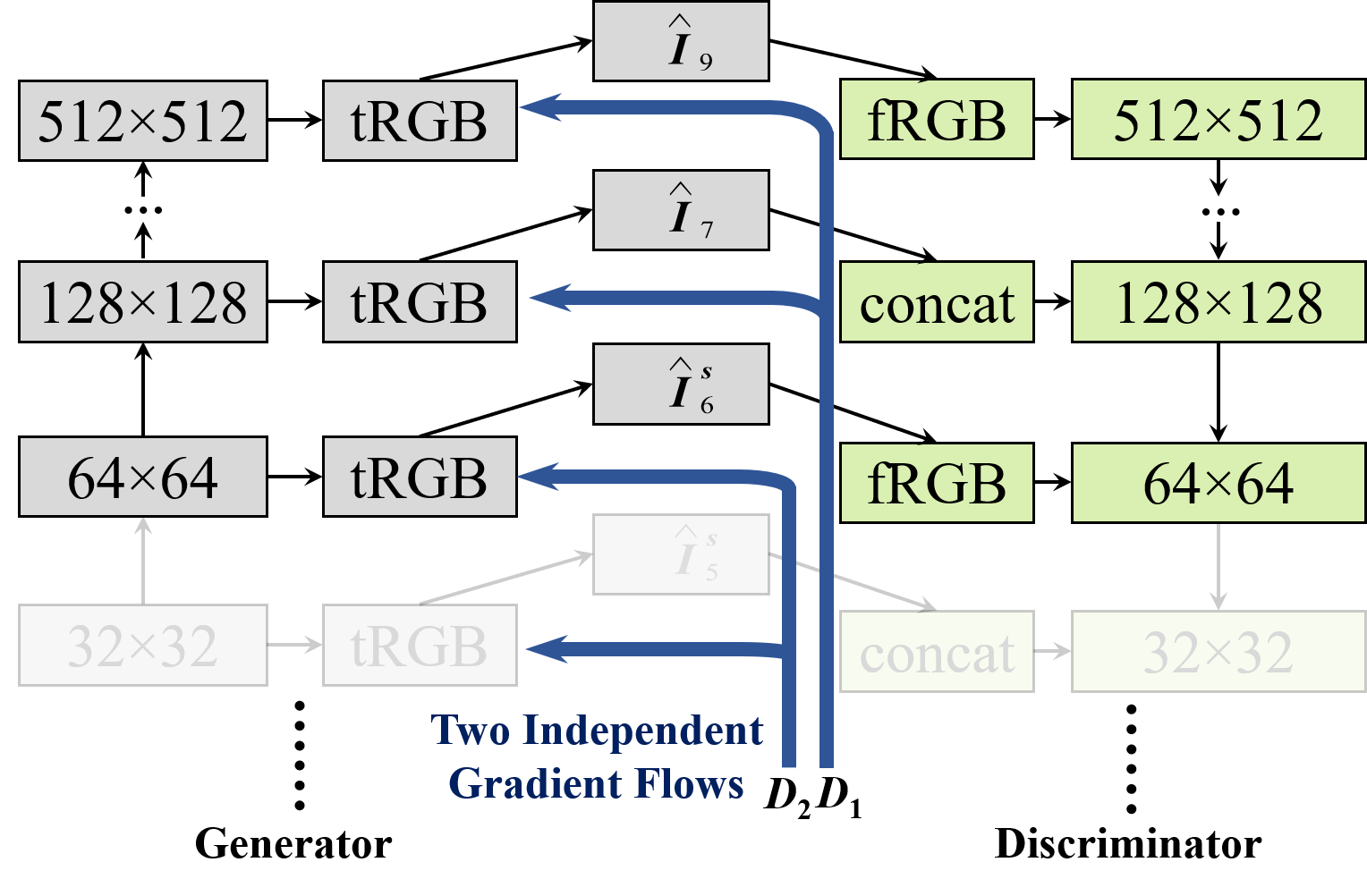}	
		\caption{Ours}
	\end{subfigure}	
	\caption{ Comparison of the gradient flows in the backpropagation process. (a) StyleGAN \cite{Karras2020}. (b) MSG-GAN \cite{Karnewar2020}. (c) Ours. $\hat{I}_{res}$ represents the generated images with resolution of $2^{res} \times 2^{res}$ pixels, ${\boxed{\textrm{$res \times res$}}}$ represents the convolutional block with size of $res \times res$, the ${\boxed{\textrm{tRGB}}}$ block represents a layer that projects feature tensor to RGB colors and ${\boxed{\textrm{fRGB}}}$ block does the reverse,  ${\boxed{\textrm{concat}}}$ is the concatenation operation}
	\label{fig:fig3}
\end{figure*}

It is clearly seen that orthogonal regularization can guarantee the independence of the feature matrix $u_nv^T_n$. Thus, applying the orthogonal regularization to constrain $U$ and $V$ in (\ref{key6}) can further ensure that each entry $s_n$  control one independent feature matrix.

Orthogonal regularization has been used to enforce the weights in the discriminator to be orthogonal \cite{Brock2016,Miyato2018} , thus guaranteeing the discriminator satisfy the Lipschitz constraint \cite{Arjovsky2017} for improving the training stability \cite{Miyato2018}. As a comparison, orthogonal regularization of (\ref{key7}) is used in the generator to constrain the feature matrices for the purpose of controlling the synthesis process.
Proof of validity of weight decomposition and orthogonal regularization is conducted in Section \ref{Exp}.

\begin{figure}[t]
	\centering
	\begin{subfigure}[b]{0.15\textwidth}
		\centering
		\includegraphics[height=0.8\linewidth]{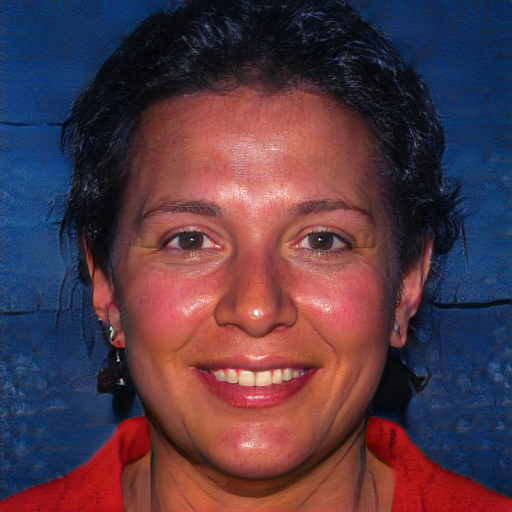}\\
		\includegraphics[height=0.8\linewidth]{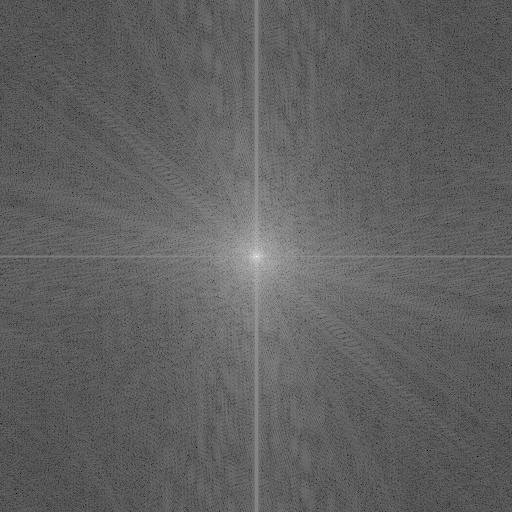}
		\caption{RGB}
	\end{subfigure}	
	\begin{subfigure}[b]{0.15\textwidth}
		\centering
		\includegraphics[height=0.8\linewidth]{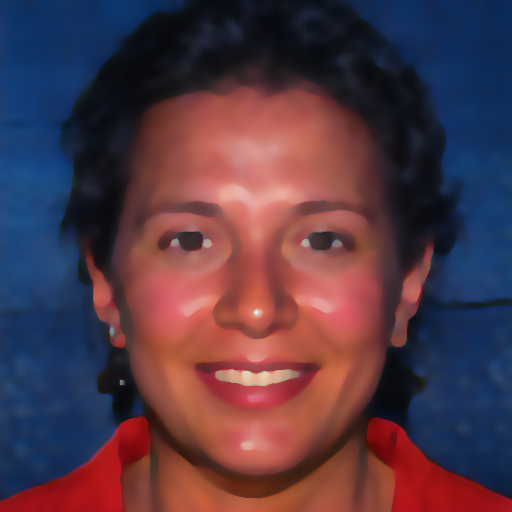}\\
		\includegraphics[height=0.8\linewidth]{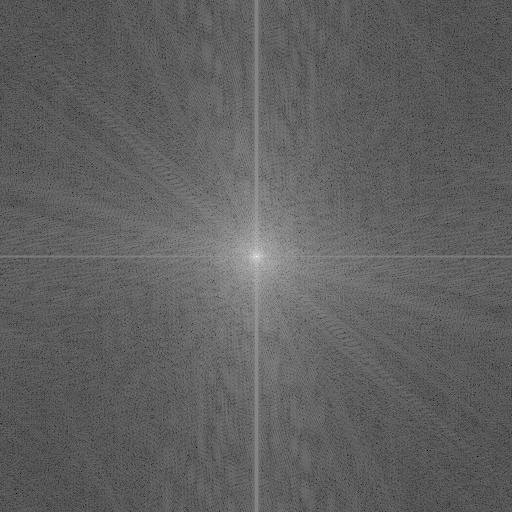}
		\caption{structure}
	\end{subfigure}	
	\begin{subfigure}[b]{0.15\textwidth}
		\centering
		\includegraphics[height=0.8\linewidth]{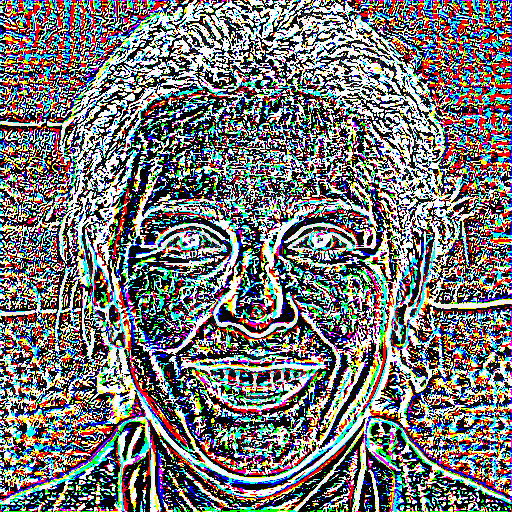}\\
		\includegraphics[height=0.8\linewidth]{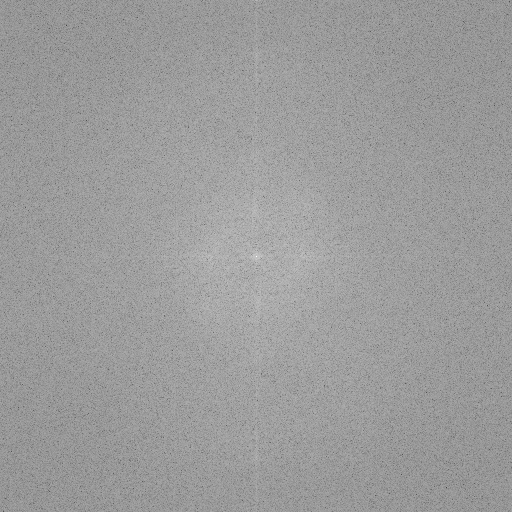}	
		\caption{texture}
	\end{subfigure}	
	\caption{Illustration of a face image as well as its corresponding structure, texture components (upper row) and the magnitude of their frequency spectra (lower row). (a) shows an example of a human face. (b) and (c) show the structure and texture components, respectively. The structure and  texture components are obtained by the structure-texture decomposition algorithm \cite{Xu2012}.}
	\label{fig:fig4}
\end{figure}

\section{Architecture}
The GAN architecture is essential for synthesizing images of high fidelity \cite{Karras2017}.
To obtain better attribute editing in the generation process, we have modified the multi-scale gradient  technique, which is used for stable training \cite{Karnewar2020}, and applied it to the proposed structure-texture independent architecture, which obtains better perceptual results on attribute disentanglement.
To clearly introduce our motivations, we first recap some essential strategies which have been used in the GANs training and deep learning community, then show our proposed GAN architecture.

\subsection{Multi-scale Gradient}
GAN training is dynamic and sensitive to nearly every aspect of its setup, from optimization parameters to model architecture. Training instability, or mode collapse, is one of the major obstacles in developing applications \cite{Liu2019}. It has been found that one of the reasons for the training instability of GANs is due to the passage of random (uninformative) gradients from the discriminator to the generator when there is insubstantial overlap between the supports of the real and fake distributions \cite{Karnewar2020}. 

\begin{figure*}[htp]
	\centering
	\begin{subfigure}[b]{0.45\textwidth}
		\centering
		\includegraphics[height=0.4\linewidth]{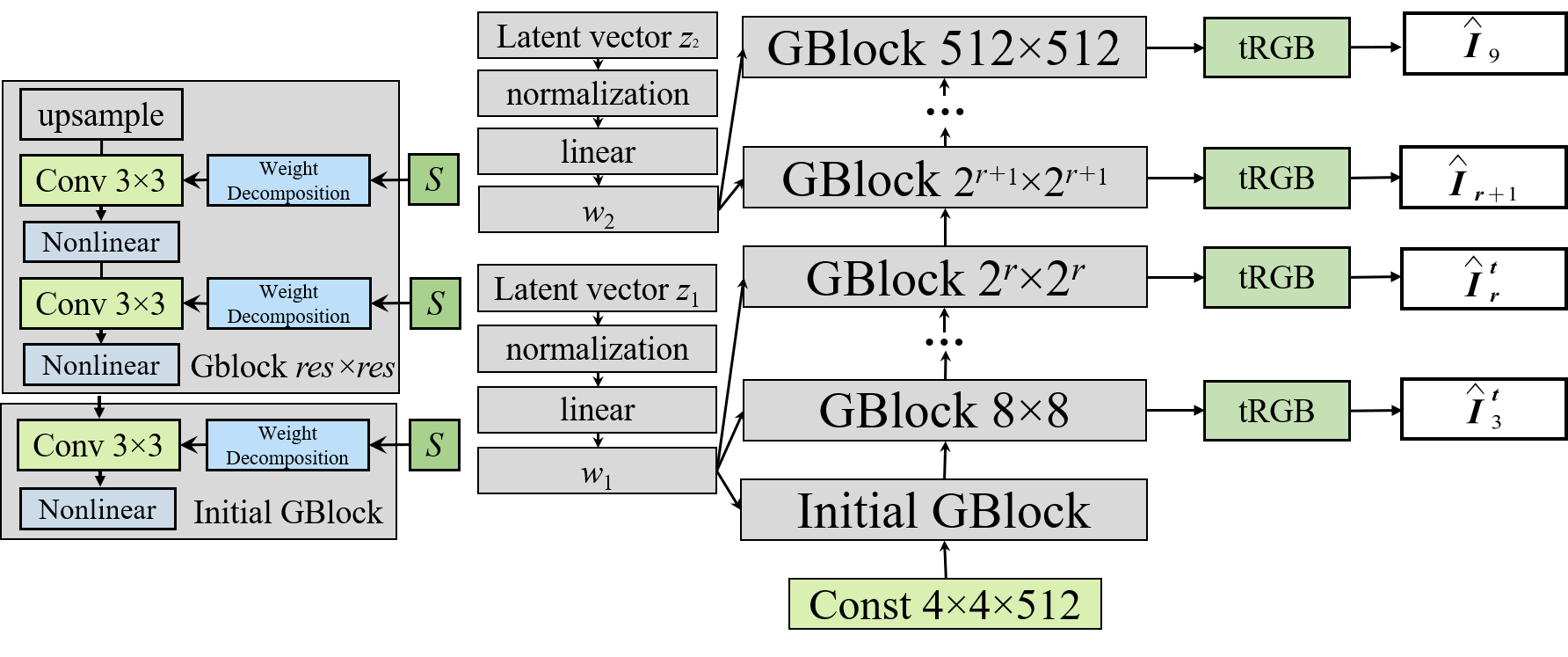}	
		\caption{Generator}
	\end{subfigure}	
	\begin{subfigure}[b]{0.45\textwidth}
		\centering
		\includegraphics[height=0.4\linewidth]{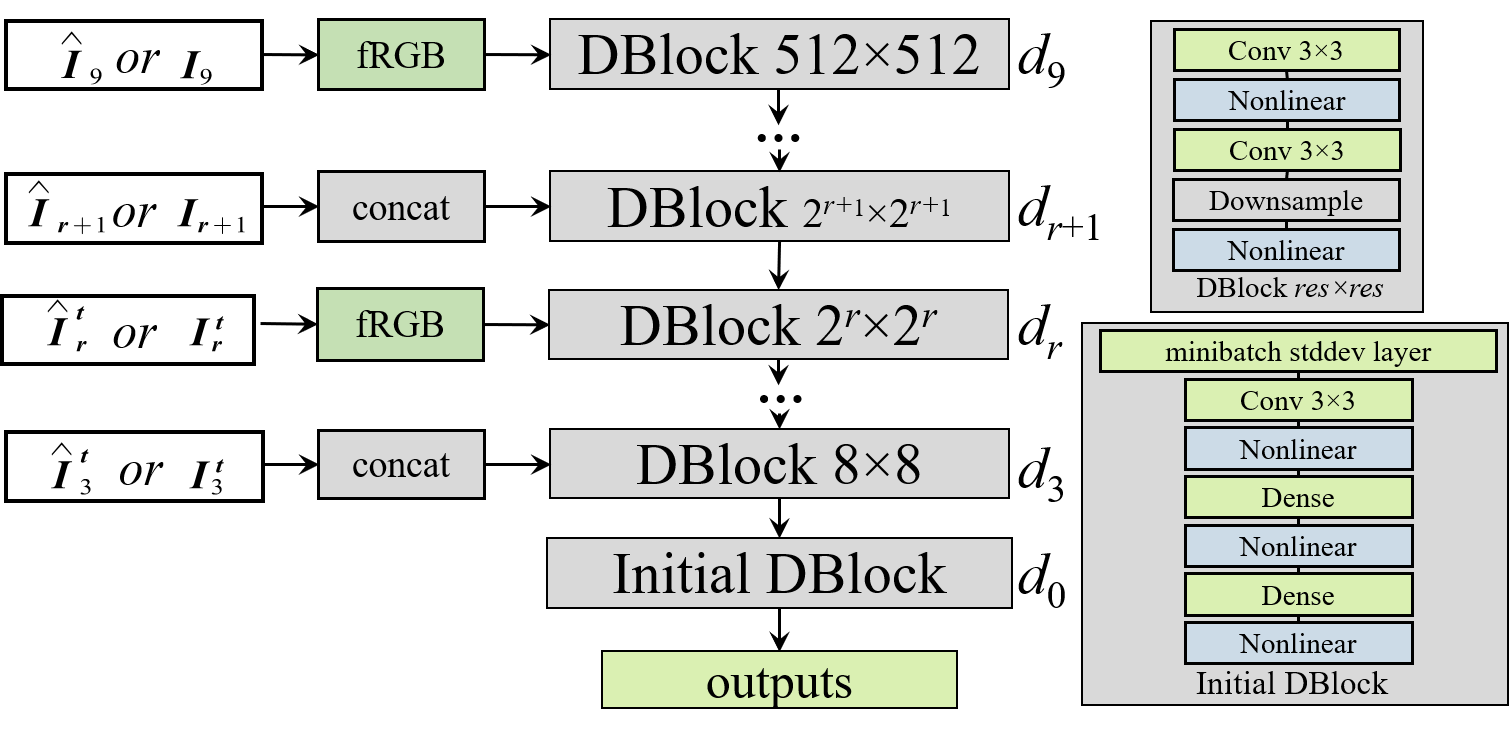}	
		\caption{Discriminator}
	\end{subfigure}	
	
	\caption{Proposed architecture. (a) the generator architecture, where $z_1 \in R^{512}$ and $z_2 \in R^{512}$ are the latent vectors with independent and identical distribution,  Initial Gblock and Gblock $res \times res$ ($res$ represent the resolution) are the convolutional blocks, whose details are shown on the left. ${\boxed{\textrm{tRGB}}}$ represents a layer that projects feature vectors to RGB colors or structure component, and $\hat{I}_{res}$, $\hat{I}^{s}_{res}$  are the outputs at corresponding resolutions. (b) the discriminator architecture, where ${\boxed{\textrm{fRGB}}}$ represents a layer that projects RGB colors or the structure component to feature vectors, ${\boxed{\textrm{concat}}}$ is the  concatenation operation, Initial DBlock and DBlock $res \times res$ are the convolutional Block, whose details are shown on the right. }
	\label{fig:fig5}	
\end{figure*}

The progressive growing strategy \cite{Karras2017} is proposed to tackle the training instability problem by gradually training the GAN layer-by-layer. Thus, the generator will initially focus on low-resolution features and then slowly shift its attention to finer details, achieving a good distribution match on lower resolutions, therefore contributing to alleviating the support overlap problem. 
The progressive growing strategy has been demonstrated to be successful in stabilizing high-resolution image synthesis. However, the  progressively grown generator  appears to have a strong location preference for details, causing charateristic artifacts in the synthesized images \cite{Karras2020}.  
In searching for a training strategy as a replacement of progressive growing, MSG-GAN provides the multi-scale gradient (MSG) technique to allow the gradients from the discriminator flow back to the generator at multiple scales \cite{Karnewar2020}.
The generator of MSG-GAN outputs images of multiple resolutions instead of an image, which are then concatenated at the corresponding scales of the discriminator, providing a pathway for gradient flows at multiple scales.  The MSG technique allows GANs to focus on low-resolution features, tackling the training instability problem in a similar way to the progressive growing technique. In the subsequent work, the improved version of StyleGAN \cite{Karras2020} has simplified this design by upsampling and summing the contributions of image outputs corresponding to different resolutions, which similarly allows the gradient flow at multiple scales of the generator. Fig. \ref{fig:fig3} plots their gradient flows in the backpropagation process of the training. To obtain stable training, we utilize the multi-scale gradient strategy and make some modifications to take the structure and texture components into consideration when designing the architecture (see Section \ref{arch}).

\subsection{Structure Vs Texture}
Convolutional Neural Networks (CNNs) are structure or texture sensitive. In the deep learning community, attempts have been made towards explaining the impressive performance of CNNs on complex perceptual tasks like object detection. One widely accepted explanation is the shape hypothesis, which holds the opinion that CNNs combine low-level features (\textit{e.g.} edges) to increasingly complex shapes in the decision process \cite{Zeiler2014_eccv,Kubilius2016,Ritter2016}. 
On the other hand, the texture hypothesis suggests that CNNs can still classify  textured images perfectly even if the global shape structure is completely destroyed \cite{Hosseini2018,Gatys2017,Brendel2019}.
From the debate between the shape and texture hypothesis, it is clear that CNNs are structure or texture sensitive.

Besides, facial attributes have a close relationship with the structure and texture components, which have different frequency properties.
As shown in Fig. \ref{fig:fig4}, the texture component, more likely related to the high frequency component, outlines the shape of human faces. As a comparison, the structure component, having intense responses in the low frequency range, indicates the attributes like face or hair color, illumination, \textit{etc}.
In addition, when conducting image generation task via GANs, some prior works have tried to consider different components, \textit{e.g.}, $S^2$-GAN \cite{Wang2016} utilized a structure GAN and a style GAN to break the difficult generative task into sub-problems, and LR-GAN \cite{Yang2017} factorized image generation into foreground and background generation with layered recursive GANs. 

To summarize, CNNs as well as facial attributes are both structure and texture sensitive. Previous studies have motivated us that generating the structure and texture separately may potentially contribute to better attribute editing. Thus, we introduce the structure-texture independent architecture to hierarchically synthesize the texture and structure parts independently, where modified multi-scale gradient strategy is used.

\subsection{Proposed Architecture} \label{arch}
According to the analysis above, we would like to utilize the multi-scale gradient strategy to guarantee stable training, and generate the structure and texture components separately for better attribute disentanglement. 

Let $I_{res}$ denote the images with the resolution of $2^{res} \times 2^{res}$ pixels, and $\hat{I}$ represents the generated image. 
$I^{s}$ and $I^{t}$ are the structure and texture components of image $I$, respectively.
Fig. \ref{fig:fig5} shows the proposed architecture, where the generator starts from a learned constant, identical to that in StyleGAN. However, the 8-layer mapping network in StyleGAN is removed because of its very limited effect on image generation and attribute disentanglement. As a replacement, the linear transformation and normalization operation is used to obtain the intermediate latent code $w$, and the learned affine transformations then specialize $w$ to style vector $S$, which controls the image synthesis via the proposed weight decomposition.
Similar to MSG-GAN, the generator would synthesize images of multiple resolutions, which enable gradient flow at multiple scales. Particular differences include that the coarse layer would output the corresponding texture component instead of the RGB images for the purpose of encouraging the separation of the texture and structure parts. To achieve this, two independently and identically distributed latent vectors $z_1$ and $z_2$ are used to control the synthesis of the generator. To be specific, $z_1$ is responsible for controlling the coarse layers, which output the texture components at corresponding resolutions. The sizes of the coarse layer are lower than $2^{r+1} \times 2^{r+1}$, where $r$ is a hyper-parameter. 
Meanwhile,  $z_2$ is utilized to control the fine layers to generate RGB images.
It is clearly seen that the generator would produce the texture component first, then synthesize its corresponding structure part to obtain the final RGB output.
The reason for generating the texture component first is that the texture component is more likely related to low-level features \cite{Karras2020}, and it has intense response in the high frequency domain, which is more difficult to generate.
Equivalently, the latent vector $z_1$ can be regarded as controlling the texture component, while $z_2$ controls the structure component. As shown in Fig. \ref{fig:fig4}, the texture component mainly outlines the human faces. Thus, changing $z_1$ would affect attributes closely related to the face outline, \textit{e.g.}, hair style, expression, \textit{etc}.
As a comparison, $z_2$ would determine attributes like face color, hair color, \textit{etc}. Hence, structure related attributes can be changed while maintaining texture related ones, or vice verse, which indeed contributes to attribute disentanglement.

Applying the multi-scale gradient strategy, the discriminators of MSG-GAN and StyleGAN judge the real samples from the fake ones according to the joint distribution $p(I_n, I_{n-1}, \cdots, I_3)$ instead of $p(I_n)$, thus allowing the discriminator to focus on low-resolution features as well as the high-resolution ones.
The proposed generator in Fig. \ref{fig:fig5} (a) would synthesize the RGB colors in the fine layers, \textit{i.e.}, $\hat{I}_{n} \cdots \hat{I}_{r+1}$, which is based on the texture components, \textit{i.e.}, $\hat{I}^{t}_{r} \cdots \hat{I}^{t}_{3}$.
In a similar way, the discriminator is supposed to distinguish the real samples from the fake ones based on the joint distribution $p(I_n,\cdots,I_{r+1})$.
Furthermore, to encourage the generator to synthesize the structure and texture parts in a disentangled way, we would like the discriminator to judge the texture part separately. 
To achieve this, we can construct two  independent gradient flows in the backpropagation process, one for the RGB images and the other for the texture component. 
Thus, the output of the discriminator $D$ can be expressed as:
\begin{equation}\label{key9}
	\begin{split}
		&D(I_n,\cdots, I_{r+1},I^{t}_{r},\cdots, I^{t}_3) \\
		=&D_1(I_n,\cdots, I_{r+1}) + D_2(I^{t}_{r}, \cdots , I^{t}_3)\\
	\end{split}
\end{equation}
where $D_1$ and $D_2$ are some parametric functions.
Interpretation of (\ref{key9}) can be given as distinguishing the real samples from the fake ones can be accomplished by two independent parts: one judges the RGB images with fine resolutions, while the other judges the texture components with coarse resolutions. $D_1$ and $D_2$ are implemented as followed:
\begin{equation}
	\begin{split}
		&D_1(I_n,\cdots, I_{r+1}) = \\
		&d_0(d_3(\cdots(d_{r+1}(([d_{n-1}([d_n(f(I_n)), I_{n-1}])\cdots), I_{r+1}]))))\\
	\end{split}
\end{equation}
\begin{equation}
	\begin{split}
		&D_2(I^{t}_{r},\cdots, I^{t}_3) =\\
		&d_0(d_3([\cdots( d_{r-1} ([d_r(f(I^t_r)), I^t_{r-1}]), \cdots), I^t_3 ]))
	\end{split}
\end{equation}
where $[ , ]$ represents the concatenation operation, $d_0, d_3, \cdots d_n$ are the convolutional blocks in the discriminator as shown in Fig. \ref{fig:fig5} (b). To calculate $D_1$ or $D_2$, $I_n$ or $I^{t}_r$ is projected to the feature tensors at corresponding scales via the ${\boxed{\textrm{fRGB}}}$ block $f$, while images from lower resolutions are concatenated to the corresponding layers, which has been demonstrated to be the most effective way in MSG-GAN \cite{Karnewar2020}.
$D_1$ and $D_2$ are calculated independently but share the coarse blocks \textit{i.e.}, $d_0$, $d_3, \cdots, d_r$. Hence, only one model is used to implement the discriminator instead of two.
As a comparison, the discriminator output in MSG-GAN and StyleGAN can be expressed as:
\begin{equation}
	\begin{split}
		D(I_n, &\cdots, I_3) \\
		&=d_0(d_3([\cdots d_{n-1}([d_n(I_n),I_{n-1}]), \cdots),I_3]))
	\end{split}
\end{equation}

To better show the differences, we plot the gradient flows in Fig. \ref{fig:fig3}. It is clearly seen that, in the back propagation process, $D_2$ would pass one independent gradient flow back to the generator, thus allowing the discriminator to emphasize on the low level features, which are closely related to the texture component. 

To sum up, the distinguishing feature of the proposed structure-texture independent architecture is the independent generation of structure and texture components.
In Section \ref{Exp}, we will conduct  experiments to verify the validity of the proposed architecture on face editing.

\section{Unsupervised Attribute Editing}
%

The goal of STGAN-WO is towards learning a disentangled space that consists of linear subspaces.	Once a disentangled representation in the latent space is learned by STGAN-WO, moving $w$ along its orthogonal directions would produce new attributes which are absent in the original face \cite{Karras2019,Karras2020}.	
As stated in Section \ref*{w_d}, each entry $s_n$ in the style vector is incapable of determining one particular attribute, and editing one particular attribute would inevitably affect multiple entries across different style vectors (see the example in Section \ref{ap1}). 
Thus, it is impossible to find a disentangled representation in the space of $S$.
As a comparison, STGAN-WO is more likely to learn a less disentangled representation in the latent space of $w$ and $z$ than that of $S$. The reason behind is stated as followed.
$u_nv_n^T$ can be regarded as an independent feature matrix, and the latent code $w$ would accordingly choose  suitable ones via the produced $S$ for accomplishing the face synthesis task. Orthogonal vectors in the $w$ space can still share those feature matrices to complete similar operations, thus making it easier to learn a disentangled representation in $w$ space than that in $S$ space.
Recalling the example in Section \ref*{w_d} that face aging would potentially turn the hair color to gray or silver, which in turn affect the attribute of hair color.
Suppose that $w_m$ and $w_k$ represent the attributes of age and hair color respectively, and the style vectors $S_m$ and $S_k$ are produced by $w_m$ and $w_k$, respectively.
Even though $w_m$ and $w_k$ are supposed to be orthogonal vectors in a disentangled space, $S_m$ and $S_k$ can both utilize the component of $u_nv_n^T$ to complete their common task, \textit{i.e.}, changing hair color, avoiding the entanglement problem when establishing the correspondence between the attribute and the entry of $S$ (see Section \ref*{w_d}). 

As we can see, STGAN-WO can learn a less disentangled representation in the $w$ space with the help of the weight decomposition and orthogonal regularization techniques.
Considering this and to obtain a more precise control of facial attributes, we can move $w$ along its orthogonal directions to obtain the new intermediate latent codes, and  utilize the new latent codes to generate new faces, whose attributes will have been changed when compared with the original face images. 
Manipulating $w$ instead of $z$
is based on the evidence that $w \in \mathbb{W}$ is less entangled than
$z \in \mathbb{Z}$ \cite{Karras2020,Karras2019}.
Taking $w_1$ as an example, moving $w_1$ along its orthogonal directions can have:
\begin{equation}\label{key15}
	w^{'}_1 = w_1 + \alpha \cdot w^{\perp}_{1}
\end{equation}
where $w^{'}_1$ is the new latent code, $ w^{\perp}_{1}$ is the orthonormal vector of $w_1$, and $\alpha$ is a coefficient. In the following sections, we would show how (\ref{key15}) affects the facial attribute.

%
%
%

\begin{figure*}[htp]
	\centering
	\begin{subfigure}[b]{0.13\textwidth}
		\centering
		\includegraphics[width=1.0\linewidth]{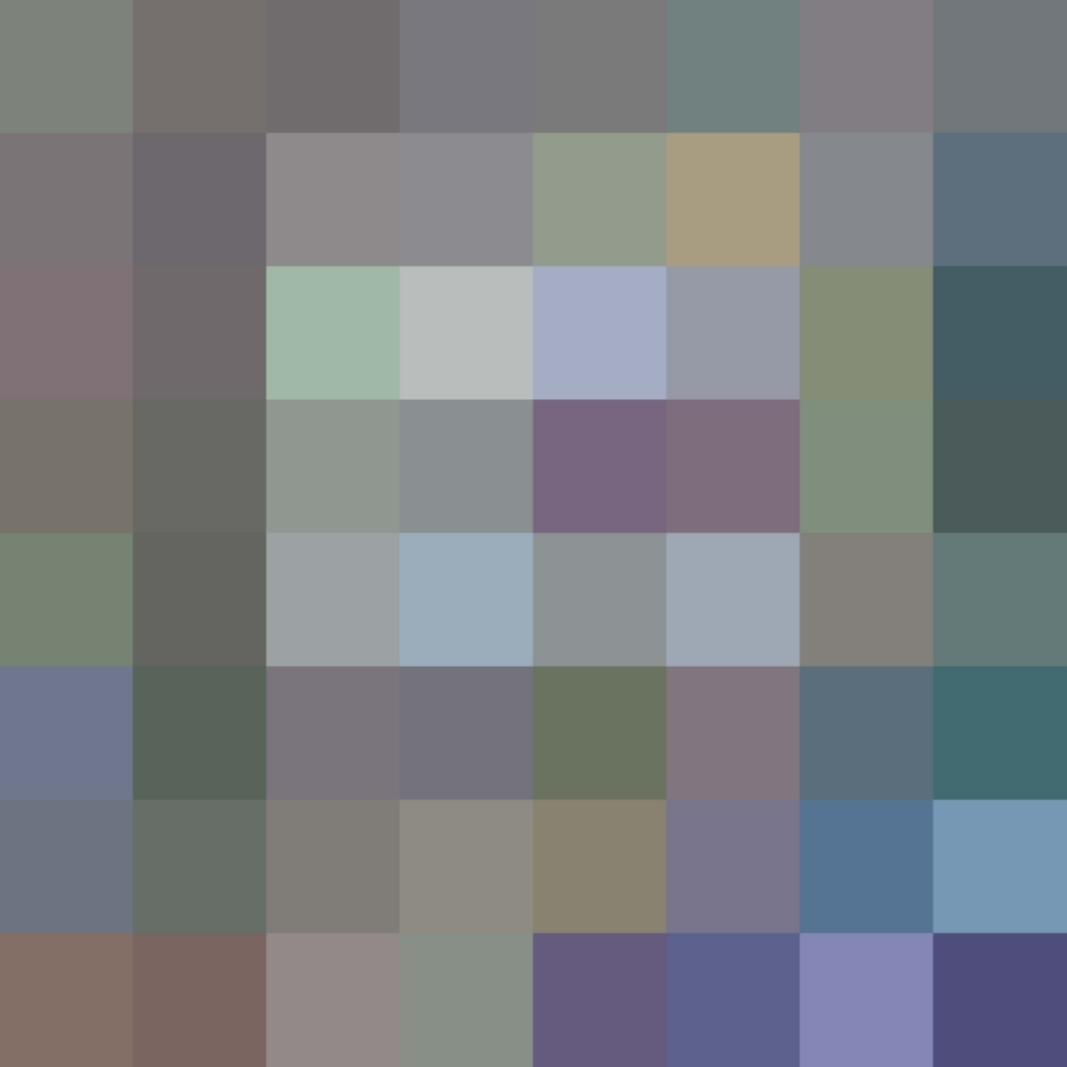}\\
		\includegraphics[width=1.0\linewidth]{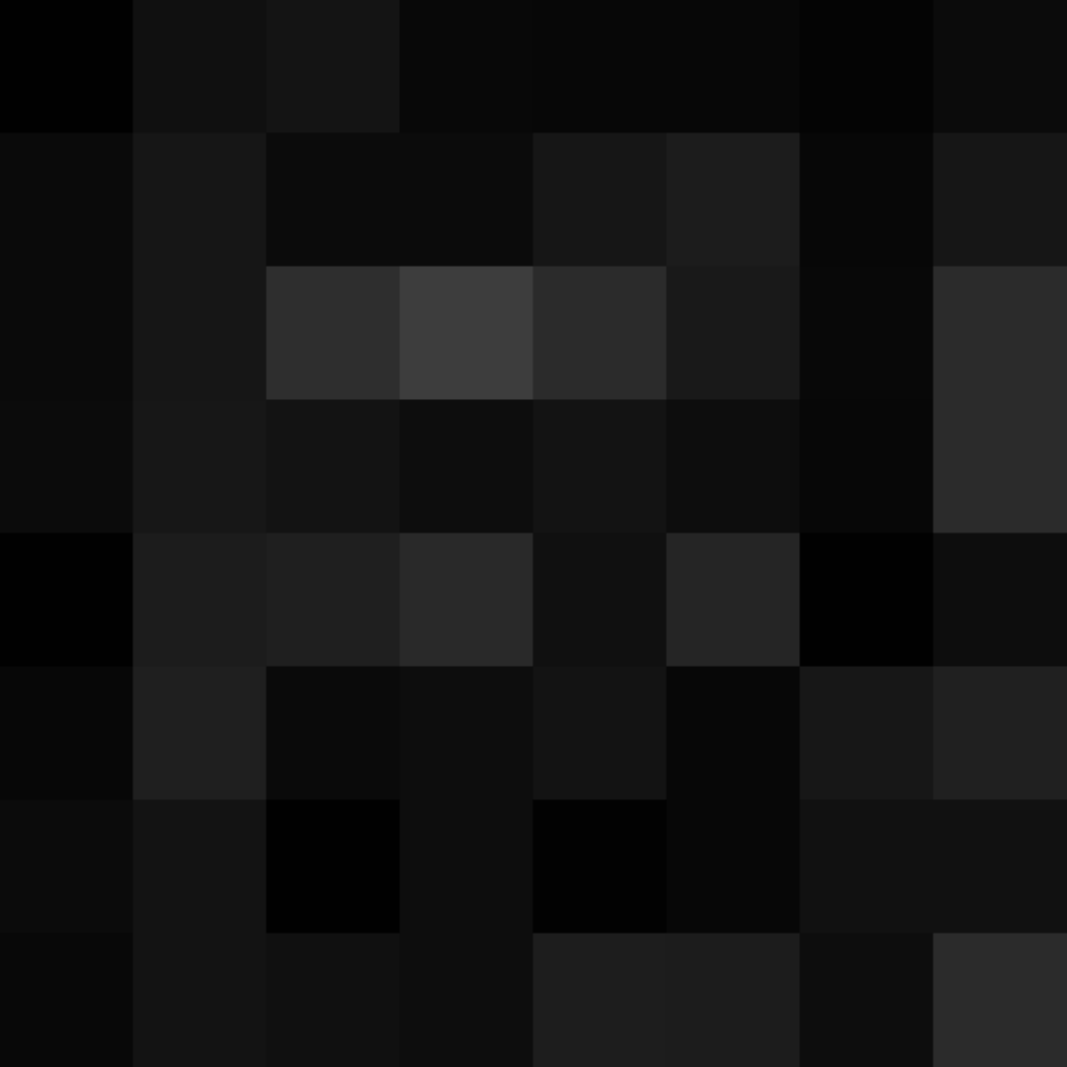}\\
		\includegraphics[width=1.0\linewidth]{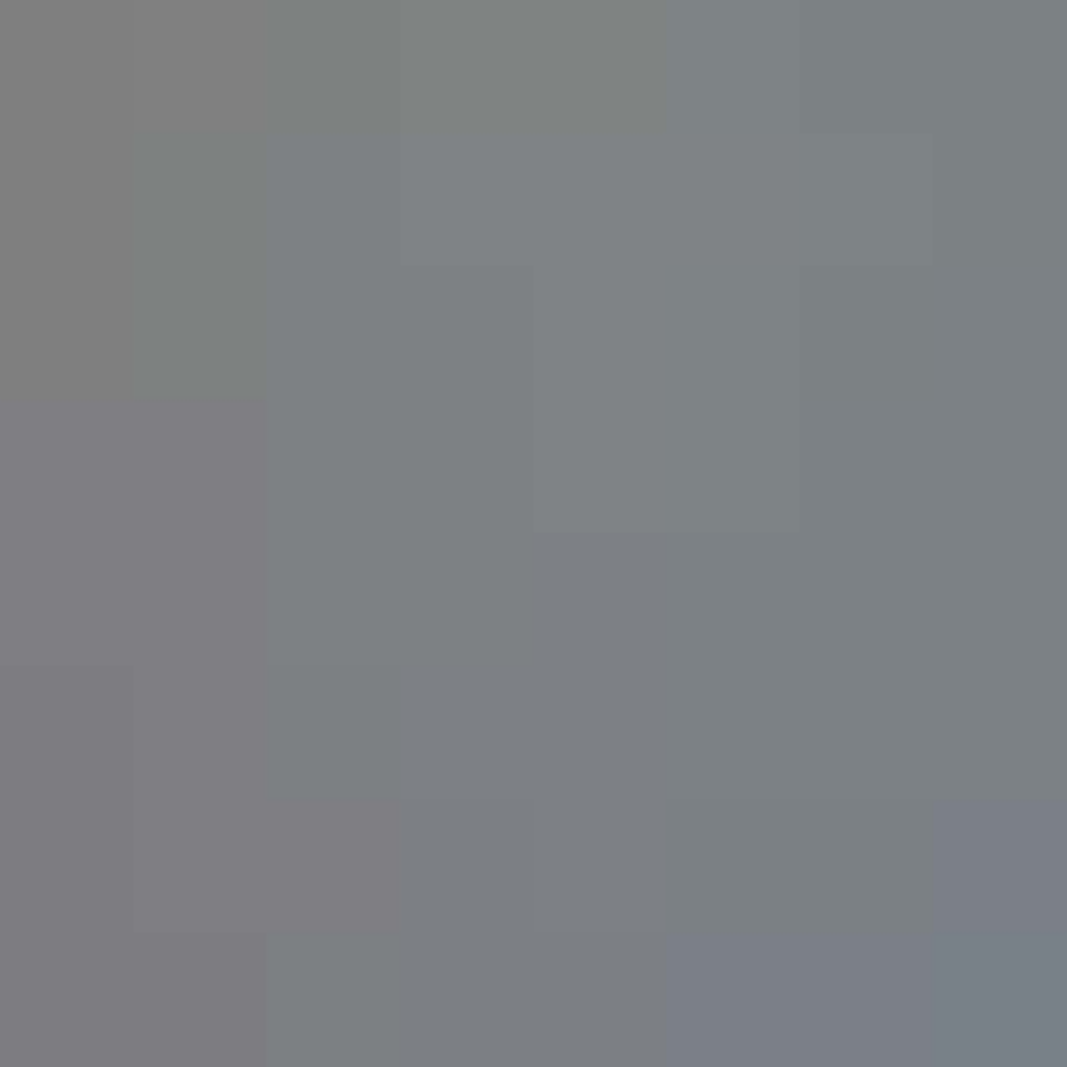}
		\caption{$\hat{I_3^t}$}
	\end{subfigure}	
	\begin{subfigure}[b]{0.13\textwidth}
		\centering
		\includegraphics[width=1.0\linewidth]{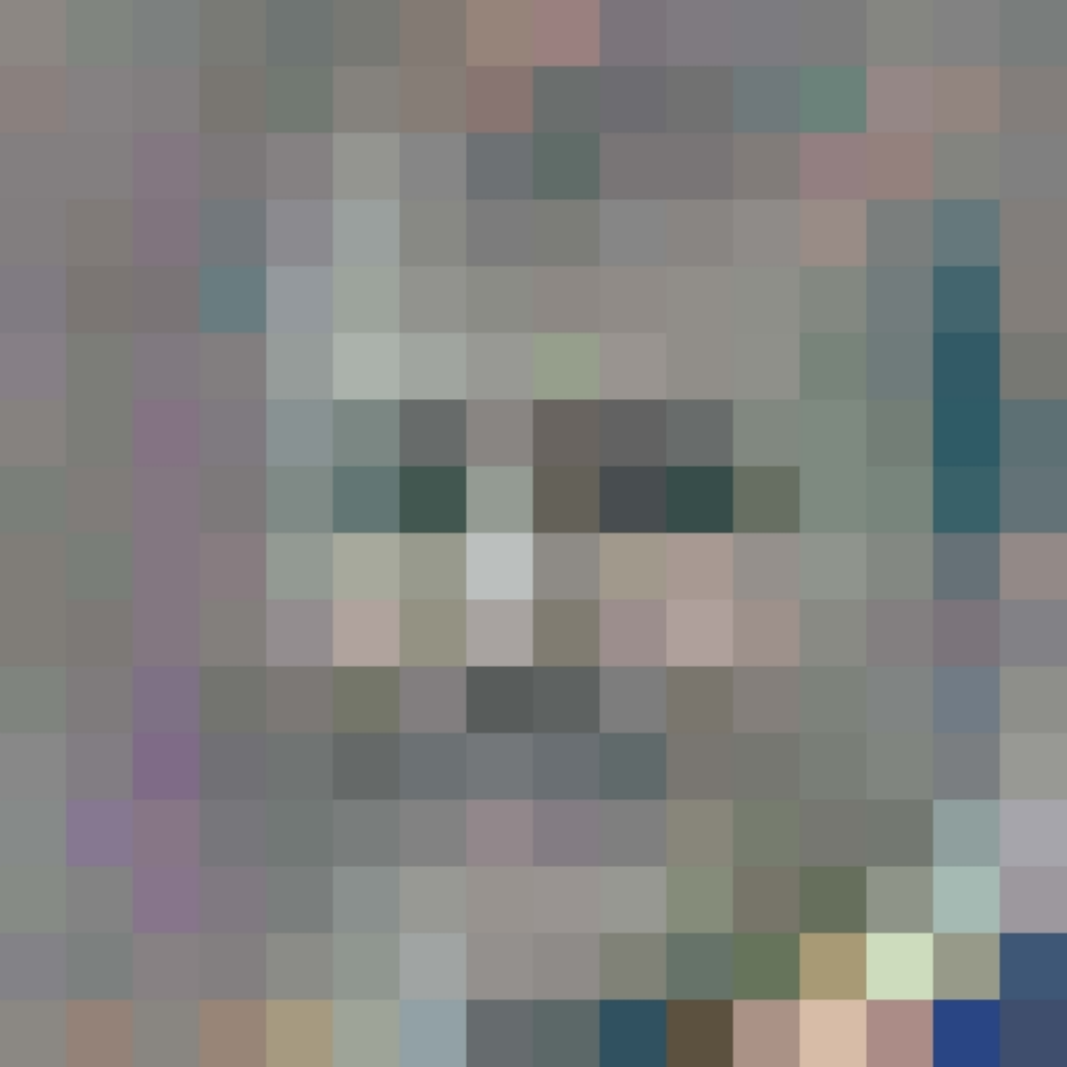}\\
		\includegraphics[width=1.0\linewidth]{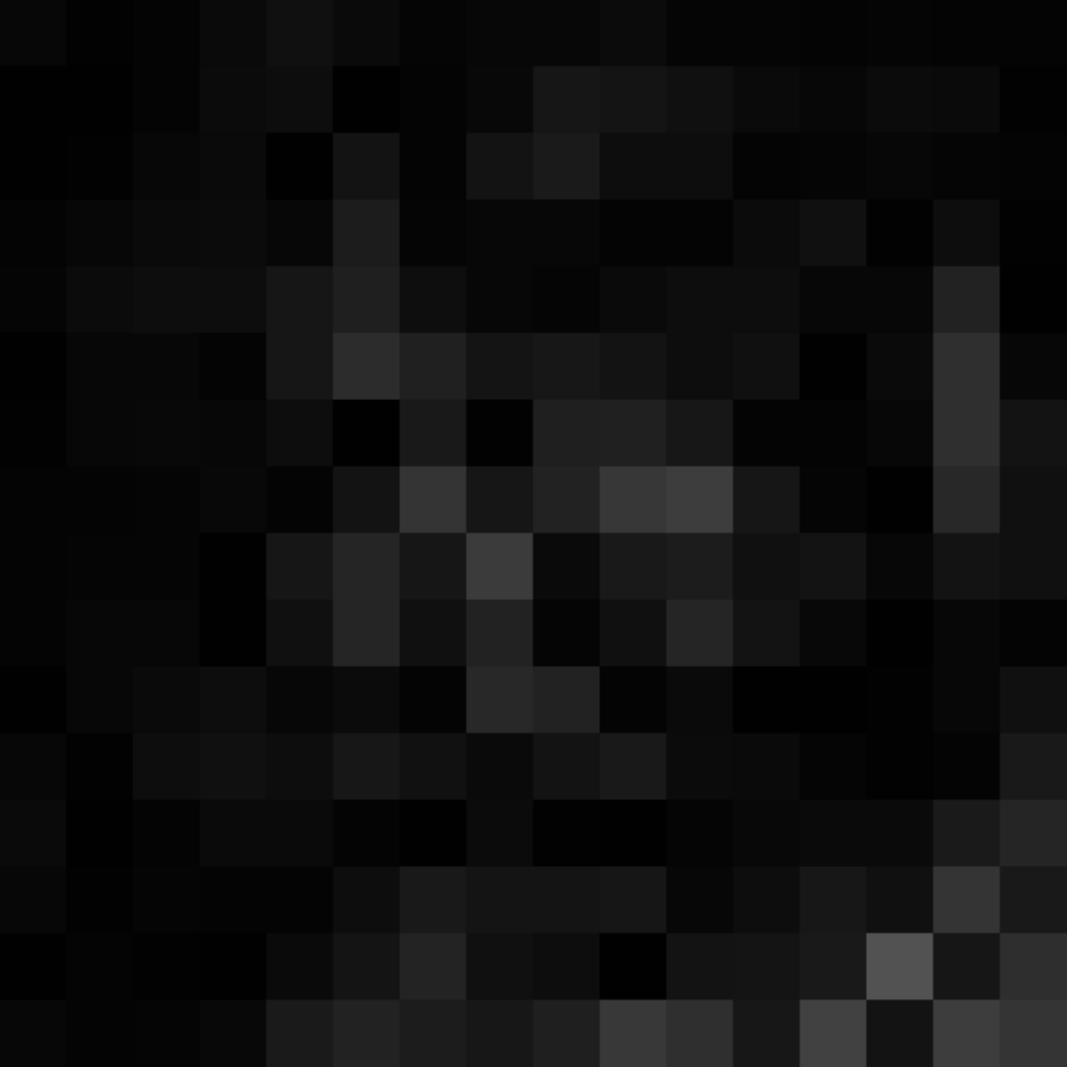}\\
		\includegraphics[width=1.0\linewidth]{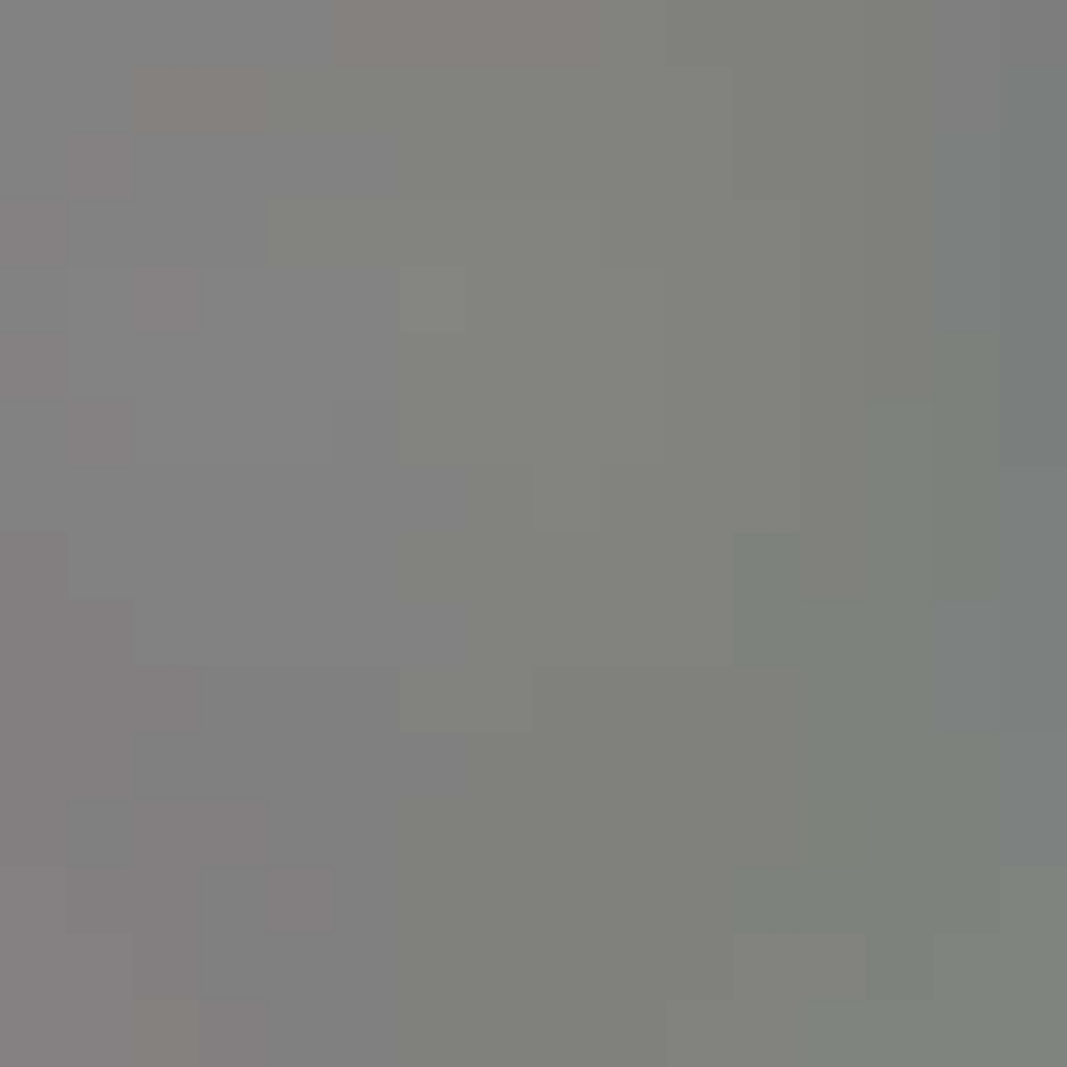}
		\caption{$\hat{I_4^t}$}
	\end{subfigure}	
	\begin{subfigure}[b]{0.13\textwidth}
		\centering
		\includegraphics[width=1.0\linewidth]{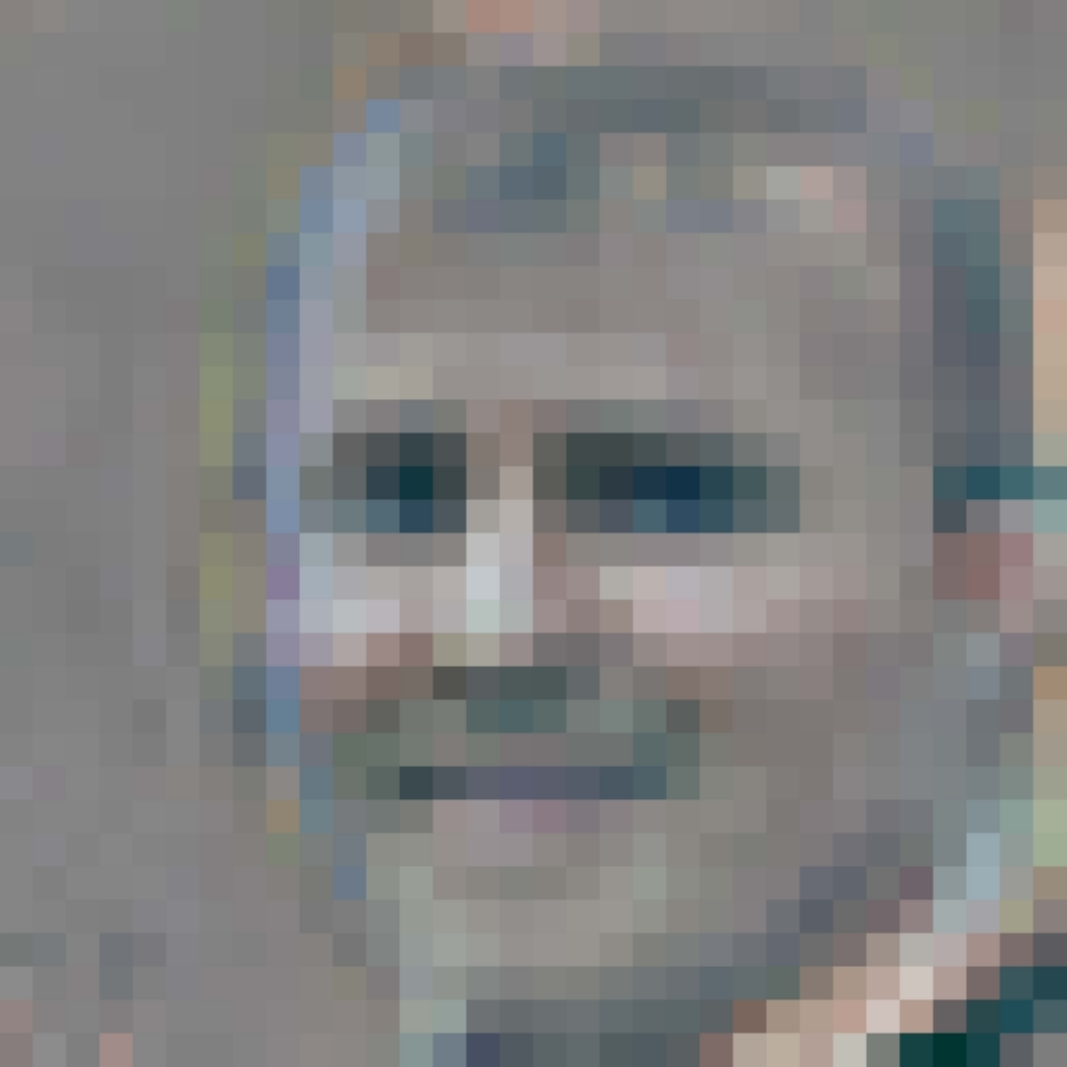}\\
		\includegraphics[width=1.0\linewidth]{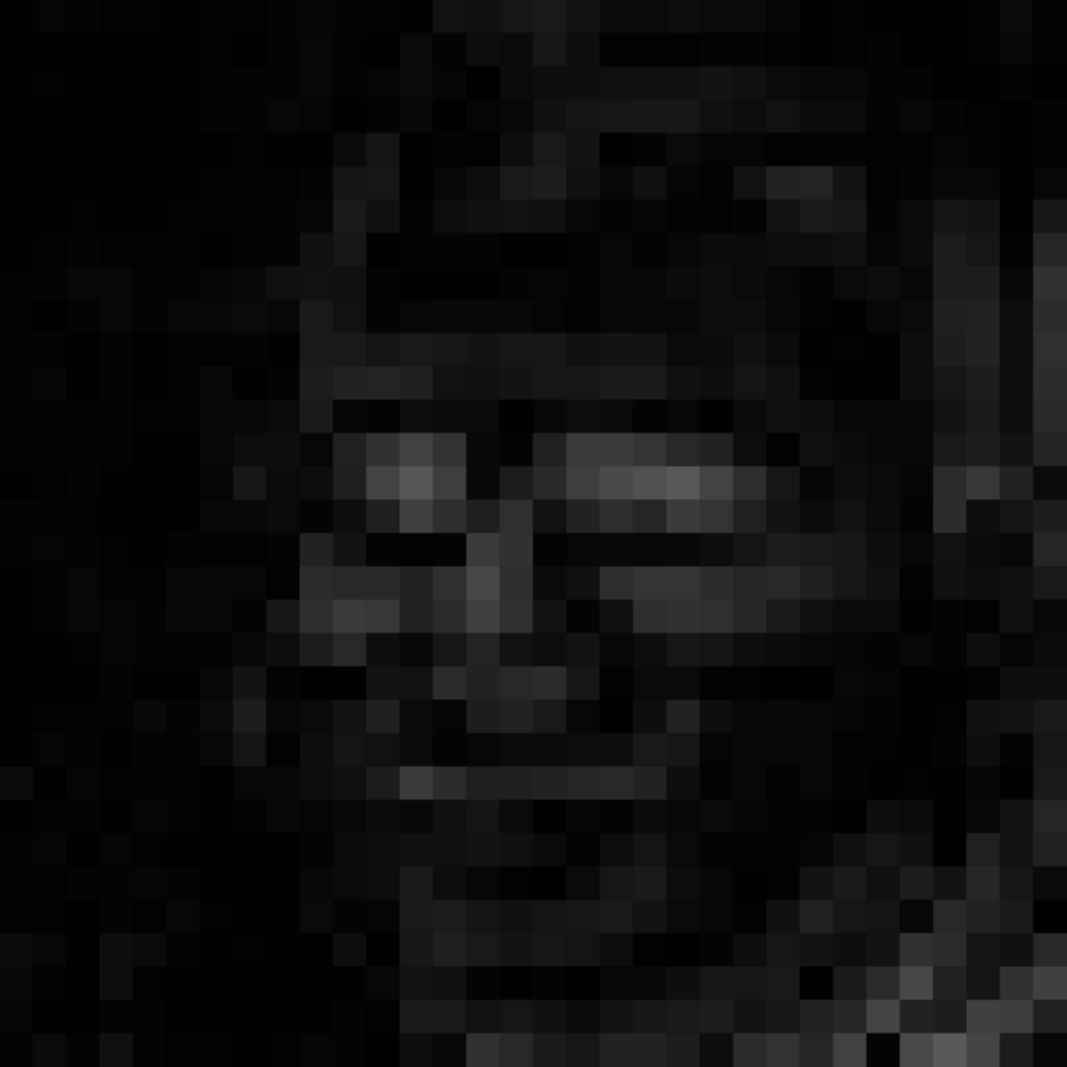}\\
		\includegraphics[width=1.0\linewidth]{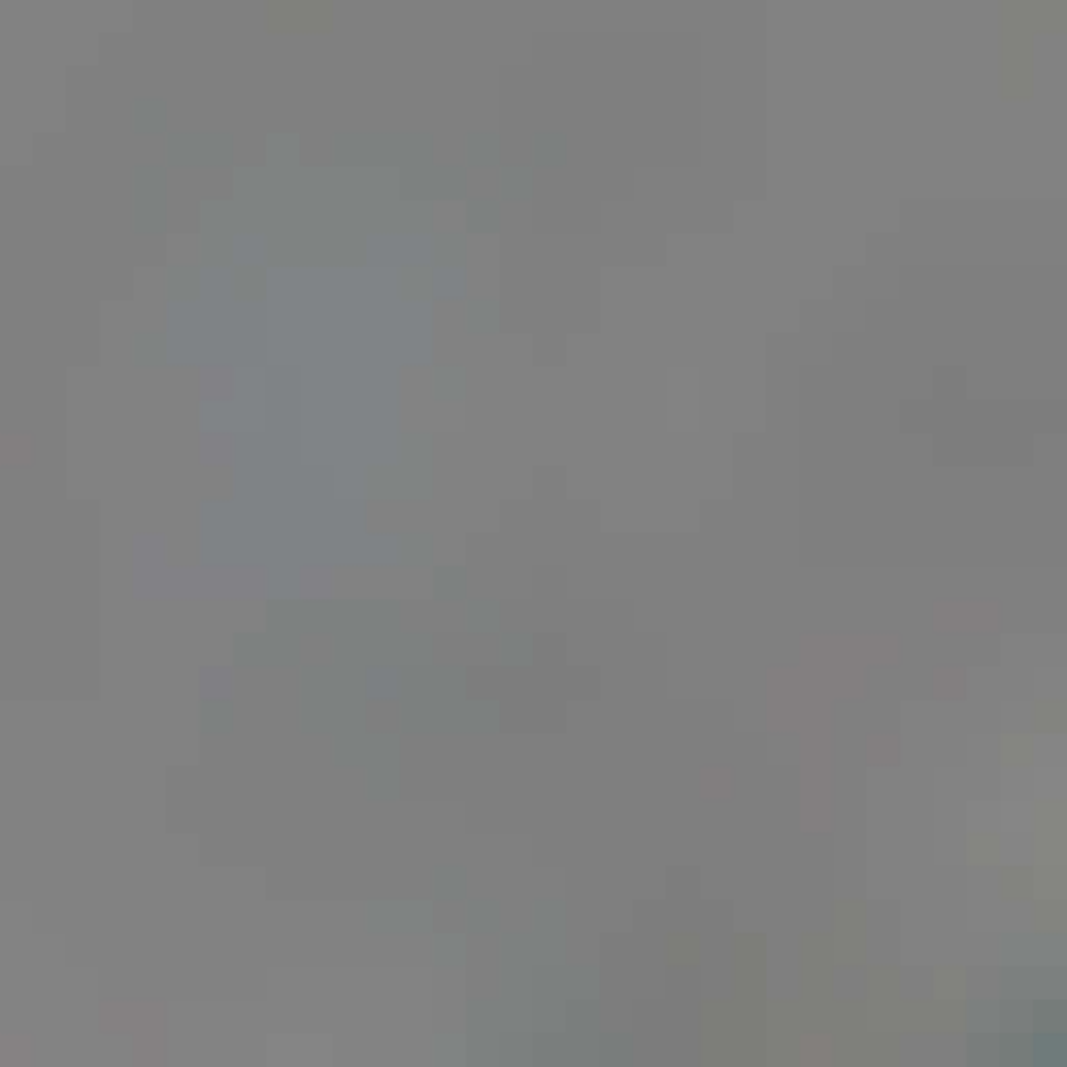}
		\caption{$\hat{I_5^t}$}
	\end{subfigure}	
	\begin{subfigure}[b]{0.13\textwidth}
		\centering
		\includegraphics[width=1.0\linewidth]{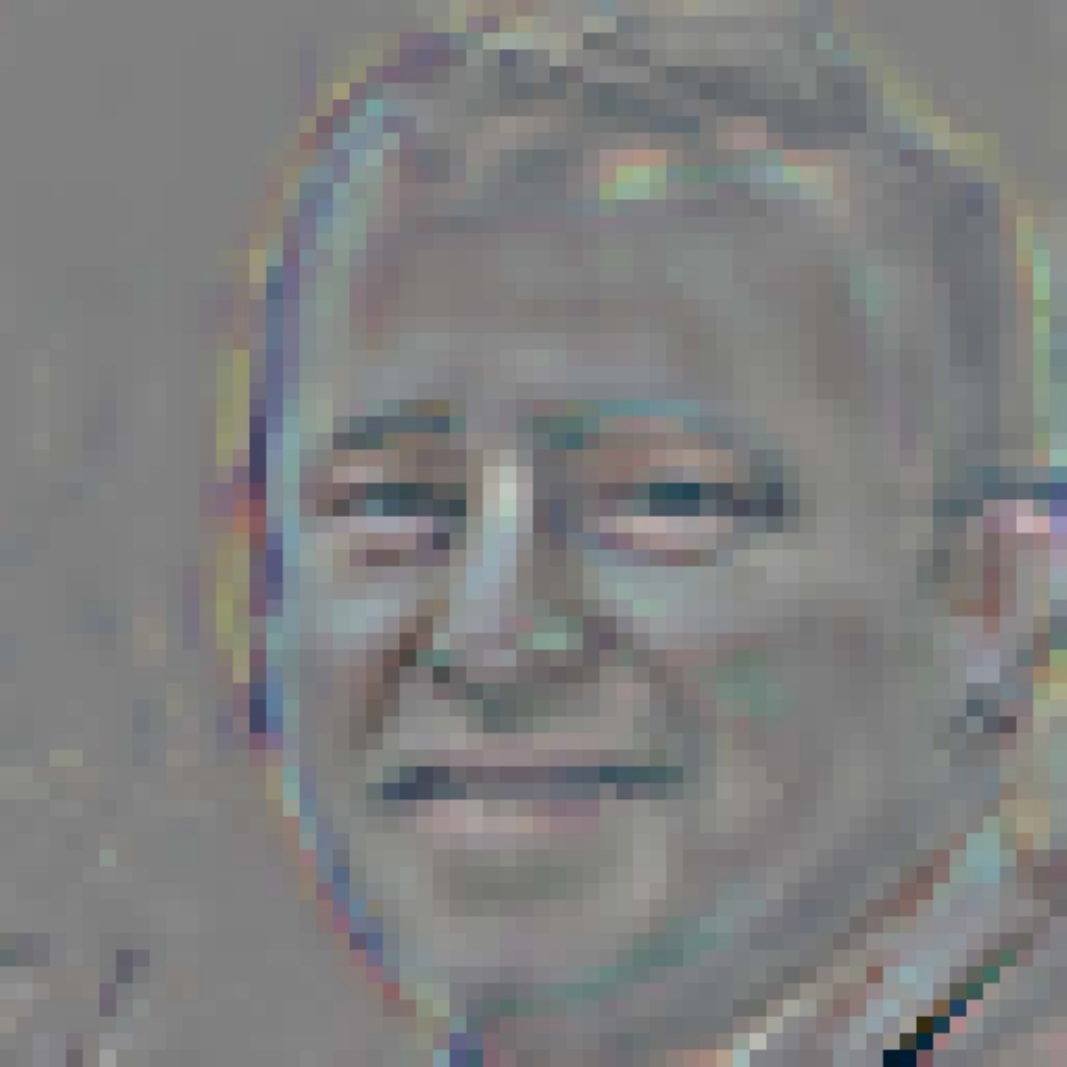}\\
		\includegraphics[width=1.0\linewidth]{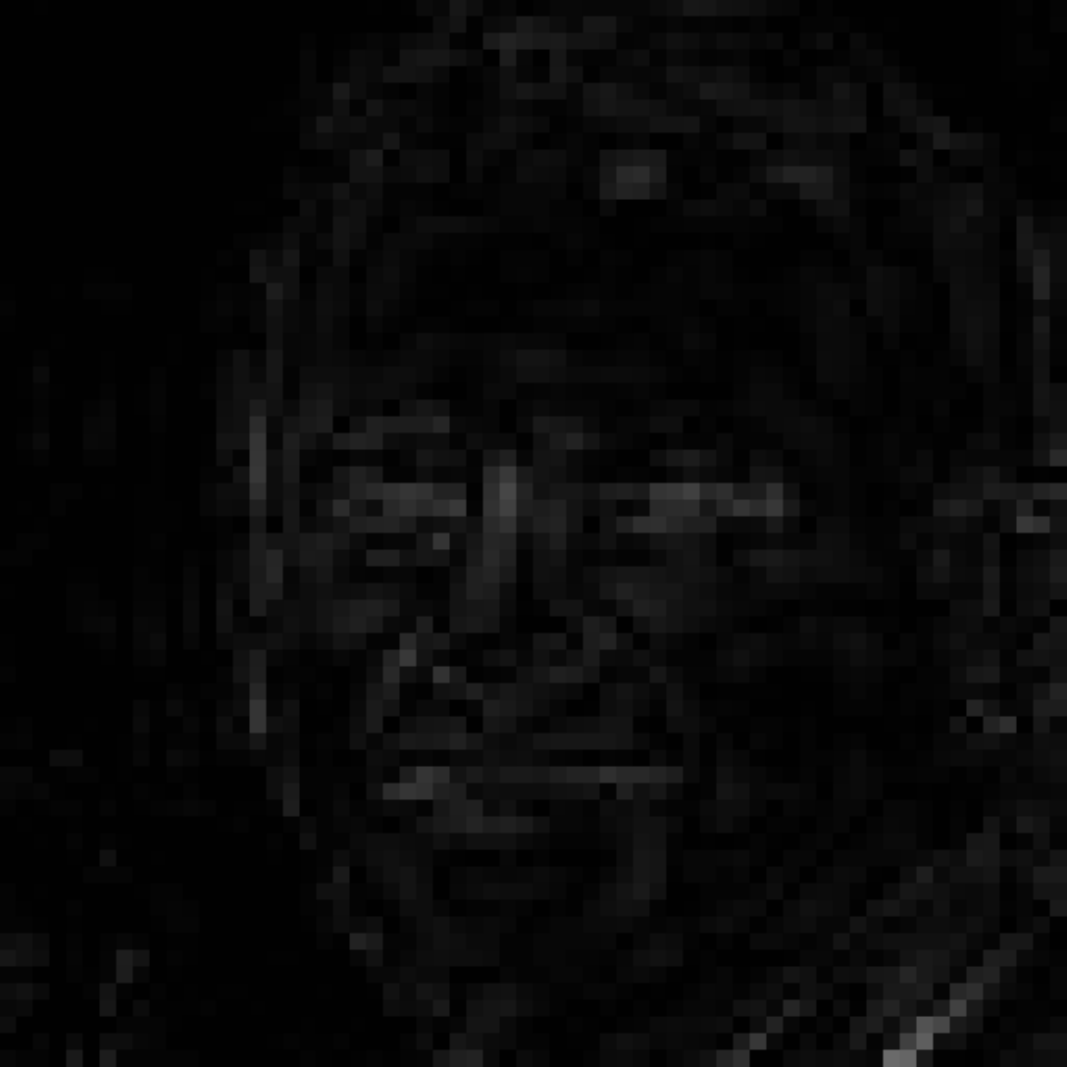}\\
		\includegraphics[width=1.0\linewidth]{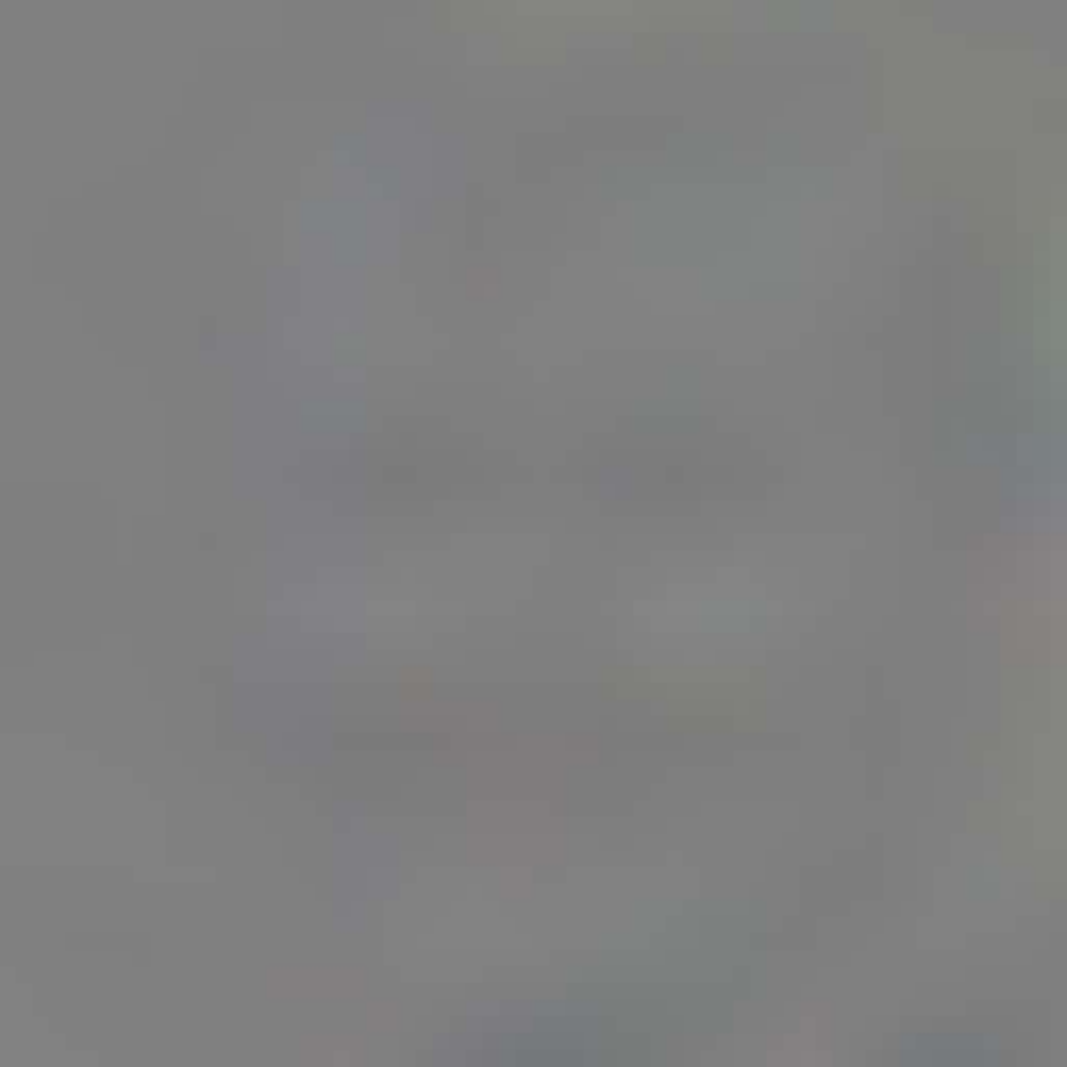}
		\caption{$\hat{I_6^t}$}
	\end{subfigure}	
	\begin{subfigure}[b]{0.13\textwidth}
		\centering
		\includegraphics[width=1.0\linewidth]{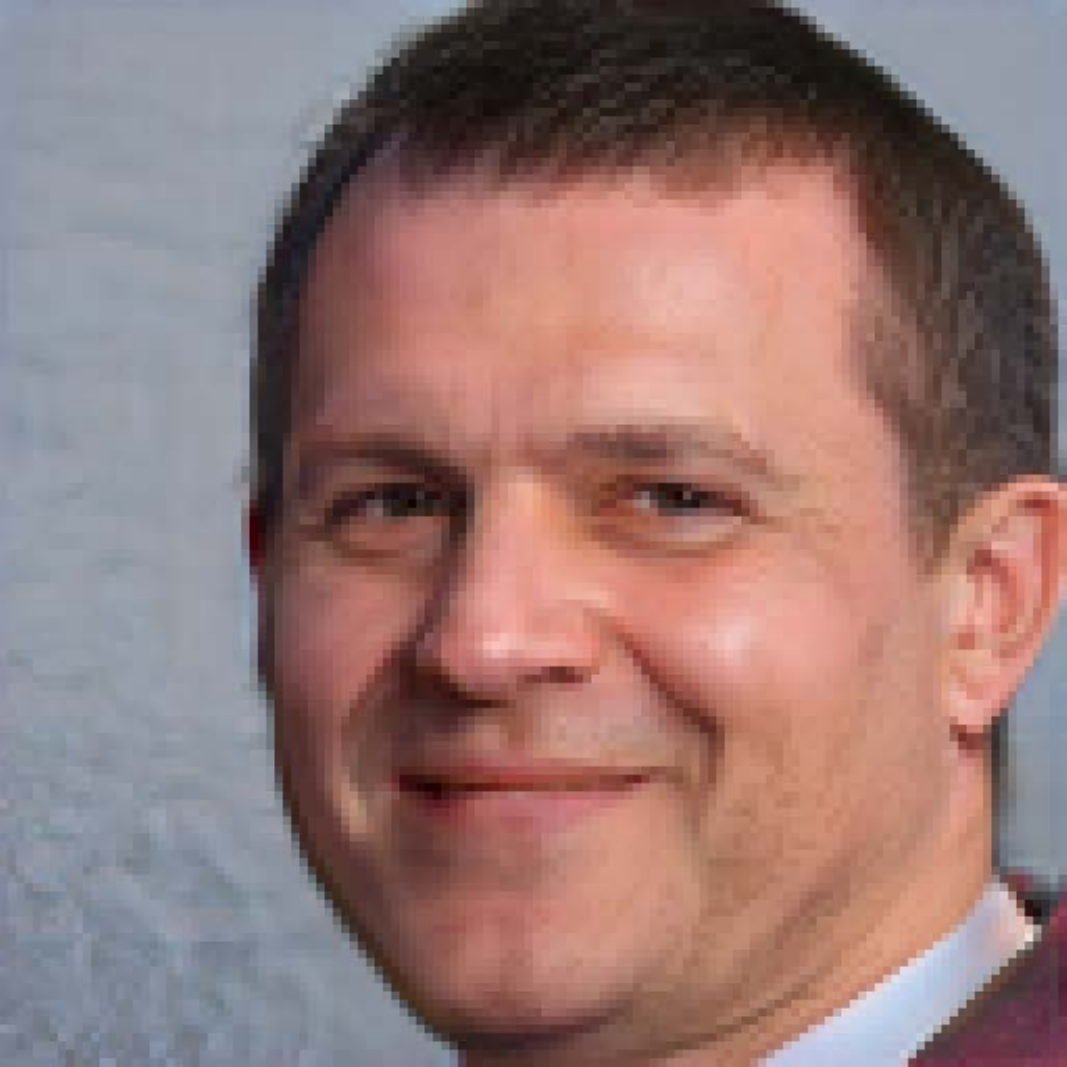}\\
		\includegraphics[width=1.0\linewidth]{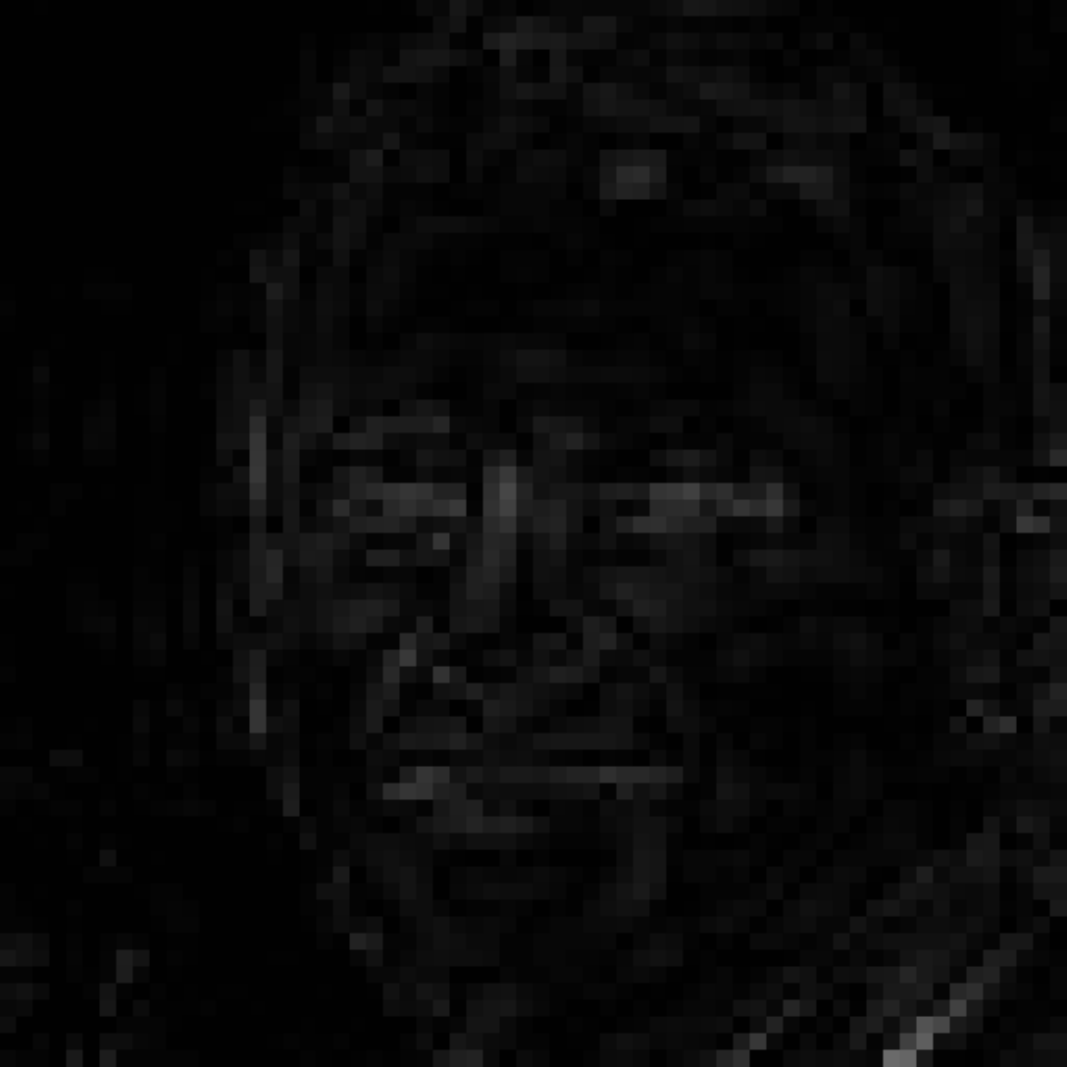}\\
		\includegraphics[width=1.0\linewidth]{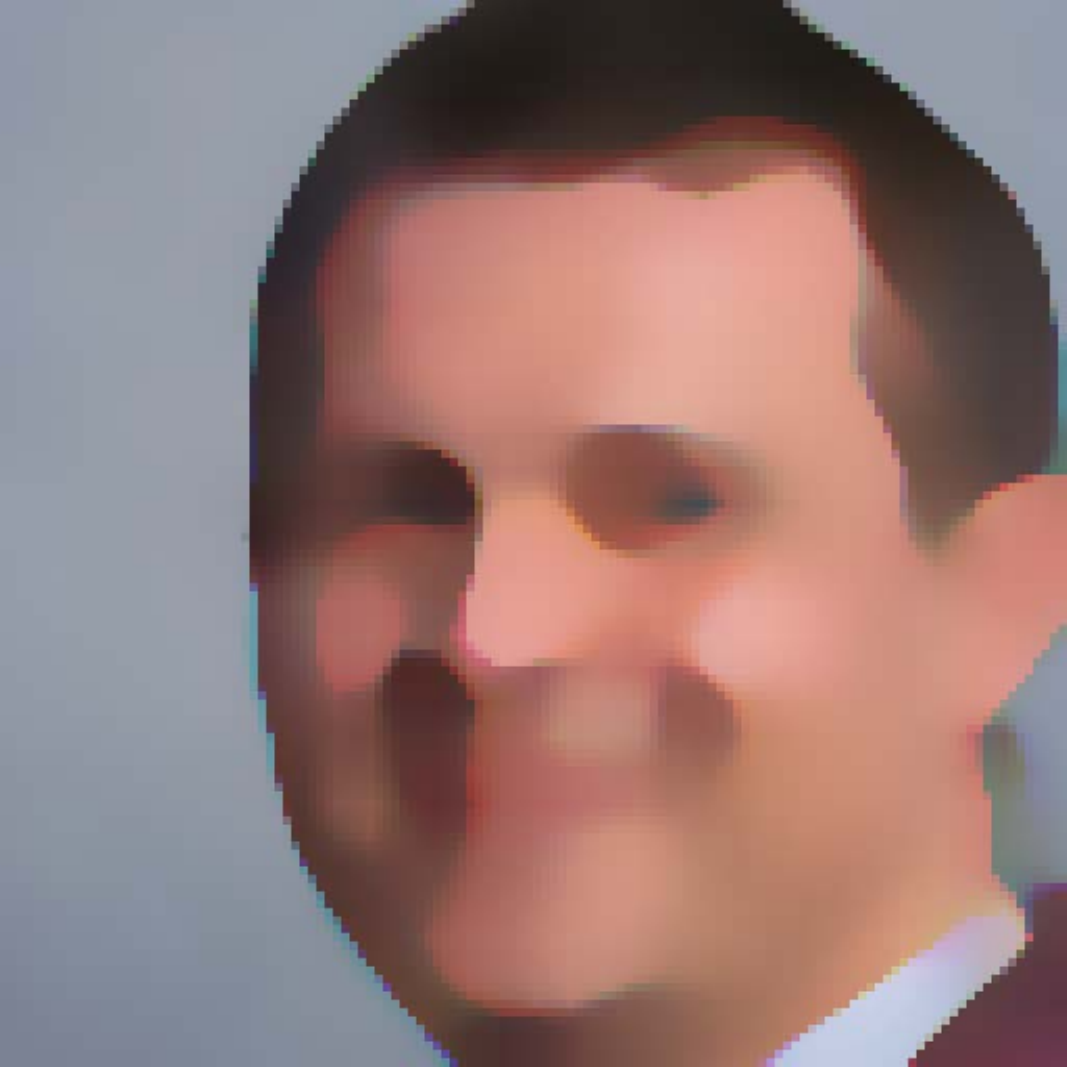}
		\caption{$\hat{I_7}$}
	\end{subfigure}	
	\begin{subfigure}[b]{0.13\textwidth}
		\centering
		\includegraphics[width=1.0\linewidth]{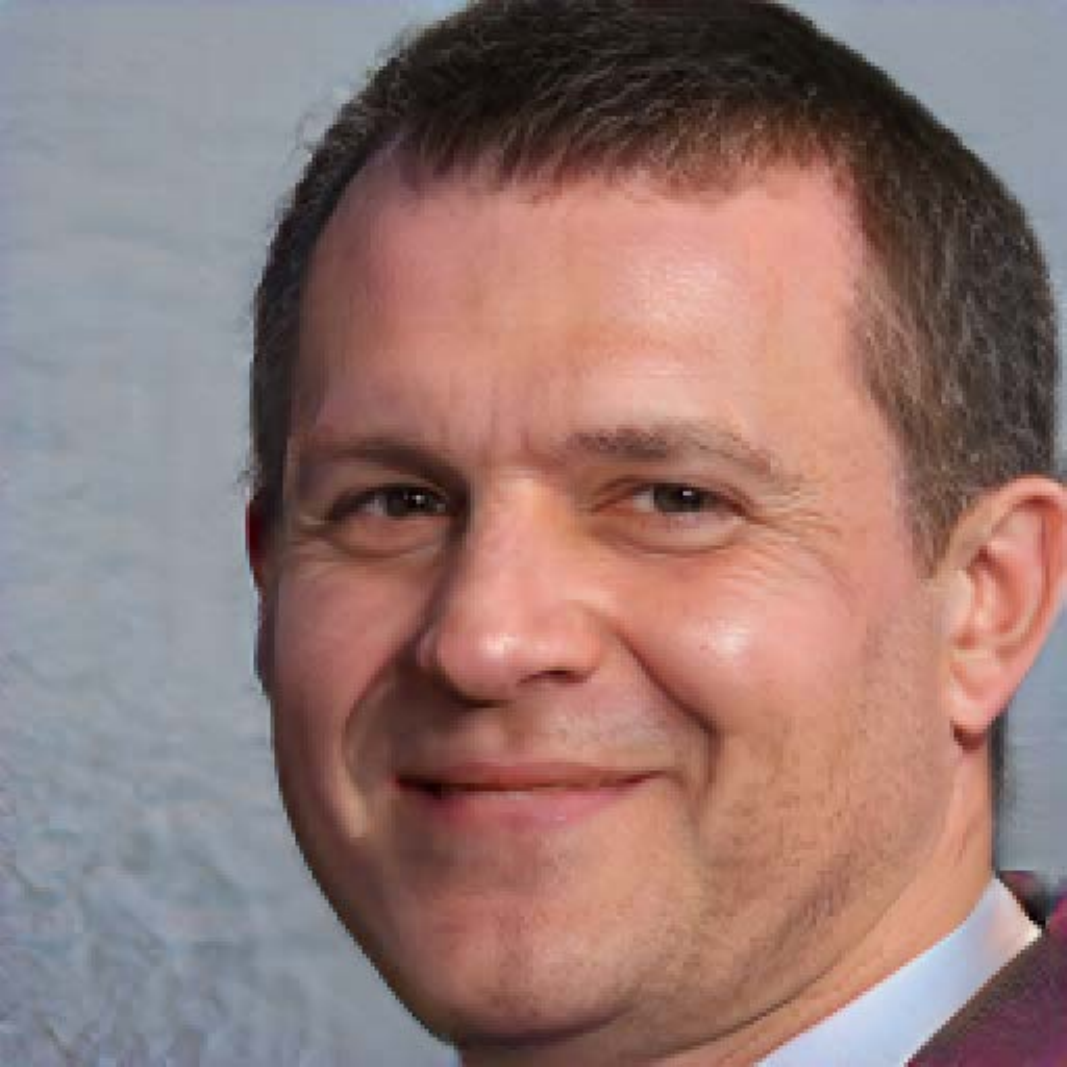}\\
		\includegraphics[width=1.0\linewidth]{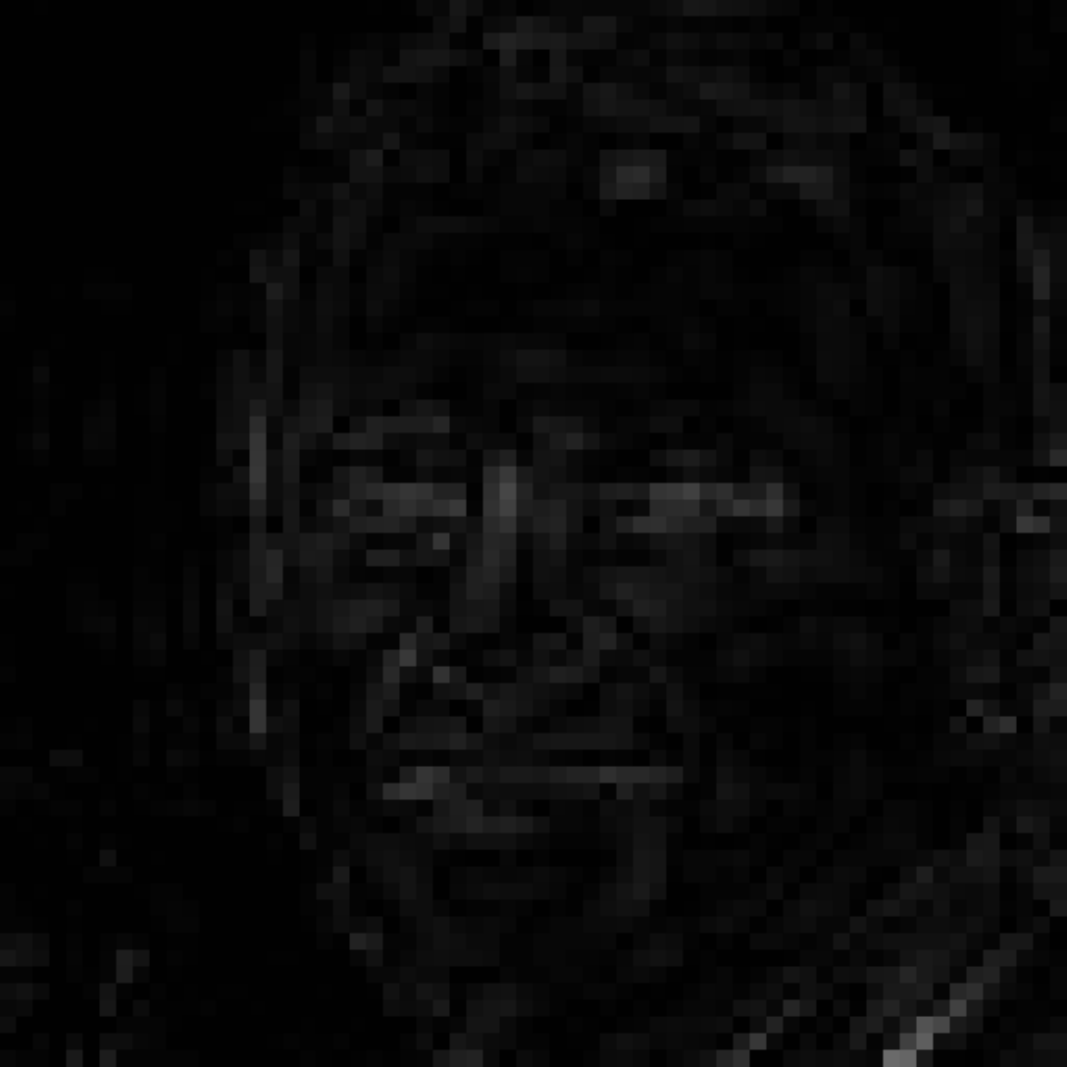}\\
		\includegraphics[width=1.0\linewidth]{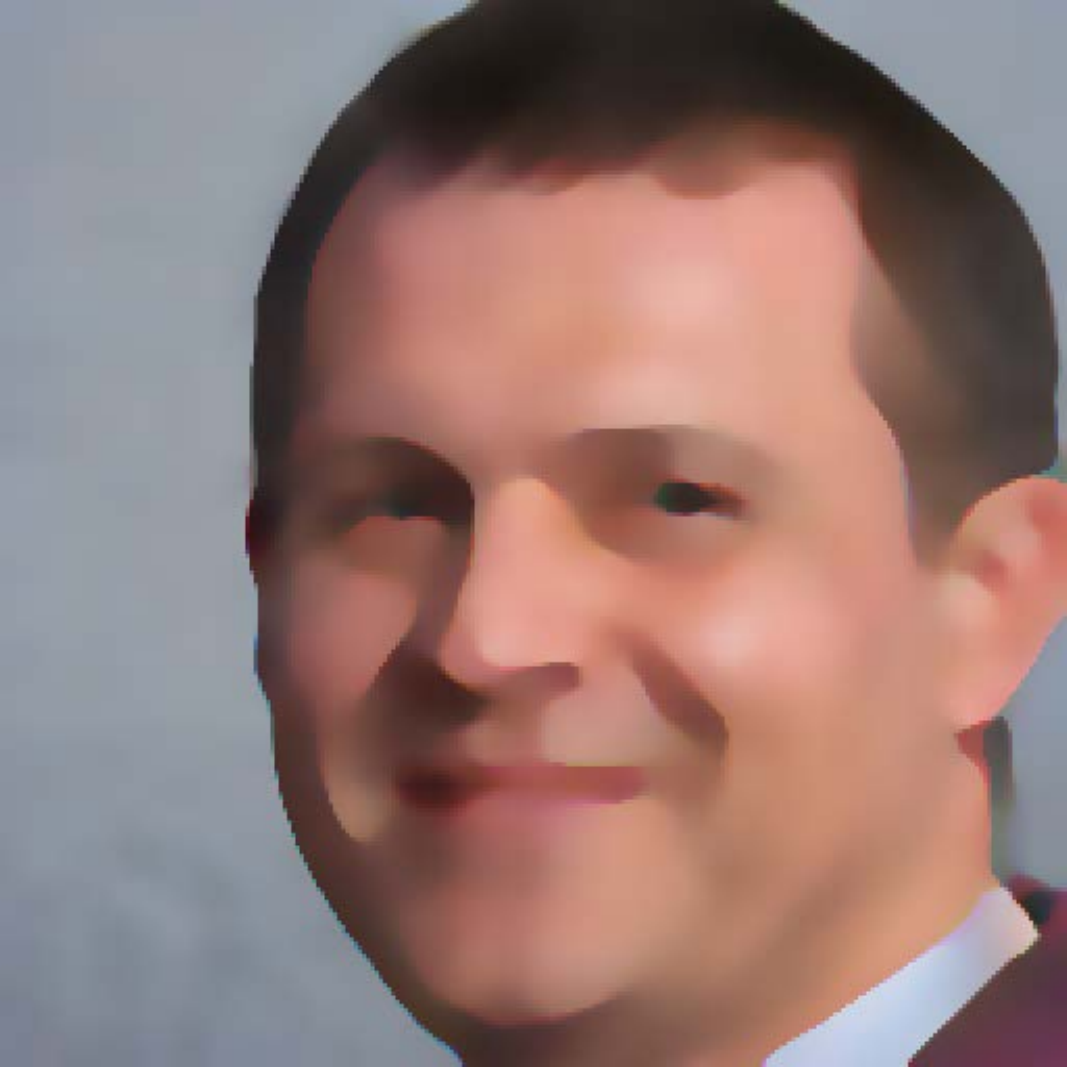}
		\caption{$\hat{I_8}$}
	\end{subfigure}	
	\begin{subfigure}[b]{0.13\textwidth}
		\centering
		\includegraphics[width=1.0\linewidth]{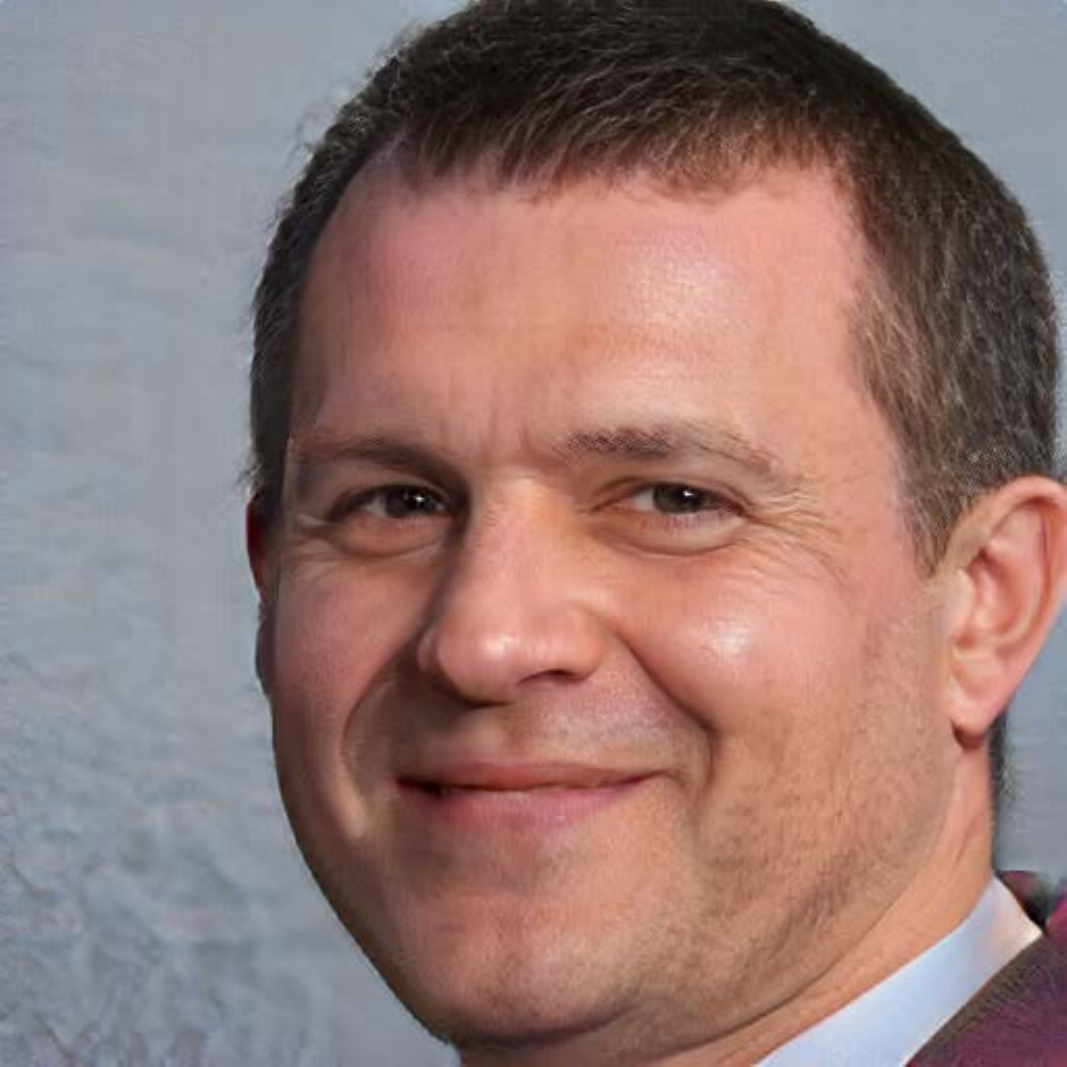}\\
		\includegraphics[width=1.0\linewidth]{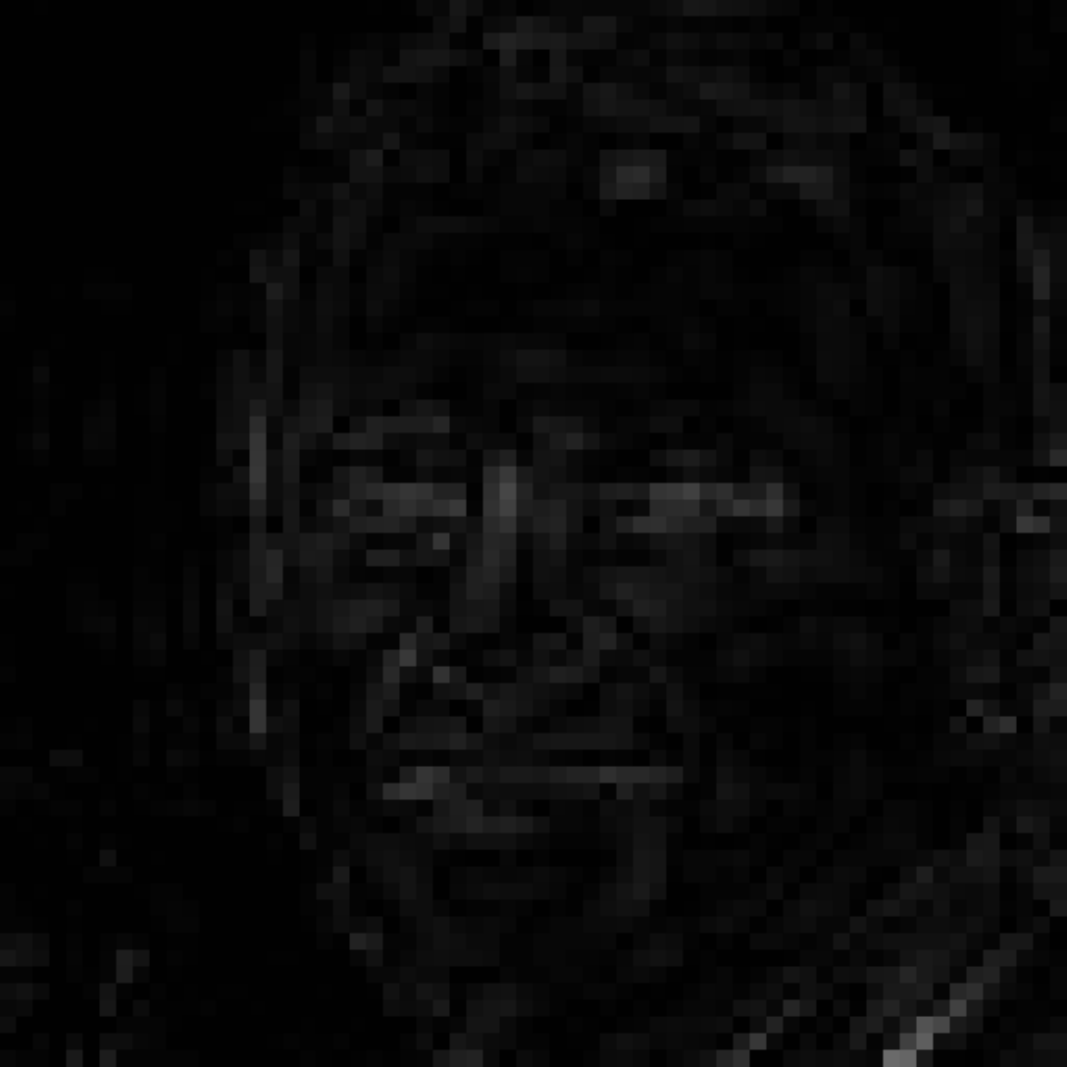}\\
		\includegraphics[width=1.0\linewidth]{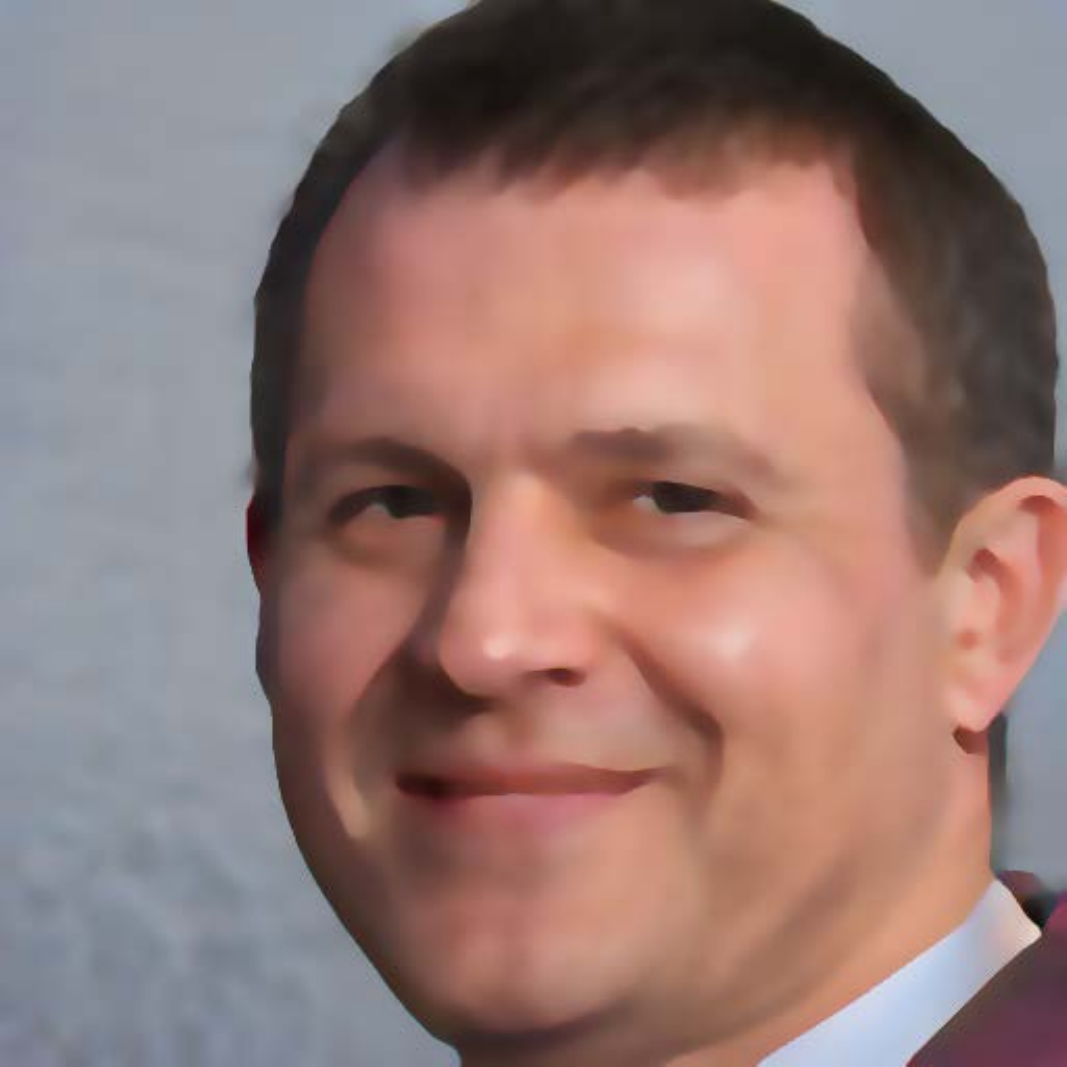}
		\caption{$\hat{I_9}$}
	\end{subfigure}	
	\caption{Synthesized faces of STGAN-WO. (a) shows $\hat{I_3^t}$ (the upper row) and its corresponding texture (the middle row) and structure (the bottom row) components. In the same way, (b)-(g) shows that of $\hat{I_4^t}$-$\hat{I_9}$, respectively.}
	\label{fig:fig34}
\end{figure*}

\section{Evaluations} \label{Exp}
Our goal is towards disentangling the latent space to achieve better face attribute editing in an unsupervised way.
To achieve this, we introduce (a) weight decomposition, (b) orthogonal regularization, and (c) the structure-texture independent architecture. To verify the validity, comparative trials are conducted. 
We first introduce our experiment setup and show how face synthesis is conducted inside STGAN-WO, then show the perceptual results by conducting attribute editing using those techniques, and report the quantitative evaluation results. 

\begin{figure*}[htp]
	\centering
	\begin{subfigure}[b]{0.95\textwidth}
		\centering
		\includegraphics[height=0.12\linewidth]{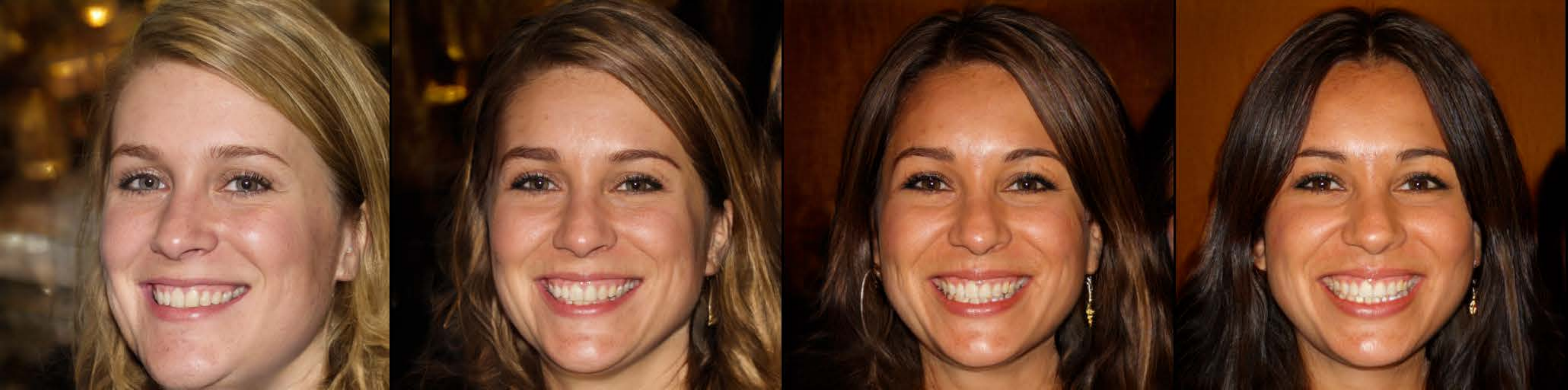}	
		\includegraphics[height=0.12\linewidth]{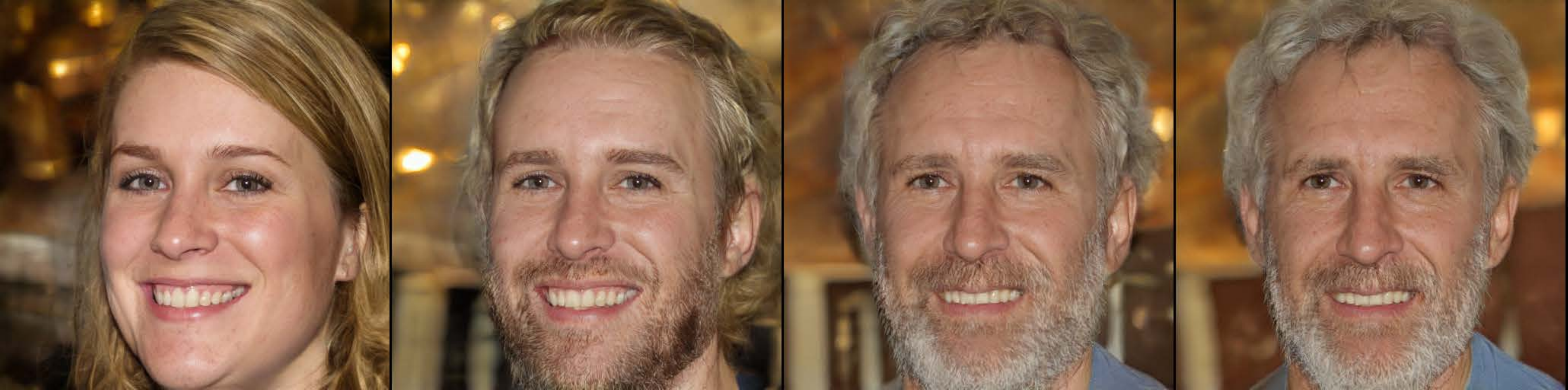}	
		\caption{Baseline}
	\end{subfigure}	
	\begin{subfigure}[b]{0.95\textwidth}
		\centering
		\includegraphics[height=0.12\linewidth]{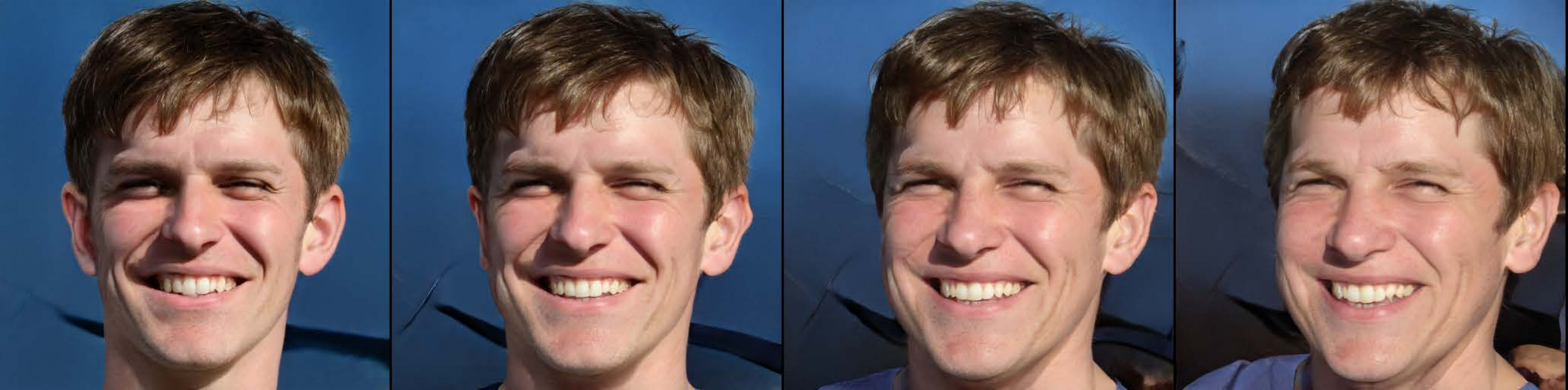}	
		\includegraphics[height=0.12\linewidth]{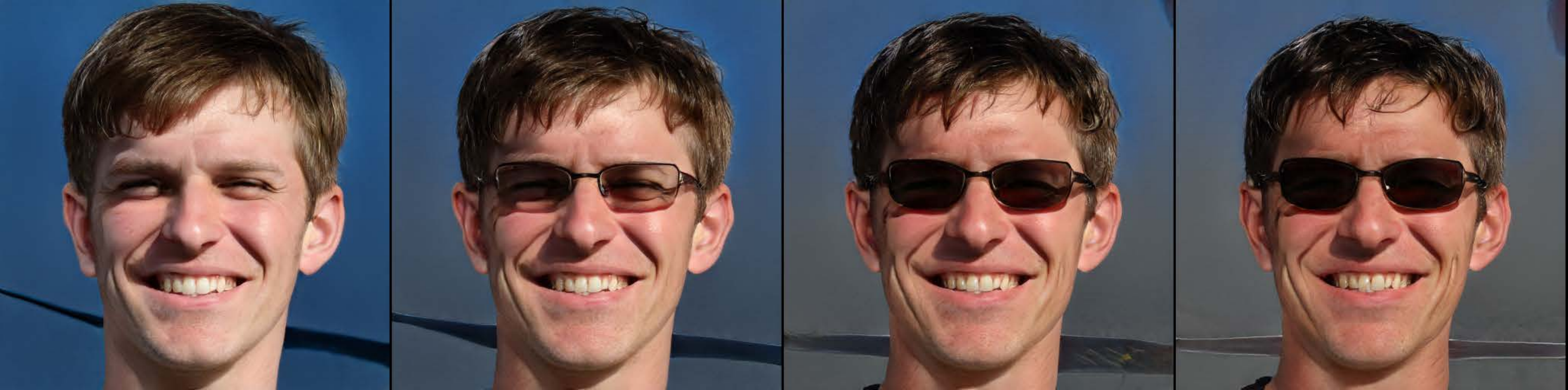}	
		\caption{Config A}
	\end{subfigure}	
	\begin{subfigure}[b]{0.95\textwidth}
		\centering
		\includegraphics[height=0.12\linewidth]{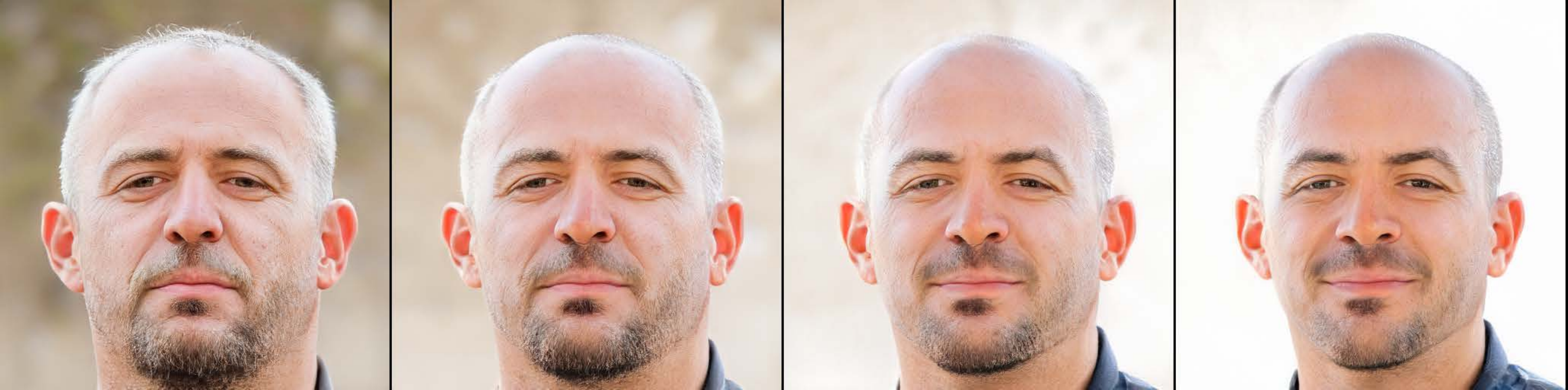}	
		\includegraphics[height=0.12\linewidth]{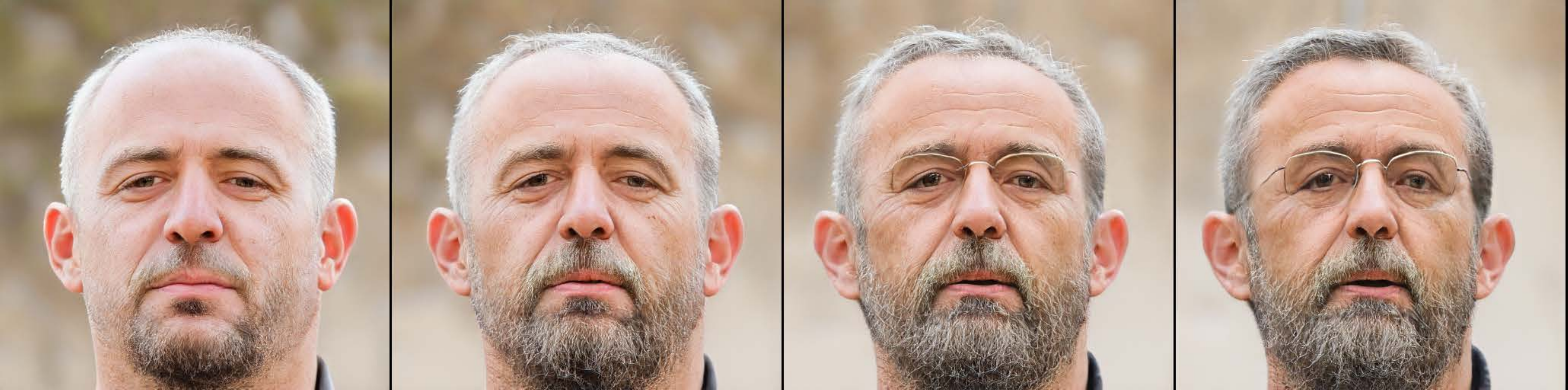}	
		\caption{Config B}
	\end{subfigure}	
	\begin{subfigure}[b]{0.95\textwidth}
		\centering
		\includegraphics[height=0.12\linewidth]{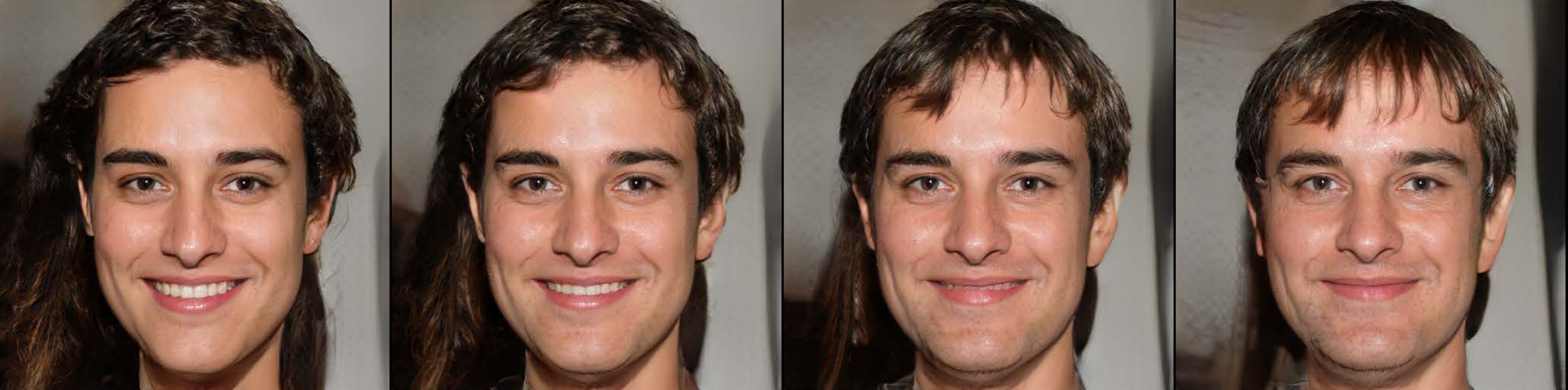}	
		\includegraphics[height=0.12\linewidth]{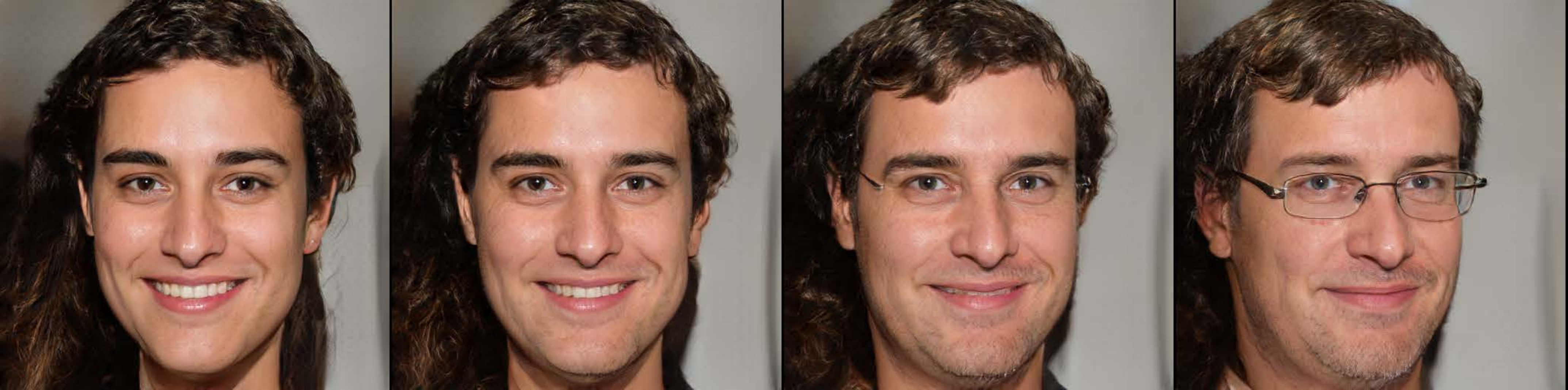}	
		\caption{Config C - moving $w_1$ along its orthogonal directions}
	\end{subfigure}		
	\begin{subfigure}[b]{0.95\textwidth}
		\centering
		\includegraphics[height=0.12\linewidth]{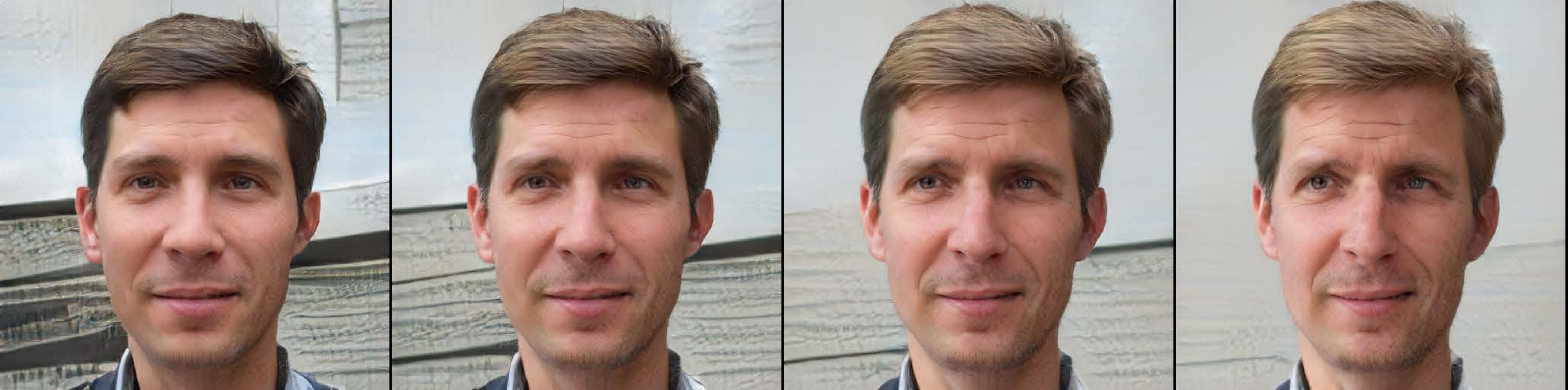}	
		\includegraphics[height=0.12\linewidth]{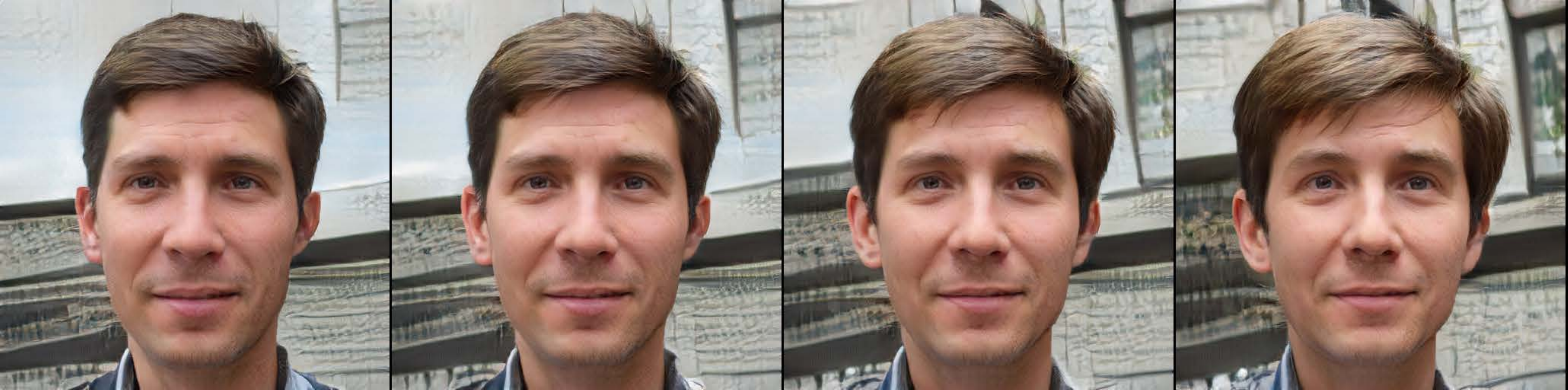}	
		\caption{Config D - moving $w_1$ along its orthogonal directions}
	\end{subfigure}	
	\begin{subfigure}[b]{0.95\textwidth}
		\centering
		\includegraphics[height=0.12\linewidth]{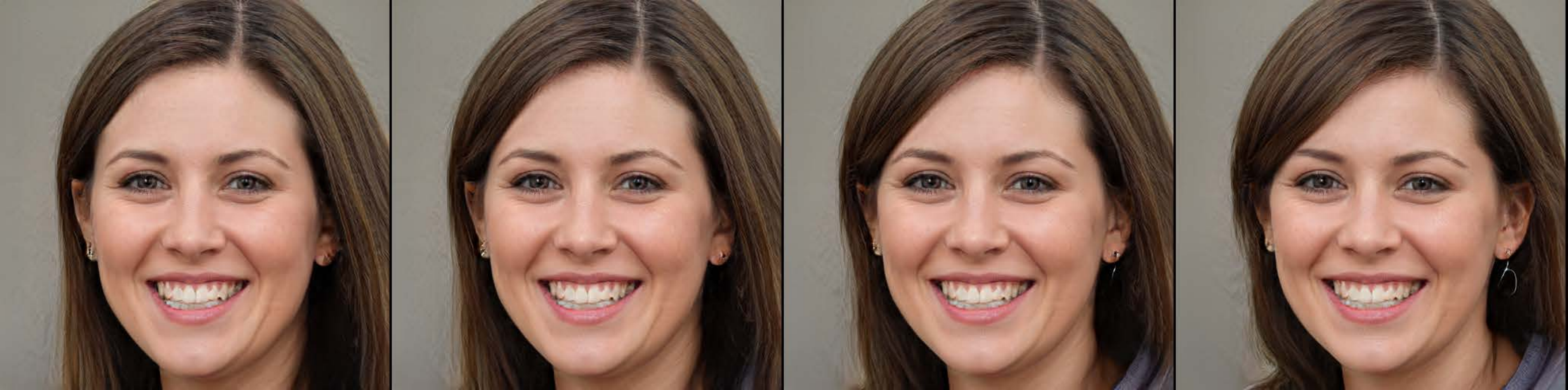}	
		\includegraphics[height=0.12\linewidth]{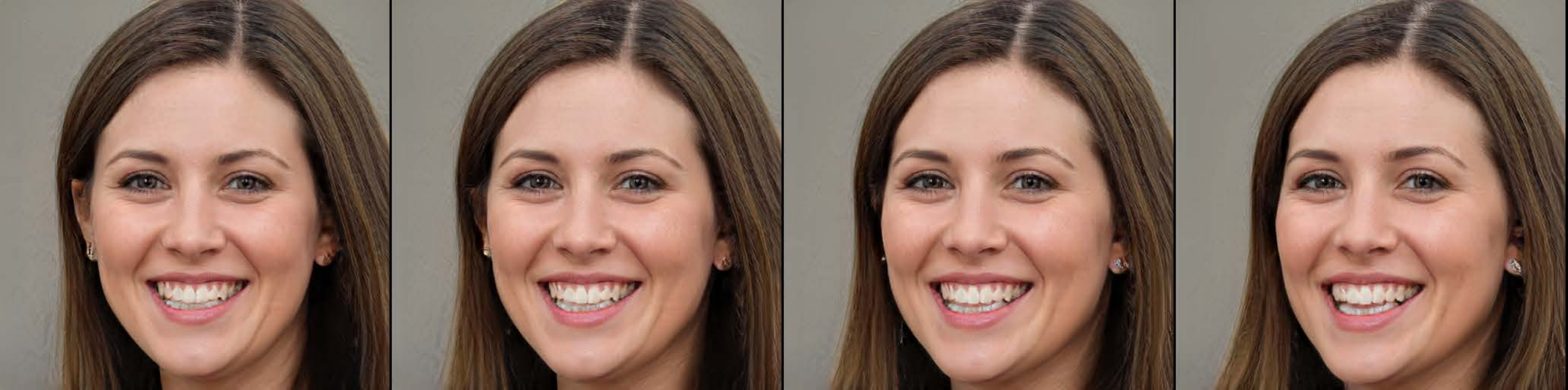}	
		\caption{STGAN-WO - moving $w_1$ along its orthogonal directions}
	\end{subfigure}	
	\caption{Face editing of various GAN models via moving the intermediate latent code $w$ along its orthogonal directions randomly. Specifically, two  latent vectors are involved in Config C, Config D and Config E, and (d)-(f) show the face editing results via moving $w_1$ along its orthogonal directions.}
	\label{fig:fig7}
\end{figure*}

\subsection{Experiment Setup}
Comparative trials are conducted to test the performances of those proposed techniques.
The improved version of StyleGAN \cite{Karras2020} is utilized as a baseline to conduct face generation based on the FFHQ dataset \cite{Karras2019}. Considering the limitation of computational facilities, face images are resized to 512 $\times$ 512 pixels to conduct the image generation tasks. All the following networks inherit the StyleGAN network, including its hyper-parameters and the loss objectives for training, except where stated otherwise. 
Specifically, Config A applies the weight decomposition technique as a replacement of weight demodulation.
Config B applies the orthogonal regularization based on Config A. As indicated by (\ref{key7}), it is obvious that orthogonal regularization would restrict the capacity of the weights $U$ and $V$, and we have found that enforcing the orthogonal regularization  on the weights of coarse layers are sufficient for better attribute disentanglement, and constraining the weights in the fine layers greatly decreases  the performance.  Hence, Config B utilizes the orthogonal regularization to constrain the weights in the coarse layers, whose size is less than $2^{r+1} \times 2^{r+1}$.
Config C, Config D and STGAN-WO use the structure-texture independent architecture based on Baseline, Config A and Config B, respectively. 
The structure and texture components are obtained by the structure-texture decomposition algorithm in \cite{Xu2012}. The  structure-texture independent architecture shows robustness to the structure-texture decomposition algorithms. Hence, users may choose other structure-texture decomposition algorithms to conduct experiments.
The hyper-parameter $r$ in  the structure-texture independent architecture is taken as 6, which has seen good performance. 

\subsection{Face Synthesis Using STGAN-WO}
Before evaluating the performance of STGAN-WO, we first show how face synthesis is conducted.
As shown in Fig. \ref{fig:fig5} (a), $z_1$ and $w_1$ are responsible for synthesizing the texture components of $\hat{I_3^t}$-$\hat{I_6^t}$, and the structure parts are then produced in $\hat{I_7}$-$\hat{I_9}$ with the control of $z_2$ and $w_2$.
Fig. \ref{fig:fig34} shows the synthesized faces and their corresponding texture and structure components.
It is clearly seen that there almost exist no structure components in $\hat{I_3^t}$-$\hat{I_6^t}$, and the structure components appear starting from $\hat{I_7}$, consistence with the design of the structure-texture independent architecture and demonstrating that $z_1$ and $z_2$ control the synthesis of the texture and structure components independently. 
In Section \ref{ap2}, we conduct ablation study and utilize the metric mentioned in the following section (Perceptual Path Length, PPL) to further demonstrate that $z_1$ and $z_2$ control the synthesis of the texture and structure components independently.

\begin{figure*}[htp]
	\centering
	\begin{subfigure}[b]{0.95\textwidth}
		\centering
		\includegraphics[height=0.10\linewidth]{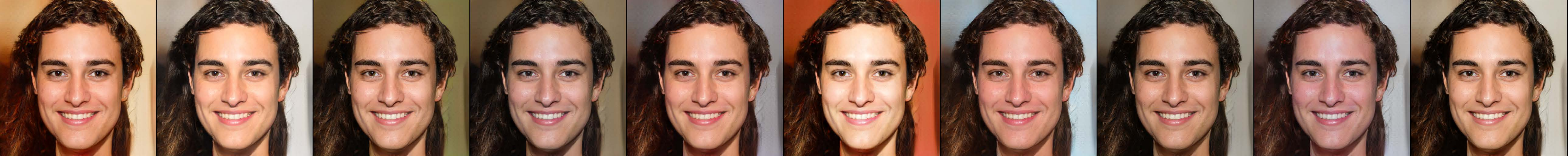}	
		\caption{Config C - change $w_2$}
	\end{subfigure}	
	\begin{subfigure}[b]{0.95\textwidth}
		\centering
		\includegraphics[height=0.10\linewidth]{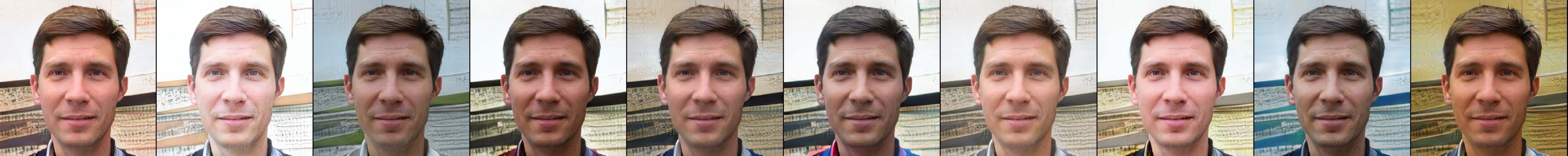}		
		\caption{Config D - change $w_2$}
	\end{subfigure}	
	\begin{subfigure}[b]{0.95\textwidth}
		\centering
		\includegraphics[height=0.10\linewidth]{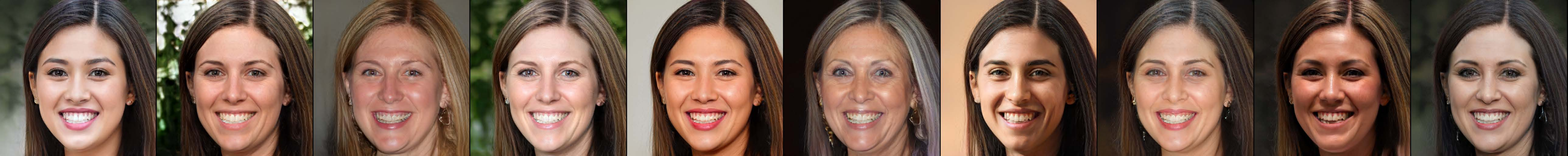}	
		\caption{STGAN-WO - change $w_2$}
	\end{subfigure}	
	\caption{Attribute editing via changing $w_2$ in Config C, Config D and STGAN-WO.}
	\label{fig:fig8}
\end{figure*}
\subsection{Perceptual Evaluation}
We would like to manipulate the intermediate latent code to figure out how perceptual changes would happen, and investigate the effect of those proposed techniques on attribute disentanglement.
Specifically, for Config C, Config D and STGAN-WO, we edit the texture related attributes via moving $w_1$ along its orthogonal directions as shown in (\ref{key15}). To manipulate structure related attributes, we change $w_2$.
In addition, we move the latent code $w$ along its orthogonal directions in Baseline, Config A and Config B to investigate the effect on perceptual results.
Perceptual results are shown in Fig. \ref{fig:fig7} and  Fig. \ref{fig:fig8}.
Firstly, Fig. \ref{fig:fig7} (a) shows the face editing results of StyleGAN.
Owing to the highly entangled latent space of StyleGAN, moving the latent code $w$ along its orthogonal directions would affect multiple attributes like expressions, face color, face outline, \textit{etc}. It is nearly impossible to change some specific attributes individually.

Secondly, as shown in in Fig. \ref{fig:fig7} (a), (b) and (c), it is obvious that the structure component  is entangled with the texture parts when conducting attribute editing, and changing the latent vector would affect texture component related attributes like face outline, as well as structure  related ones like face color. As a comparison, Config C, Config D and STGAN-WO have improved the perceptual results on decoupling the attributes over Baseline, ConfigA and Config B, respectively. To be specific, as shown in Fig. \ref{fig:fig7} (d), (e) and (f), conducting face editing with  Config C, Config D and STGAN-WO affect a decreasing number of attributes, and structure component related attributes have been disentangled with texture related ones, demonstrating that attribute editing would benefit from the structure-texture independent architecture.
This is because the structure-texture independent architecture hierarchically generates the texture and structure parts, which therefore contributes to the attribute disentanglement.

Thirdly, Config A and Config D apply the weight decomposition technique as a replacement of weight demodulation, which is used in Baseline and  Config C.
It is easier to find the correlation between face images before and after editing in Fig. \ref{fig:fig7} (b) and (e) than that in Fig. \ref{fig:fig7} (a) and (d), indicating that less attributes are affected when weight decomposition is applied, and further demonstrating the effectiveness of weight decomposition.

Fourthly, Fig. \ref{fig:fig7} (f) shows excellent performances on the attribute editing task, where moving the latent vector $w_1$ along its orthogonal directions only change the hair style (the left four images) or the poses (the right four images), while the other attributes are hardly affected. More samples of attribute editing with STGAN-WO can be found in Fig. \ref{fig:fig1}, where specific attribute editing is achieved by moving $w_1$ toward its randomly initialized directions. Compared with the results in Fig. \ref{fig:fig7} (e), it is clearly seen that orthogonal regularization is essential for better attribute disentanglement. 
In addition, Fig. \ref{fig:fig8} shows how Config C, Config D and STGAN-WO perform on face editing task by change $w_2$.
Changing $w_2$ in Config C and Config D appears to only affect the face color, and many structure component related attributes has disappeared, \textit{e.g.}, illumination, thus greatly affecting the diversity of synthesized samples. 
We reason that this is caused by the unsuccessful attribute disentanglement, and the texture related attributes are still entangled with the structure related ones, so that only part of the structure related attributes can be synthesized.
To address this, STGAN-WO utilizes the orthogonal regularization to guarantee that each $s_n$ only controls one factor of variation.
As shown in Fig. \ref{fig:fig8} (c) and Fig. \ref{fig:fig1} (a), generated faces shows more diversities, further demonstrating the importance of orthogonal regularization on attribute editing.

\begin{table*}[htp]
	\centering
	\caption{Quantitative evaluation of various GAN models. IS is Inception Score and FID is Fr\'{e}chet Inception Distance. For IS, higher is better, while lower is better for FID. Perceptual Path Length are computed based on path endpoints in $w$ according to \cite{Karras2019}. 
		For Config C, Config D and STGAN-WO, two intermediate latent codes $w_1$ and $w_2$ are involved. Thus, perceptual path length $l_{\perp}$ and  $l_{w_2}$ are calculated based on $w_1$ and $w_2$, respectively.}
	\begin{tabular}{c|c|c|c||c|c|c|c|c}
		\hline\hline	
		
		Configuration	&FID	&IS	&$l_{w}$	&Configuration	&FID	&IS	&$l_{w_2}$	&$l_{\perp}$\\
		\hline		
		Baseline 	&3.82	&5.09 $\pm$ 0.04	&130.86	&Config C	&5.51	&5.01 $\pm$ 0.11	&0.86	&0.68\\
		Config A	&3.47	&5.19 $\pm$ 0.06	&113.74	&Config D	&5.53	&5.12$\pm$ 0.05		&1.41	&0.64 \\
		Config B	&3.83  	& 5.26 $\pm$ 0.06 	&101.40 &STGAN-WO	&10.13	&4.33 $\pm$ 0.07	&13.04	&0.42\\
		\hline\hline
	\end{tabular}
	\label{t1}
\end{table*}
\begin{figure*}[htp]
	\centering
	\begin{subfigure}[b]{0.475\textwidth}
		\centering
		\includegraphics[height=0.24\linewidth]{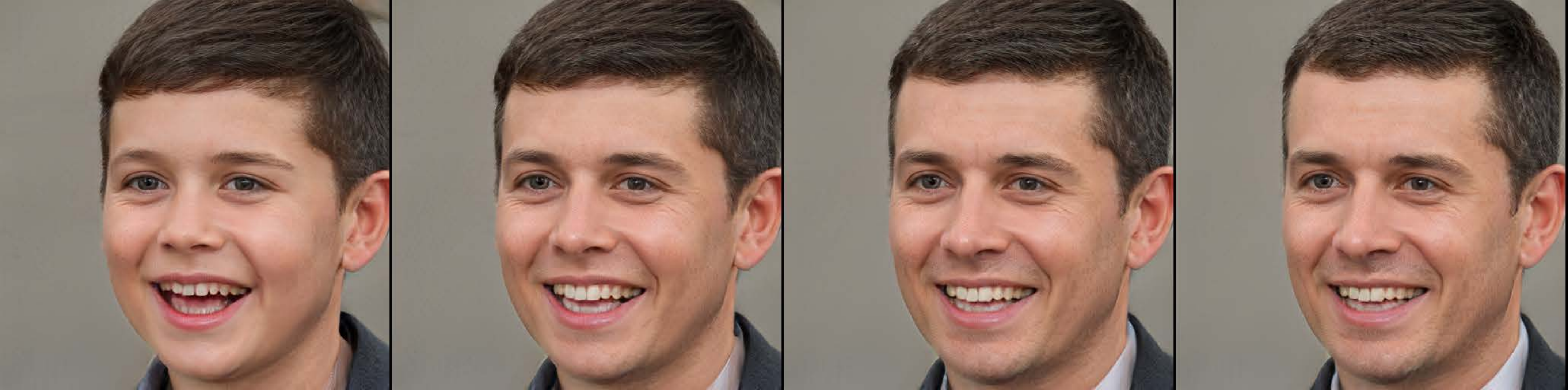}	
		\caption{Aging from teenager to adults by changing $w_1$}
	\end{subfigure}	
	\begin{subfigure}[b]{0.475\textwidth}
		\centering
		\includegraphics[height=0.24\linewidth]{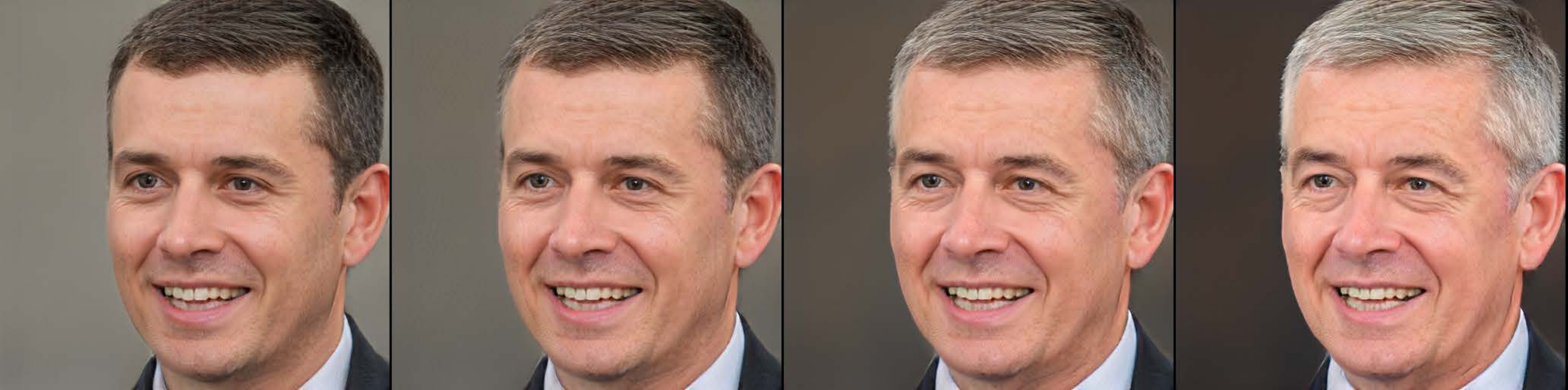}	
		\caption{Aging from adults to old age by changing $w_2$}
	\end{subfigure}	
	
	\begin{subfigure}[b]{0.950\textwidth}
		\centering
		\includegraphics[height=0.12\linewidth]{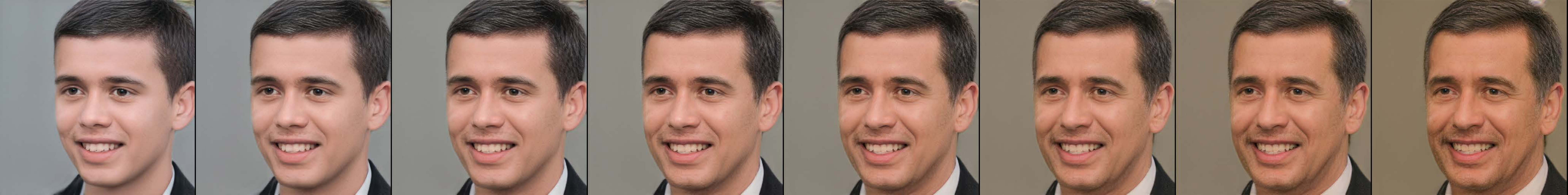}	
		\caption{Aging using InterFaceGAN}
	\end{subfigure}	
	\caption{Attribute editing of aging with STGAN-WO and InterFaceGAN. (a) and (b) show face aging using STGAN-WO, and (c) shows that of InterFaceGAN.}
	\label{fig:fig9}
\end{figure*}

To summarize, the weight decomposition technique, orthogonal regularization as well as the structure-texture independent architecture all contribute to better attribute editing, and removing any of them would decrease the performance on attribute disentanglement. In addition, it has been demonstrated that STGAN-WO has achieved good performance of face editing in an unsupervised way.

\subsection{Quantitative Evaluation}
Quantitative evaluation is performed as well to provide a comprehensive investigation of the proposed techniques.
Following the traditions in literatures, Inception score (IS) \cite{Salimans2016} and Fr\'{e}chet Inception Distance (FID) \cite{Dowson1982} are utilized as approximate measures of sample quality.
In order to quantify attribute disentanglement, the authors of StyleGAN \cite{Karras2019} has introduced the Perceptual Path Length (PPL), which measures how drastic changes the image undergoes when performing interpolation in the latent space.  Intuitively, a less curved latent space should result in perceptually smoother transition than a highly curved latent space.
Following \cite{Karras2020}, we compute the PPL metric based on path endpoints in $w$:
\begin{equation}\label{key14}
	\begin{split}
		l_w = E[&\frac{1}{\epsilon^2}d(g(lerp(w^1,w^2, t)), \\
		&g(lerp(w^1,w^2, t+\epsilon )))]
	\end{split}
\end{equation}
where $l_w$ is the PPL metric computed based on endpoint $w$, $t \sim U(0,1)$, $\epsilon$ is a small value and usually taken as $10^{-4}$, $d(\cdot, \cdot)$ evaluates the perceptual distance between the resulting images, which is computed according to \cite{Karras2019}, $w^1$ and $w^2$ are two randomly sampled latent code, $g$ is the synthesis network, and $lerp$ denotes the linear interpolation operation:
\begin{equation}
	lerp(a, b, t) = a + (b-a)*t
\end{equation}

For Baseline, Config A and Config B, we calculate $l_w$ using (\ref{key14}), while, for Config C, Config D and STGAN-WO, where two intermediate latent code $w_1$ and $w_2$ are involved, $l_{w_1}$ and $l_{w_2}$ are computed.

$w_1$ in Config C, Config D and STGAN-WO is moved along its orthogonal directions to edit some specific attributes. However, $w^1$ and $w^2$ in  (\ref{key14}) are randomly sampled, leading to the problem that $l_{w_1}$ computed by (\ref{key14}) is incapable of measuring the perceptual changes when performing attribute editing via the method in (\ref{key15}).
To address this, we can compute PPL along the orthogonal directions:
\begin{equation}\label{key16}
	l_{\perp} = E[\frac{1}{\epsilon^2}d(g(w^{1}_1), g(w^{1}_1+ \epsilon \cdot w^{\perp}_1))]
\end{equation}
where $w^{1}_1$ is the random sample from the intermediate latent space $w_1 \in \mathbb{W}_1$, and $w^{\perp}_1$ is the corresponding orthonormal vector.
Obviously, $l_{\perp}$ measures how dramatic changes happen when moving $w_1$ along its orthogonal directions. Intuitively, small value of $l_{\perp}$ means perceptually smooth transition and less attributes are affected.
Hence, we utilize $l_{\perp}$ instead of $l_{w_1}$ to quantify attribute disentanglement when conducting face manipulation in Config C, Config D and STGAN-WO.

Results of quantitative evaluation are listed in Table \ref{t1}.
Firstly, Conifg A in Table \ref{t1} indicates that applying the weight decomposition technique contributes to improving the image quality of synthesized samples and attribute disentanglement, demonstrating the effectiveness of weight decomposition again. 

Secondly, applying the orthogonal regularization and utilizing the structure-texture independent architecture indeed contribute to attribute disentanglement. 
It is clearly seen that, compared with $l_w$ in Baseline, Config A and Config B, $l_{\perp}$ and $l_{w_2}$ for Config C, Config D and STGAN-WO are of rather small values, indicating that moving $w_1$ along its orthogonal directions or changing $w_2$ results in perceptually much smoother transition in Config C, Config D and STGAN-WO than that in Baseline, Config A and Config B, and further demonstrating the effectiveness of the structure-texture independent architecture on decoupling attributes.
In addition, there is a decreasing trend for the value of $l_{\perp}$ among Config C, Config D and STGAN-WO, verifying the validity of weight decomposition and orthogonal regularization on attribute disentanglement again.

It is seen that the values of $l_{w_2}$ in Config C and Config D are exceptionally small, which can be explained as follows.
Considering the contribution of weight decomposition and orthogonal regularization on attribute disentanglement, $l_{w_2}$ for Config C and Config D are expected to have larger value than that in STGAN-WO. However, $l_{w_2}$ for Config C and ConfigD is of rather small value as listed in Table \ref{t1}. We reason that this is due to the limited diversity of generated samples in Config C and Config D.
As shown in Fig. \ref{fig:fig8}, changing $w_2$ only affects few structure related attributes, greatly limiting the diversity of generated samples, and leading to the exceptional small value of $l_{w_2}$ in Config C and Config D.

Thirdly, the orthogonal regularization and the structure-texture independent architecture would decrease the image quality of synthesized samples in Config C, Config D and STGAN-WO. 
As indicated by (\ref{key7}), it is obvious that the orthogonal regularization would greatly restrict the capacity of the weight $U$ and $V$, thus decreasing the image quality for those settings where the orthogonal regularization is applied.
In a similar way, utilizing the structure-texture independent architecture to hierarchically generate the texture and structure parts requires the discriminator to capture the underlying distributions that the texture and structure components obey, and requires the coarse layers and fine layers in the generator to synthesize the corresponding parts, which obviously is a more complex task. As a result, applying the orthogonal regularization or the structure-texture independent architecture has seen decreased performance on image quality.
Notwithstanding, the orthogonal regularization and the structure-texture independent architecture are practical technique for their contribution to attribute disentanglement. Furthermore, as shown in Fig. \ref{fig:fig1} and Fig. \ref{fig:fig9}, the synthesized images are still of very good quality

To sum up, we experimentally confirm that  weight decomposition technique, orthogonal regularization and structure-texture independent architecture contribute to attribute disentanglement, which is implemented at the cost of a slight decreasing in image quality.

\subsection{InterFaceGAN vs STGAN-WO}
StyleGAN \cite{Karras2019} and its sequel StyleGAN2 \cite{Karras2020} provide a novel generative scheme, enabling to synthesize faces of high quality and fidelity. However, StyleGAN and StyleGAN2 suffer from the entanglement problem that controlling a specific attribute would inevitably affect the others, \textit{e.g.}, as shown in Fig. \ref{fig:fig7} (a), moving the latent code $w$ towards its orthogonal direction would affect multiple attributes, and it is unclear how to edit the attribute individually.
To address this, STGAN-WO intends to learn a disentangled representation in an unsupervised way, in contrast to the supervised scheme like InterFaceGAN \cite{Shen2020_cvpr,Interfacegan2020}.
To clearly show their differences, we conduct comparisons between STGAN-WO and InterFaceGAN, as a representation of the supervised scheme.

InterFaceGAN provides a novel semantic face editing method by moving the latent code along a particular direction, which is obtained under the supervision of an auxiliary attribute prediction model, and annotations
from the CelebA \cite{liu2015faceattributes} datasets are utilized to train the attribute prediction model, which in turn enable InterFanceGAN to edit attributes according to users’ demands. As a comparison, STGAN-WO as well as another unsupervised schemes, \textit{e.g.}, StyelGAN, are not able to determine which attributes to edit in advance as no annotation or label is used in the training process. To conduct semantic face editing, STGAN-WO would randomly initialize some orthogonal directions, then move the latent code towards these directions and investigate which attribute has been changed by a particular label code. From this perspective, supervised schemes provide more convenience for attribute editing than the unsupervised one.

As we can see, InterFaceGAN depends on the annotation to train the attribute prediction model, and obtaining such datasets are rather labor-consuming. Besides, InterFaceGAN can only edits attributes which have been annotated in the dataset. Specifically, InterFaceGAN can only edit five facial attributes, \textit{i.e.}, pose, smile (expression), age, gender, and eyeglasses. 
In contrast, STGAN-WO learns a disentangled representation of the face distribution, and it can edit much more attributes than InterFaceGAN. As shown in Fig. \ref{fig:fig1}, STGAN-WO is able to edit attributes like hat, face shape, hair, \textit{etc.} in addition to those in InterFaceGAN. Furthermore, STGAN-WO achieves this in an unsupervised way, \textit{i.e.}, no annotation is used in the entire training process. From this perspective, the unsupervised method shows its superiority over the supervised method.

To visualize their differences, Fig. \ref{fig:fig9} plots the face aging results of STGAN-WO and InterFaceGAN.
As we can see, there are two stages of aging: (a) from teenager to adult, and (b) from adult to old age. Obviously, aging from teenager to adult would mostly affect the shapes of faces. As a comparison, aging from adult to old age mainly affect structure related attributes and the face shape would hardly be changed.
STGAN-WO can easily accomplish the attribute editing task of face aging. Specifically, we can change $w_1$ to show the effect of aging stage (a) on the face outline as shown in  Fig. \ref{fig:fig9} (a), and change $w_2$ to figure out stage (b) of aging as shown in Fig. \ref{fig:fig9} (b).
InterFaceGAN has also successfully accomplished the face aging task as seen in Fig. \ref{fig:fig9} (c). 

We hold the opinion that both supervised and unsupervised schemes are important
as each has their own advantages,  together they complement each other and enrich the techniques available to users.

\section{Ablation Study}
In this section, we conduct ablation study to illustrate the implication of (\ref{key6}), and to demonstrate that $z_1$ and $z_2$ control the synthesis of the texture and structure parts independently.

\subsection{Effect of Attribute Editing on $S$} \label{ap1}
\begin{figure}[htp]
	\centering
	\begin{subfigure}[b]{0.475\textwidth}
		\centering
		\includegraphics[height=0.24\linewidth]{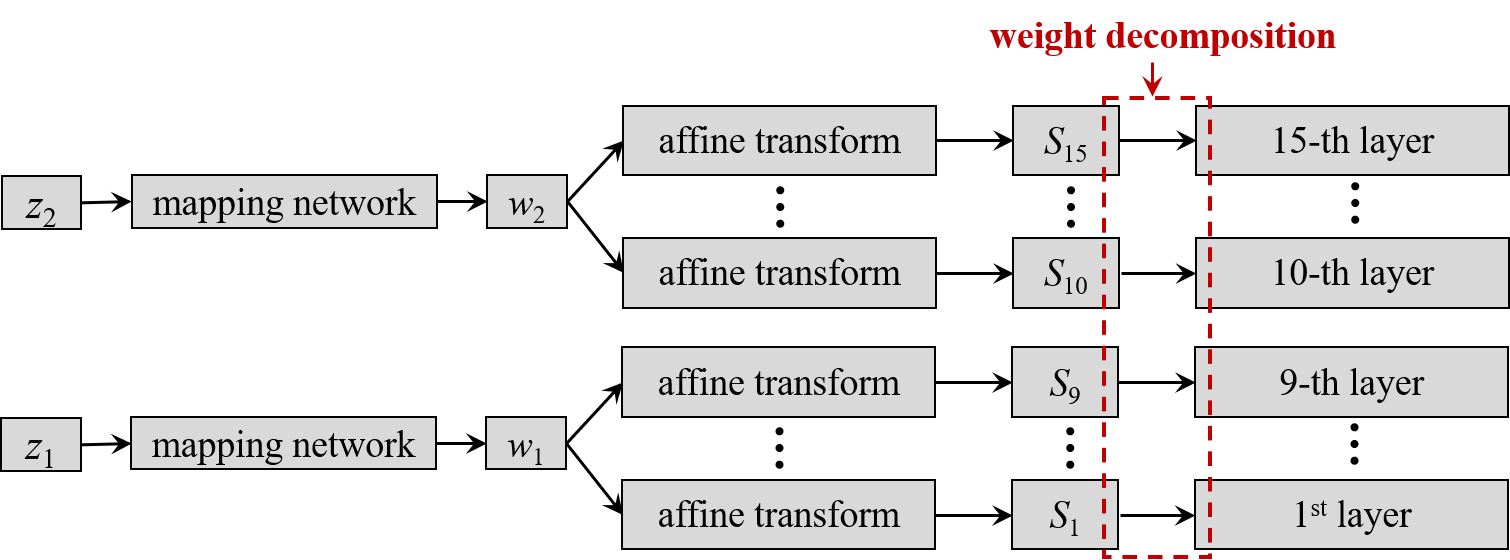}	
	\end{subfigure}	
	\caption{Illustration of $G$ architecture}
	\label{fig:figa1}
	\begin{subfigure}[b]{0.475\textwidth}
		\centering
		\includegraphics[height=0.24\linewidth]{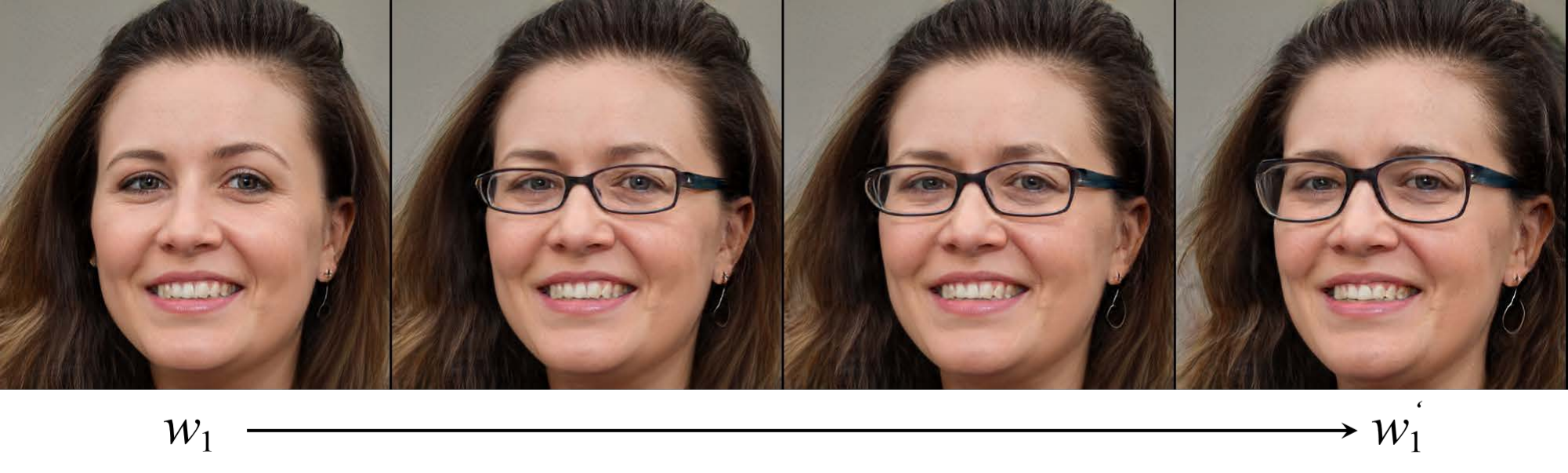}	
	\end{subfigure}		
	\caption{The influence of $w_1$ on facial attributes.}
	\label{fig:figa2}
\end{figure}
Equation (\ref{key6}) may cause the misunderstanding that each entry $s_n$ corresponds to a particular attribute, and attribute editing is achieved by changing $s_n$. To explain this clearly, we show a specific example.
Fig. \ref{fig:figa1} shows how the latent code $z$ produces its corresponding style vectors to control the image synthesis process. Specifically, there are a total of 15 layers in the generator of STGAN-WO. When conducting face synthesis, the latent vector $z_1$ and $z_2$ would produce their corresponding $w_1$ and $w_2$ first, then produce 9 and 6 style vectors, respectively, each of which is responsible for one particular layer.
Manipulating the latent code $z$ or $w$ would affect the generated faces. For example, we move the latent code $w_1$ along its orthogonal direction and obtain $w_1^{'}$, then the synthesized faces would be changed as shown in Fig. \ref{fig:figa2}. It is clearly seen that the attribute "eyeglass" has been edited while the others are hardly affected.
Let $S_i, S_i^{'} (i = 1, \cdots, 15)$ denote the style vector produced by $w_1$, and $w_1^{'}$, respectively.
\begin{equation}\label{key0}
	S_i, S_i^{'} \in \left\{\begin{matrix}
		\mathbb{R}^{512}& i = 1, \cdots, 9   \\ 
		\mathbb{R}^{256}& i = 10, 11  \\
		\mathbb{R}^{128}& i = 12, 13  \\
		\mathbb{R}^{64}&  i = 14, 15 \\
	\end{matrix}\right.	 
\end{equation}

We utilize $p_i$ to indicate how moving $w_1$ towards $w_1^{'}$ affects each entry of $S_i$, which is element-wisely computed by:
\begin{equation}\label{keya1}
	p_i = \frac{|S_i-S_i^{‘}|}{|S_i|}
\end{equation}

\begin{table}[htp]
	\centering
	\caption{The effect of semantic face editing on the entries in $S_i$. }
	\resizebox{85mm}{8mm}{
		\begin{tabular}{c c c c c c c c c c}
			\hline\hline	
			&$S_1$ & $S_2$ & $S_3$ & $S_4$ & $S_5$ & $S_6$ & $S_7$ & $S_8$ & $S_9$ \\
			\hline	
			$N(p_i \leqslant 0.1)$&67&77&71&62&71&75&78&72&66\\
			$N(0.1<p_i< 0.5)$&207&212&216&222&207&220&202&216&199\\
			$N(p_i \geqslant 0.5)$&238&223&225&228&234&217&232&224&247\\		
			\hline\hline
		\end{tabular}
	}
	\label{t2}
\end{table}
The results are listed in Table \ref{t2}. $N(p_i \leqslant 0.1)$ is the number of entries in $p_i$, whose value is in the range of $[0, 0.1]$. It is clearly seen that $N(p_i \leqslant 0.1)$ indicates how many entries are slightly affected when moving $w_1$ towards $w_1^{'}$, and $N(p_i \geqslant 0.5)$ shows those entries which have been greatly changed.

Suppose that each entry of $S_i$ corresponds to one particular attribute, then in the case of Fig. \ref{fig:figa2}, moving $w_1$ towards $w_1^{'}$ would only affect few entries of $S_i$, because only the attribute "eyeglass" is affected. However, as shown in Table \ref{t2}, most of the entries in $S_i$ have been greatly changed, indicating that each entry does not correspond to one particular attribute.

\subsection{Ablation Study about Independent Synthesis} \label{ap2}
Fig. \ref{fig:fig34} has demonstrated that $z_1$ is responsible for the synthesis of texture components, and $z_2$ controls that of the structure parts.
In addition, we utilize the structure-texture decomposition algorithm \cite{Xu2012} to obtain the structure and texture components of $\hat{I_9}$, then compute the PPL based on the structure and texture components.
Thus, we can quantitatively determine how $w_1$ and $w_2$ affect the synthesis of the structure and texture components. Results are listed in Table \ref{t3}.
It is clearly seen that changing $w_2$ hardly affects the texture components as the $l_{w_2}$ for the texture components is much smaller than that of the structure components. In a similar way, $w_1$ is closely related to the texture components, and hardly affects the structure one.

\begin{table}[hp]
	\centering
	\caption{Ablation study to show that $w_1$ and $w_2$ can control structure and texture components independently.}
	\begin{tabular}{c c c }
		\hline\hline	
		&$l_{w_2}$& $l_{\perp}$\\
		\hline
		STGAN-WO&13.04&0.42\\
		Texture components&0.12&0.42\\
		Structure components&12.20&0.04\\
		\hline\hline
	\end{tabular}
	\label{t3}
\end{table}
%
%
%

\section{Concluding Remarks}
To perform better facial attribute editing in an unsupervised way, we have introduced (a) weight decomposition, (b) orthogonal regularization, (c) the structure-texture independent architecture, and proposed STGAN-WO.  We experimentally confirm that the proposed techniques contribute to better attribute editing, \textit{i.e.}, individual attribute editing can be achieved by STGAN-WO via moving the intermediate latent code $w_1$ along its randomly initialized directions or changing $w_2$. 


%

\ifCLASSOPTIONcaptionsoff
  \newpage
\fi



%
%
%
\bibliographystyle{IEEEtran}
\bibliography{IEEEexample}
%




\end{document}